\documentclass[letterpaper]{article} 
\usepackage{aaai24}  
\usepackage{times}  
\usepackage{helvet}  
\usepackage{courier}  
\usepackage[hyphens]{url}  
\usepackage{graphicx} 
\usepackage{natbib}  
\usepackage{caption} 
\usepackage{amsmath}
\usepackage{amssymb}
\usepackage{amsthm}
\usepackage{subfiles}
\usepackage{subcaption}
\usepackage{algorithm}
\usepackage{algorithmic}

\usepackage{newfloat}
\usepackage{listings}
\DeclareCaptionStyle{ruled}{labelfont=normalfont,labelsep=colon,strut=off} 
\floatstyle{ruled}
\newfloat{listing}{tb}{lst}{}
\floatname{listing}{Listing}
\newtheorem{theorem}{Theorem}
\newtheorem{lemma}[theorem]{Lemma}
\newtheorem{remark}[theorem]{Remark}
\newcommand{\independent}{\perp \!\!\! \perp}
\makeatletter
\renewcommand*\env@matrix[1][*\c@MaxMatrixCols c]{%
  \hskip -\arraycolsep
  \let\@ifnextchar\new@ifnextchar
  \array{#1}}
\makeatother
\title{Tightening Bounds on Probabilities of Causation By Merging Datasets}
\author {
    Numair Sani\textsuperscript{\rm 1},
    Atalanti A. Mastakouri\textsuperscript{\rm 2}
}
\affiliations {
    \textsuperscript{\rm 1}Johns Hopkins University\\
    \textsuperscript{\rm 2}Amazon Research\\
    snumair1@jh.edu, atalanti@amazon.de
}

\usepackage{bibentry}

\begin{document}

\maketitle

\begin{abstract}
Probabilities of Causation (PoC) play a fundamental role in decision-making in law, health care and public policy. Nevertheless, their point identification is challenging, requiring strong assumptions, in the absence of which only bounds can be derived. Existing work to further tighten these bounds by leveraging extra information either provides numerical bounds, symbolic bounds for fixed dimensionality, or requires access to multiple datasets that contain the {\it same} treatment and outcome variables. However, in many clinical, epidemiological and public policy applications, there exist \textit{external} datasets that examine the effect of {\it different} treatments on the same outcome variable, or study the association between covariates and the outcome variable. These external datasets cannot be used in conjunction with the aforementioned bounds, since the former may entail different treatment assignment mechanisms, or even obey different causal structures. Here, we provide \textit{symbolic} bounds on the PoC for this challenging scenario. 
We focus on combining either two randomized experiments studying different treatments, or a randomized experiment and an observational study, assuming causal sufficiency. Our symbolic bounds work for arbitrary dimensionality of covariates and treatment, and we discuss the conditions under which these bounds are tighter than existing bounds in literature. Finally, our bounds parameterize the difference in treatment assignment mechanism across datasets, allowing the mechanisms to vary across datasets while still allowing causal information to be transferred from the external dataset to the target dataset.

\end{abstract}

\section{Introduction}
Probabilities of Causation (PoC) play a fundamental role in decision-making in law, health care and public policy \cite{mueller2022personalized, faigman2014group}. For example, in medical applications, if a medication for a disease has similar side effects to the disease itself, we must calculate the probability that the adverse side effect was caused by the medication, for safety assessments. In epidemiology, we often need to determine the likelihood that a particular outcome is caused by a specific exposure, or if a particular subgroup that experienced an adverse outcome would benefit from an intervention. Probabilities of Causation defined in \cite{pearl2009causality} provide a logical framework to reason about such counterfactuals, as well as necessary assumptions required to identify them from the observed data.

While causal parameters such as the Average Treatment Effect are \emph{point identified} from the observed data distribution, under reasonable assumptions, such as exogeneity, PoC are not. Specifically, in addition to exogeneity, these require monotonicity to hold for point identification \citep{pearl2022probabilities, khoury1989measurement}. However, when we do not have sufficient justification for assuming monotonicity, then, the PoC are no longer \emph{point identified}. Rather, PoC are \emph{partially identified}, i.e. bounded as a function of the observed data. 

There exists a rich literature on partial identification and its use in bounding causal quantities; some examples include \cite{manski1990nonparametric, tian2000probabilities, balke1994counterfactual, robins1989probability, dawid2017probability}. The specific question of bounding PoC has also been explored in \cite{tian2000probabilities, zhang2022partial,cuellar2018causal, robins1989probability, dawid2017probability, padh2022stochastic, sachs2022general}, however, this body of literature assumes the joint probability distribution for the variables of interest is known. Our contribution differs fundamentally from existing work in that we do not assume access to the joint probability distribution for the variables of interest. Rather, we consider the case where multiple datasets of non-overlapping treatments that study the same outcome are given. Specifically, consider the \emph{target} dataset containing treatment $X$ and outcome $Z$. In addition, we are given an \emph{external} dataset that studies the same outcome $Z$, and contains different treatment or covariates $Y$ which are randomised and independent of $X$. We demonstrate how to merge the external dataset with the target dataset to tighten the bounds on the PoC of $X$ on $Z$, while allowing for the treatment $X$ to be confounded in the external dataset. We demonstrate the importance of this scenario using the following example. 

Suppose we have treatment $X$ (medication) for a disease, and an adverse side effect $Z$, that could be caused by the disease or the medication itself. Suppose $X$ is randomized, and for safety reasons, we are interested in calculating the probability that $X$ caused $Z$. The Probabilities of Causation provide a logical framework to reason about this probability, so we aim to calculate it using the observed data. Since we don't know whether the medication is protective or harmful, we are not justified in assuming monotonicity. Then, the Probabilities of Causation are no longer point identified, and we must settle for bounds on them. Given the target dataset containing observations on $X$ and $Z$, these bounds can be calculated - however, these bounds may be too wide for us to arrive at a conclusion. Re-running the study on the same population while recording a richer set of covariates is not an option due to time and financial constraints. This raises the question ``Can other external datasets, studying the same adverse effect on similar populations, but not necessarily studying the same treatment, be leveraged to tighten the bounds on the PoC in the target dataset?''

While \citealp[]{zhang2022partial, li2022probabilities, pearl2022probabilities, cuellar2018causal, dawid2017probability} tighten bounds on the Probabilities of Causation (and more generally, other counterfactuals), they all assume access to a dataset recording all of the variables of interest. Therefore, such bounds cannot be straightforwardly applied to use cases like the one we described above. To address this gap, in this paper we focus on the challenging cases that assume only access to datasets that contain the same outcome, but do not record $X$ and $Y$ at the same time. These datasets can differ in their treatment assignment mechanism for $X$, allowing it to be randomized or confounded (assuming a sufficient set of confounders is observed). While \citealp[]{duarte2021automated, zeitler2022causal} do not need access to the joint distributions over all the variables, the bounds that they provide are only \textit{numerical}. Here, we provide \textit{symbolic} bounds. Additionally, the linear programming approach in \cite{balke1994counterfactual} cannot be applied for the bound estimation since the dual of the linear program would still have symbolic constraints. From a non-causal perspective, \citealp[]{charitopoulos2018multi} present a symbolic linear programming approach. However, knowing which constraints to supply to the linear program to derive bounds in the problem we describe requires knowledge of the causal graph and invariances. As we show in this paper, these constraints are not trivial to derive. Moreover, the solution in \cite{charitopoulos2018multi} only works for fixed dimensionality for $Y$, since different dimensionalities of $Y$ correspond to different linear programs. In our work, we explicitly target the aforementioned use case allowing arbitrary dimensionality of $Y$. 

\paragraph{Structure and Contributions} We start by reviewing the structural causal model (SCM) framework and its semantics for counterfactual reasoning (\S \ref{sec:preliminaries}). Counterfactuals are the basis of the Probabilities of Causation, which we introduce in \S \ref{sec:probabilities-of-causation}. In this section, we formally state the Probability of Sufficiency and Necessity, and we review existing bounds on it. In \S \ref{sec:merge-ext-targ} we describe the data-generating process for the target and external datasets. We then describe the assumed invariances, and how these can be used to transfer information across datasets. Leveraging this invariance principle, we present theorems that provide symbolic bounds on the PoC in our target dataset, after merging it with an external dataset containing a randomized covariate or treatment of arbitrary dimensionality. In \S \ref{sec:merging-experimental-observational} we further relax the strict assumption of $X$ being randomized in the external dataset, and provide symbolic bounds in the presence of observed confounding. We conclude with remarks in \S \ref{sec:discussion}. We provide all our proofs and derivations in the Technical Appendix.

\section{Preliminaries}\label{sec:preliminaries}
While there exist many formulations of causal models in the literature, such as the Finest Fully Randomized Causally Interpretable Structured Tree Graph (FFRCISTG) of \cite{robins1986new} and the agnostic causal model of \cite{spirtes2000causation}, in this work, we utilise the SCM defined in \cite{pearl2009causality}. 
Formally, a SCM $\mathcal{M}$ is defined as a tuple $\langle \mathbf{U}, \mathbf{V}, \mathcal{F}, \mathbb{P} \rangle$ where $\mathbf{U}$ and $\mathbf{V}$ represent a set of exogenous and endogenous random variables respectively. $\mathcal{F}$ represents a set of functions that determine the value of $V \in \mathbf{V}$ through $v \leftarrow f_V(pa_V, u_V)$ where $pa_V$ denotes the parents of $V$ and  $u_V$ denotes the values of the noise variables relevant to $V$. $\mathbb{P}$ denotes the joint distribution over the set of noise variables $\mathbf{U}$, and since the noise variables $\mathbf{U}$ are assumed to be mutually independent, the joint distribution $\mathbb{P}(\mathbf{U})$ factorises into the product of the marginals of the individual noise distributions. $\mathcal{M}$ induces an observational data distribution on $\mathbf{V}$, and is associated with a Directed Acyclic Graph (DAG) $\mathcal{G}$. 

Defining an SCM allows us to define submodels, potential responses and counterfactuals, as defined in \cite{pearl2009causality}. Given a causal model $\mathcal{M}$ and a realisation $x$ of random variables $\mathbf{X} \subset \mathbf{V}$, a submodel $\mathcal{M}_x$ corresponds to deleting from $\mathcal{F}$ all functions that set values of elements in $\mathbf{X}$ and replacing them with constant functions $X = x$. The submodel captures the effect of intervention $do(X = x)$ on $\mathcal{M}$. Given a subset $\mathbf{Y} \subset \mathbf{V}$, the potential response $\mathbf{Y}_x(u)$ denotes the values of $Y$ that satisfy $\mathcal{M}_x$ given value $u$ of the exogenous variables $\mathbf{U}$. Thus, the counterfactual $\mathbf{Y}_x(u) = y$ represents the scenario where the potential response $\mathbf{Y}_x(u) $ is equal to $y$, if we possibly contrary to fact, set $X = x$. When $u$ is generated from $P(\mathbf{U})$, we obtain counterfactual random variables $\mathbf{Y}_{x}$ that have a corresponding probability distribution. Counterfactual random variables lie on rung three of the Ladder of Causation \cite{pearl2009causality}, needing additional assumptions for their identification. In the following section, we explore a special class of counterfactual probabilities, known as the Probabilities of Causation.

\section{Probabilities of Causation}\label{sec:probabilities-of-causation}
\begin{figure}[h]
    \centering
    \includegraphics{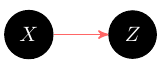}
    \caption{Causal graph containing treatment $X$ and outcome $Z$, where treatment $X$ is randomized}
    \label{fig:single-rct-graph}
\end{figure}

An important class of counterfactual probabilities that have applications in law, medicine, and public policy are known as the Probabilities of Causation \cite{pearl2009causality}. These are a set of five counterfactual probabilities related to the Probability of Necessity and Sufficiency ($PNS$), which we define as follows. Consider the causal graph in Fig. \ref{fig:single-rct-graph}, where the outcome $Z$ and treatment $X$ are binary random variables.

Let $x$ denote the event that the random variable $X$ has value $1$ and let $x^{\prime}$ denote the event that $X$ has value $0$. Then, the $PNS$ is defined as
\begin{align}
    PNS \equiv P(Z_x = 1, Z_{x^{\prime}} = 0)
\end{align}
$PNS$ represents the joint probability that the counterfactual random variable $Z_x$ takes on value $1$ and the counterfactual random variable $Z_{x^{\prime}}$ takes on value $0$. Under conditions of exogeneity, defined as $Z_x \independent X$, the rest of the Probabilities of Causation such as Probability of Necessity ($PN$) and Probability of Sufficiency ($PS$), are all defined as functions of $PNS$ (see Theorem 9.2.11 in \cite{pearl2009causality}). Consequently, when $PNS$ is identified, all the other Probabilities of Causation are straightforwardly identified from the observed data as well. 

\begin{figure}[h]
    \centering
    \includegraphics{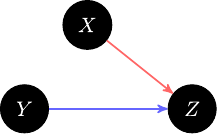}
    \caption{Causal graph containing binary treatment $X$, binary outcome $Z$ and  additional treatment or covariate $Y$}
    \label{fig:single-rct-with-covariates}
\end{figure}

However, to identify $PNS$, we must make assumptions such as \emph{monotonocity} \citep{tian2000probabilities}, which may not be justified in settings involving experimental drugs, legal matters and occupational health. Without this assumption, $PNS$ is no longer point identified, but it can still be meaningfully bounded using tools from the partial identification literature. Assuming still the graph in Fig. \ref{fig:single-rct-graph}, an important bound on $PNS$ is defined in \citet{tian2000probabilities}, and is presented below, with $p_{11}$, $p_{10}$ and $p_{00}$ denoting $P(Z = 1 \mid X = 1)$, $P(Z = 1 \mid X = 0)$, and $P(Z = 0 \mid X = 0)$ respectively. 
\begin{equation}\label{eq:pearls-bounds}
    \max\begin{bmatrix}
        0\\
        p_{11} - p_{10} 
    \end{bmatrix} 
    \leq PNS \leq
    \min\begin{bmatrix}
        p_{11}\\
        p_{00}
    \end{bmatrix}
\end{equation}
Given additional data (i.e. $Y$ in Fig. \ref{fig:single-rct-with-covariates}), the bounds on $PNS$ can be further tightened, as shown in \citet{dawid2017probability}. We present one of their results here, as we utilize this to tighten the bounds on the Probabilities of Causation by merging target and external datasets. 
In Fig. \ref{fig:single-rct-with-covariates}, $X$ and $Z$ represent the treatment and outcome respectively, and $Y$ represents additional treatments or an additional set of covariates. Then, \citealp[]{dawid2017probability} show that given $P(Z, X, Y)$, the bounds on $PNS$ can be further tightened as
\footnote{While \citealp[]{dawid2017probability} provide bounds on a different counterfactual probability called $PC$, its relation to $PNS$ is described in Theorem 9.2.11 in \cite{pearl2009causality}}
\begin{equation}\label{eq:pns-dawid-bounds-main}
    \Delta \leq PNS \leq P(Z = 1 \mid X = 1) - \Gamma
\end{equation}
Where
\begin{align*}
    \Delta &= \sum_y P(Y = y)\max \{0, P(Z = 1 \mid X = 1, Y = y)\\ & \hspace{1in}- P(Z = 1 \mid X = 0, Y = y)\} \\
    \Gamma &= \sum_y P(Y = y)\max \{0, P(Z = 1 \mid X = 1, Y = y)\\
    &\hspace{1in} - P(Z = 0 \mid X = 0, Y = y)\} 
\end{align*}
\citealp[]{dawid2017probability} show that this interval is always contained in the one given in Eq. \ref{eq:pearls-bounds}. Note that the bounds in Eq. \ref{eq:pns-dawid-bounds-main} assume access to the joint distribution $P(Z, X, Y)$.

In the use case we tackle in this paper, we do not have this luxury. On the contrary, we are given a \textit{target} dataset that studies a treatment $X$ and an outcome $Z$, and additional information in the form of an \textit{external} dataset with the same outcome $Z$, that studies a different treatment (or covariate) $Y$ which is randomized and independent of $X$. We denote \footnote{We abuse notation by denoting all distributions associated with the target dataset by $P^T$. A similar approach is used for $P^E$} the distribution associated with the target dataset as $P^T$, and the distribution associated with the external dataset as $P^E$. We are interested in using this external dataset to tighten the bounds on the PNS of $X$ on $Z$ in our target dataset. In other words, we do not have access to the joint distribution $P^T(Z, X, Y)$ for our target dataset, rather, we only have access to $P^T(Z, X)$. Hence, we cannot straightforwardly apply the bounds in Eq. \ref{eq:pns-dawid-bounds-main}.

To this end, we borrow information from the external dataset to constrain the possible choices for the \textit{target} distribution $P^T(Z, X, Y)$. Typically, the target distribution is not identified, and as such, we constrain the set of possible distributions compatible with both our target and external datasets. Then, we can utilize Eq. \ref{eq:pns-dawid-bounds-main} to pick the most conservative bounds implied by the set of joint distributions which are compatible with the target and the external dataset. 

Formally, let $i$ index this set of compatible joint distributions. Then the most conservative bound will be the smallest lower bound on $PNS$, and the greatest upper bound on $PNS$. Denoting $P^T(Z = 1 \mid X = 1)$ as $p^T_{11}$, the bounds are given as
\begin{equation}\label{eq:max-baby-form}
\min_{P^T_i} \Delta(P^T_i) \leq PNS \leq \max_{P^T_i} p^T_{11} - \Gamma(P^T_i)
\end{equation}
From the properties of the $\max$ operator, and since $p^T_{11}$ is known, the bounds in Eq. \ref{eq:max-baby-form} and are re-written as
\begin{equation}\label{eq:max-adult-form}
    \min_{P^T_i} \Delta(P^T_i) \leq PNS \leq p^T_{11} - \min_{P^T_i}\Gamma(P^T_i)
\end{equation}
This raises the question of how exactly to utilize the external dataset over $Z$ and $Y$ to constrain the set of possible possible joint distributions $P^T(Z, X, Y)$ for our target dataset. To this end, we utilize the principle of independent causal mechanisms (see Definition 4 in \citet{janzing2010causal}, \citet{scholkopf2012causal} and Principle 2.1 in \citet{peters2017elements}) to transfer causal information across datasets.

\section{Merging Target and External Datasets}\label{sec:merge-ext-targ}
\begin{figure*}[]
    \centering
    \begin{subfigure}{0.25\textwidth}
        \centering
        \includegraphics[width=0.95\textwidth]{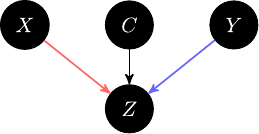}
        \caption{}\label{tr1}
    \end{subfigure}%
    \begin{subfigure}{0.25\textwidth}
        \centering
        \includegraphics[width=0.95\textwidth]{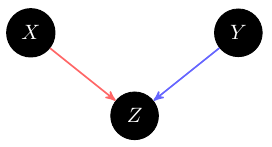}
        \caption{}\label{v_structure}
    \end{subfigure}%
    \begin{subfigure}{0.25\textwidth}
        \centering
        \includegraphics[width=0.95\textwidth]{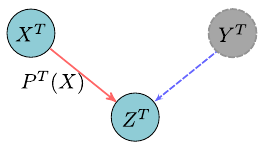}
        \caption{}\label{v1}
    \end{subfigure}%
    \begin{subfigure}{0.25\textwidth}
        \centering
        \includegraphics[width=0.95\textwidth]{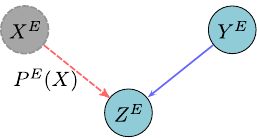}
        \caption{}\label{v2}
    \end{subfigure}%
    \caption{Causal graphs representing the different types of interactions between the outcome $Z$, the treatments $X$ and $Y$, and the covariates $C$. (a) A causal graph where both treatments $X$ and $Y$ are assigned randomly, and together with $C$ determine the outcome $Z$. (b) A causal graph where the causal mechanisms in the target and external datasets are the same for $X$ and $Y$, i.e. $P^T(X) = P^E(X)$ and $P^T(Y) = P^E(Y)$, and as such, $P^T(Z, X)$ and $P^E(Z, Y)$ are the result of marginalizing $Y$ and $X$ from $P(Z, X, Y)$ respectively. (c) A causal graph describing the data-generating process for our target dataset. In the latter, the treatment $X$ is assigned randomly and is observed (cyan node), and the node $Y$ is not observed (grey node). (d) A causal graph describing the data-generating process for our external dataset. Here, although the treatment $X$ is still assigned randomly, its mechanism differs from that in the target dataset and is not observed. As we have already stated, we assume $P^T(Y) = P^E(Y)$.}
    \label{fig:invariant-graphs}
\end{figure*}
To transfer causal information from our external dataset to our target dataset, we must first identify causal quantities that remain invariant across these different data sources. Similar approaches of defining invariant quantities and utilizing them to borrow information across datasets have been used in transportability, distribution shift and robustness \cite{pearl2011transportability, christiansen-generalization, buhlmann2020invariance}. 
Throughout this paper, we assume we are given a treatment variable $X$, a set of covariates $C$, and either an additional treatment or covariate $Y$, along with outcome $Z$, where $Z$ is a causal descendant of $X$, $Y$ and $C$. Motivated by the principle of independent mechanisms (see Principle 2.1 of \citet{peters2017elements}), we assume the interventional distribution $P(Z_{x, y, c})$ is invariant across datasets. This assumption is justified given a rich enough set of covariates $C$, and similar assumptions have been made by other works in the literature \cite{christiansen-generalization, muandet2013domain, daume2006domain}.

This invariance is assumed in all the data-generating processes shown in Fig. \ref{fig:invariant-graphs} and Fig. \ref{fig:invariant-graphs-experimental-observational}, representing scenarios where we combine datasets with different study designs. Throughout this paper, to avoid cases with undefined quantities, we assume that $0 < P^T(X) < 1$, $0 < P^T(Y) < 1$ as well as $0 < P^E(X) < 1$ and $0 < P^E(Y) < 1$.

\subsection{Merging Datasets With Randomized Treatments}\label{sec:borrowing-information-randomized}
First, we consider the case where the target dataset contains an outcome $Z$ and a randomized treatment $X$, and the external dataset contains the same outcome $Z$, and either a different randomized treatment or additional randomized covariate $Y$, while still having $X$ assigned randomly. Unobserved covariates $C$ that are independent of $X$ and $Y$ can exist, and these are marginalized out.

Under these assumptions, we derive invariances across our target and external datasets. We denote the distribution associated with our target dataset as $P^T(Z, X)$, and the one associated with our external dataset as $P^E(Z, Y)$. In the scenario we consider, $P^T(Z, X, Y, C)$ and $P^E(Z, X, Y, C)$ \footnote{Note, that we do not have access to these joint distributions.} obey the causal structure in Fig. \ref{tr1}. Then the interventional distributions are:
\begin{align*}
    P^T(Z_{x, y}) &= P^T(Z \mid x, y) = \sum_{c} P(Z \mid x, y, c)P(c)\\
    P^E(Z_{x, y}) &= P^E(Z \mid x, y) = \sum_{c} P(Z \mid x, y, c)P(c)
\end{align*}
Consequently, when both the populations considered in our internal and external datasets have the same distribution of covariates $P(C)$, we expect $P(Z_{x, y})$ to be invariant across $P^T$ and $P^E$. To transfer causal information from the external dataset to the target dataset, we must also consider the data-generating process for $P^T(Z, X)$ and $P^E(Z, Y)$. Specifically, the distribution for the target dataset $P^T$ can be factorized as
\begin{align*}
    P^T(Z = 1\mid X = 1) &= P(Z \mid X = 1, Y = 1)P^T(Y = 1) \\
    & + P(Z \mid X = 1, Y = 0)P^T(Y = 0)
    \intertext{Here, $P(Z_{x, y}) = P(Z = 1 \mid X = 1, Y = 1)$ since $Z_{x, y} \independent X, Y$. Similarly, the external distribution can be factorized as}
    P^E(Z = 1\mid Y = 1) &= P(Z \mid X = 1, Y = 1)P^E(Y = 1) \\
    & + P(Z \mid X = 0, Y = 1)P^E(Y = 0)
\end{align*}
When $P^E(X) = P^T(X)$, and $P^E(Y) = P^T(Y)$, both $P^E(Z, Y)$ and $P^T(Z, X)$ can be viewed as marginalized distributions obtained from a joint distribution over $P(Z, X, Y)$, where $X$, $Y$ and $Z$ follow the collider shaped causal structure given in Fig. \ref{v_structure}. Then, the target dataset and external dataset can be used to constrain $P^T(Z, X, Y)$ as
\begin{align*}
    P^T(Z = 1 \mid X = 1) &= \sum_{y} P(Z = 1 \mid X = 1, Y = y)\\
    &\hspace{0.5in}\times P^E(Y = y)\\
    \vdots\\
    P^E(Z = 1 \mid Y = 1) &= \sum_{x} P(Z = 1 \mid X = x, Y = 1)\\
    &\hspace{0.5in}\times P^T (X = x).\\
\end{align*}
Since $X \independent Y$, $P^E(X) = P^T(X)$, and $P^E(Y) = P^T(Y)$, these constraints form a system of equations with multiple solutions for $P(Z \mid X, Y)$. When $Y$ is binary, these constraints form a system of equations with a single free parameter $P(Z = 1 \mid X = 1, Y = 1)$. Then, bounds on $PNS$ using $P^T(Z, X)$ and $P^E(Z, Y)$ are obtained by solving
\begin{align}\label{eq:minimax-dawid-bound}
    & \min_{P(Z = 1 \mid X = 1, Y = 1)} \Delta(P(Z = 1 \mid X = 1, Y = 1))\notag \\
    & \hspace{0.5in} \leq PNS \leq \notag\\
    & p^T_{11} - 
    \min_{P(Z = 1 \mid X = 1, Y = 1)}  \Gamma(P(Z = 1 \mid X = 1, Y = 1))
\end{align}
We present the bounds on $PNS$ in our target dataset when $Y$ is binary in Theorem \ref{thm:binary-case}.
\begin{theorem}\label{thm:binary-case}
Let $X$, $Y$ and $Z$ be binary random variables, obeying the causal structure in Fig. \ref{v_structure}. Then, given distributions $P^T(Z, X)$ and $P^E(Z, Y)$ where $P^E(X) = P^T(X)$ and $P^E(Y) = P^T(Y)$, the bounds of $PNS$ of $X$ on $Z$ in the target dataset are
\begin{align}
    &\max\begin{bmatrix}
            0\\
            p^T_{11} - p^T_{10}
          \end{bmatrix} \leq PNS \leq 
    \min\begin{bmatrix}
        p^T_{11} - \sum_{i = 0}^1\Phi_i\\
        p^T_{00} - \sum_{i = 0}^1\Theta_i\\
    \end{bmatrix}
\end{align}
Where $p^T_{11} = P^T(Z = 1 \mid X = 1)$, $p^T_{10} = P^T(Z = 1 \mid X= 0)$, $p^T_{00} = P^T(Z = 0\mid X = 0)$, $p_x = P^T(X = 1)$ and $p_{y_i} = P^E(Y = y_i)$, and $\Phi_i$ and $\Theta_i$ are
\begin{align*}
    &\Phi_i = \mathbb{I}(p^E_{1i} \geq \max\{1 - p_x, p_x\})\frac{p_{y_i}(p^E_{1i} - \max\{1 - p_x, p_x\})}{\min\{1 - p_x, p_x\}}\\
    &\Theta_i = \mathbb{I}(p^E_{1i} \leq \min\{1 - p_x, p_x\})\frac{p_{y_i}(\min\{1 - p_x, p_x\} - p^E_{1i})}{\min\{1 - p_x, p_x\}}
\end{align*}
\end{theorem}

The bounds in Theorem \ref{thm:binary-case} will be tighter than the bounds in Eq. \ref{eq:pearls-bounds} whenever either $P(Z = 1 \mid Y = 1)$ or $P(Z = 1 \mid Y = 0)$ is greater than the maximum of $p_x$ and $1 - p_x$, or less than the minimum of $p_x$ and $1 - p_x$. Note that these bounds recover Proposition 4 in \citet{gresele2022causal}, showing the lower bound on $PNS$ cannot be tightened. These bounds can be extended to the case when $Y$ is discrete, taking values in $\{0, \dots, N\}$. We provide bounds for this case in Theorem \ref{thm:y-multivalued}. 

\begin{theorem}\label{thm:y-multivalued}
Let $X$ and $Z$ be binary random variables, and let $Y$ be a discrete random variable taking on $N + 1$ discrete values in $\{0, 1, \dots ,N\}$. Assume $Z$, $X$ and $Y$ obey the causal structure in Fig. \ref{v_structure}. Then, given distributions $P^T(Z, X)$ and $P^E(Z, Y)$ where $P^E(X) = P^T(X)$ and $P^E(Y) = P^T(Y)$, the bounds of $PNS$ of $X$ on $Z$ in the target dataset are
\begin{align}
    &\max\begin{bmatrix}
            0\\
        p^T_{11} - p^T_{10}
        \end{bmatrix} \leq PNS \leq 
        \min\begin{bmatrix}
        p^T_{11} - \sum_{i = 0}^N\Phi_i\\
        p^T_{00} - \sum_{i = 0}^N\Theta_i\\
    \end{bmatrix}
\end{align}
The notation used is identical to that used in Theorem \ref{thm:binary-case}.
\end{theorem}
The assumptions of $P^T(X) = P^E(X)$ and $P^T(Y) = P^E(Y)$ are very restrictive - so we discuss approaches to either satisfy or relax these assumptions. 

\subsection{Mismatch Between Treatment Assignment Mechanisms}
First, the assumption $P^T(Y) = P^E(Y)$ can be satisfied by choosing a suitable $Y$, such that $P(Y)$ is invariant across our target and external datasets. An example of such a $Y$ would be the presence of a genetic mutation, which based on Mendelian Randomization, is assigned randomly, and is expected to have similar prevalence across populations that share similar characteristics. More generally, suitable choices of $Y$ would be variables that do not affect the treatment assignment of $X$ in the external or target dataset, and are expected to have identical prevalence across the target and external datasets. Next, the restrictive assumption on $P^T(X) = P^E(X)$ can be relaxed by parameterizing the difference in data generating processes between the external and target dataset using a parameter $\delta_X$, i.e. $P^E(X) = P^T(X) + \delta_X$, where $\delta_X$ is adequately restricted to ensure valid probabilities as well as $0 < P^E(X) < 1$. Using this parameterization, we constrain $P(Z \mid X, Y)$ in terms of $\delta_X$ and the given target and external datasets as
\begin{align*}
    P^E(Z = 1 \mid Y = 1) &= \sum_{x} P(Z = 1 \mid X = x, Y = 1)\\
    &\times \Big(P^T(X = x) + \delta_X\Big)\\
    &\hspace{0.5in}\vdots\\
    P^T(Z = 1 \mid X = 0) = \\
    \sum_{y} P(Z = 1 &\mid X = 0, Y = y)P^T(Y = y)
\end{align*}

Under this parameterization, the bounds in Theorem \ref{thm:binary-case} and Theorem \ref{thm:y-multivalued} can be re-derived in terms of the parameter $\delta_X$, and we present these in Theorem \ref{thm:multivalued-Y-delta}.
\begin{theorem}\label{thm:multivalued-Y-delta}
Let $X$ and $Z$ be binary random variables, and let $Y$ be a discrete random variable taking on $N + 1$ discrete values in $\{0, 1, \dots ,N\}$. Assume the target distribution $P^T(Z, X)$ and external distribution $P^E(Z, Y)$ obey the causal structure in Fig. \ref{v1} and  Fig. \ref{v2} respectively. Then, assuming $P^T(Y) = P^E(Y)$ and $P^E(X) = P^T(X) + \delta_X$, the bounds of $PNS$ of $X$ on $Z$ in the target dataset are given as
\begin{align}
\max\begin{bmatrix}
        0\\
    p^T_{11} - p^T_{10}
    \end{bmatrix} \leq PNS \leq 
    \min\begin{bmatrix}
    p^T_{11} - \sum_{i = 0}^N\Phi_{i, \delta_X}\\
    p^T_{00} - \sum_{i = 0}^N\Theta_{i, \delta_X}\\
\end{bmatrix}
\end{align}
Where $p^T_{11}$, $p^T_{10}$ and $p^T_{00}$ are defined in Theorem \ref{thm:binary-case}, and $\Phi_{i, \delta_X}$ and $\Theta_{i, \delta_X}$ are defined as
\begin{align*}
    &\Phi_i = \mathbb{I}(p^E_{1i} \geq \max\{1 - p_x - \delta_X, p_x + \delta_X \})\\
    &\times \frac{p_{y_i}(p^E_{1i} - \max\{1 - p_x - \delta_X, p_x + \delta_X\})}{\min\{1 - p_x - \delta_X, p_x +\delta_X\}}\\
    &\Theta_i = \mathbb{I}(p^E_{1i} \leq \min\{1 - p_x - \delta_X, p_x + \delta_X\})\\
    &\times \frac{p_{y_i}(\min\{1 - p_x - \delta_X, p_x + \delta_X\} - p^\prime_{1i})}{\min\{1 - p_x - \delta_X, p_x + \delta_X\}}
\end{align*}
\end{theorem}
So, introducing $\delta_X$ maintains the overall structure of the bounds introduced in Theorems \ref{thm:binary-case} and \ref{thm:y-multivalued}, but it does require the external dataset to have a stronger treatment effect ($p^E_{1i}$) of $Y$ on $Z$ to tighten the bounds on the target dataset.

While the above bounds relax the assumption $P^T(X) = P^E(X)$ by parameterizing their difference, they still require the external dataset $P^E(Z, Y)$ to have $X$ randomized. As this does not always hold in practice, in the following section we tackle a more realistic scenario; one where the treatment assignment mechanism for $X$ in $P^E$  is confounded by a set of covariates $C$. We assume that both the target and external datasets record this $C$.

\subsection{Merging Experimental And Observational Datasets}\label{sec:merging-experimental-observational}
\begin{figure*}[]
    \centering
        \begin{subfigure}{0.5\textwidth}
        \centering\includegraphics{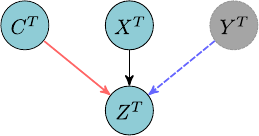}
        \caption{}\label{simple_trident}
    \end{subfigure}%
    \begin{subfigure}{0.5\textwidth}
        \centering
        \includegraphics
        {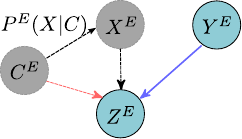}
        \caption{}\label{confounded_trident}
    \end{subfigure}%
    \caption{Causal graphs representing the different types of interactions between the outcome $Z$, treatments $X$ and $Y$, and covariates $C$. (a) A causal graph describing the data generating process for the target dataset where both $X$ and $Y$ are assigned randomly, and together with $C$ determine the outcome $Z$. (b) A causal graph describing the data-generating process for the external dataset, where $X$ is confounded by $C$. Here recall that $P^T(C) = P^E(C)$. We parameterize the difference in treatment assignment mechanisms for $X$ across datasets using $\delta^C_X$ as $P^E(X \mid C) = P^T(X) + \delta^C_X$. We denote with cyan the observed nodes in each dataset and with grey the unobserved ones.}
    \label{fig:invariant-graphs-experimental-observational}
\end{figure*}

When the treatment $X$ in the \textit{external} dataset is allowed to be confounded by a set of discrete \textit{observed} confounders $C$, the causal graph representing the data generating process for the external dataset is given in Fig. \ref{confounded_trident}. Throughout this section, we assume that $P^T(C) = P^E(C)$. Note that $X$ is still randomized in the target dataset. The causal graph corresponding to the data-generating process for the target dataset is depicted in Fig. \ref{simple_trident}. To employ a similar approach to deriving bounds as before, we must first derive bounds in the ideal case where we have access the the joint distribution $P^T(Z, X, Y, C)$ for our target dataset. Following a similar derivation to the bounds presented in \citet{dawid2017probability}, first, we derive bounds on $PNS$ for when the joint $P^T(Z, X, Y, C)$ is observed in Theorem \ref{thm:covariate-instrument-bounds}.

\begin{theorem}\label{thm:covariate-instrument-bounds}
Let $X$ and $Z$ be binary random variables, and let $C$ and $Y$ be discrete random variables. If $X$, $Y$, $Z$ and $C$ follow the causal graph in Fig. \ref{simple_trident}, then given $P(Z, X, Y , C)$, we can obtain bounds on $PNS$ as
    \begin{equation}\label{eq:trident-bounds-main}
        \sum_{C}P(C)\Delta_C \leq PNS \leq p_{11} - \sum_{C}P(C)\Gamma_C
    \end{equation}
    Where
    \begin{align*}
    \Delta_C &= \sum_y P(Y = y)\max \{0, P(Z = 1 \mid X = 1, Y = y, C)\\ & \hspace{1in}- P(Z = 1 \mid X = 0, Y = y, C)\} \\
    \Gamma_C &= \sum_y P(Y = y)\max \{0, P(Z = 1 \mid X = 1, Y = y, C)\\
    &\hspace{1in} - P(Z = 0 \mid X = 0, Y = y, C)\} 
    \end{align*}
\end{theorem}
The conditions under which these bounds are tighter than bounds in Eq. \ref{eq:pns-dawid-bounds-main} are described in Lemma \ref{lem:joint-bound-tightening-conditions}. 
\begin{lemma}\label{lem:joint-bound-tightening-conditions}
    The bounds in Eq. \ref{eq:trident-bounds-main} are contained in the bounds given in Eq. \ref{eq:pns-dawid-bounds-main}, and will be tighter when for any $C$, $P(Z = 1\mid X = 1, Y, C)$ is sometimes, but not all the time, greater than $P(Z = 1\mid X = 0, Y, C)$ or $P(Z = 0\mid X = 0, Y, C)$.
\end{lemma}

Having established bounds when given access to the joint distribution $P^T(Z, X, Y, C)$, we now derive bounds on our target dataset when the joint distribution $P^T(Z, X, Y, C)$ is unknown, but additional information is available from the external dataset. Specifically, we consider the case where the target distribution $P^T(Z, X, C)$ obeys the causal structure in Fig. \ref{simple_trident}, with treatment $X$ being randomized, and does not measure $Y$. In addition, we are given access to an external distribution $P^E(Z, Y, C)$, which obeys the causal structure in Fig. \ref{confounded_trident}. Note that in $P^E$, the treatment $X$ is confounded by $C$, while this is not the case in $P^T$. We assume $P(Z_{x,y,c})$ is invariant across these datasets, however, since $X$ is confounded by $C$ in the external dataset, we must also parameterize the difference between the treatment assignment mechanism of $X$ in $P^T$ and $P^E$. This must be done for every level of covariates $C$. Hence we index this parameter for every level $C$ as $\delta^C_X$. Then, the following constraints can be utilized to restrict the set of choices of joint distributions $P^T(Z, X, Y, C)$ compatible with the target and external dataset: 
\begin{align*}
    &P^E(Z = 1 \mid Y = 1, C) = \\
    &\sum_{x} P(Z = 1 \mid X = x, Y = 1, C)\times \Big(P^T(X = x) + \delta^C_X\Big)\\
    &\hspace{0.6in}\vdots\\
    &P^T(Z = 1 \mid X = 0, C) = \\
    &\sum_{y} P(Z = 1 \mid X = 0, Y = y, C)\times P(Y = y)
\end{align*}
Note that when $\delta^C_X = 0$ for all levels of $C$, this corresponds to the case where the external dataset has $X$ randomized as well, and additionally, both the target and external datasets are results of marginalizing the distribution $P(Z, X, Y, C)$ over $X$ and $Y$ respectively.  Similarly, note that when $\delta^{c_0}_X = \delta^{c_1}_X = \dots = \delta^{c_i}_X$ for all levels of $C$, this corresponds to the case where $X$ is randomized in the external dataset, but it has a different treatment mechanism than the target dataset. We provide bounds for arbitrary values of $\delta^C_X$ (ensuring valid probabilities) in Theorem \ref{thm:experimental-observational-bounds}.
\begin{theorem}\label{thm:experimental-observational-bounds}
Let $X$ and $Z$ be binary random variables, and let $Y$ be a discrete random variable taking on $N + 1$ discrete values in $\{0, 1, \dots ,N\}$, and $C$ be a discrete random variable taking on $M + 1$ discrete values in $\{0, \dots, M\}$. Assume $P^T(Z, X, C)$ and $P^E(Z, Y, C)$ are generated generated according to the causal graphs given in Fig. \ref{simple_trident} and \ref{confounded_trident} respectively. Then, assuming $P^T(Y) = P^E(Y)$, and $P^E(X \mid C) = P^T(X) + \delta^C_X$, then $P^T$ and $P^E$ can be merged to tighten the bounds on $PNS$ for $X$ on $Z$ in the target dataset as
\begin{equation}
    \Delta \leq PNS \leq
    \sum_{C}P^T(C)\min\begin{bmatrix}
        p^T_{11C} - \sum_{i = 0}^N\Phi^C_{i, \delta^C_X}\\
        p^T_{00C} - \sum_{i = 0}^N\Theta^C_{i, \delta^C_X}\\
    \end{bmatrix}
\end{equation}
Where $p^T_{11C} = P(Z = 1\mid X = 1, C)$, $p^T_{00C} = P(Z = 0\mid X = 0, C)$, $p^E_{1iC} = P^E(Z = 1 \mid Y = i, C)$ and $\Delta$, $\Phi^C_{i, \delta^C_x}$ and $\Theta_{i, \delta^C_x}$ are defined as
\begin{align*}
    \Delta &= \sum_{C}P^T(C)\max\{0, P^T(Z = 1 \mid X = 1, C)\\
    &\hspace{0.5in} -  P^T(Z = 0 \mid X = 1, C)\}\\
    \Phi^C_{i, \delta^C_X} &= \mathbb{I}(p^E_{1iC} \geq \max\{1 - p_x - \delta^C_X, p_x + \delta^C_X \})\\
    &\times \frac{p_{y_i}(p^E_{1iC} - \max\{1 - p_x - \delta^C_X, p_x + \delta^C_X\})}{\min\{1 - p_x- \delta^C_X, p_x + \delta^C_X\}}\\
    \Theta^C_{i, \delta^C_X} &= \mathbb{I}(p^E_{1iC} \leq \min\{1 - p_x- \delta^C_X, p_x + \delta^C_X\})\\
    &\times \frac{p_{y_i}(\min\{1 - p_x - \delta^C_X, p_x + \delta^C_X\} - p^E_{1iC})}{\min\{1 - p_x - \delta^C_X, p_x + \delta^C_X\}}
\end{align*}
\end{theorem}
These bounds allow for the target and external datasets to have different treatment assignment mechanisms for $X$, allowing for a greater variety of external datasets to be used to tighten the bounds on $PNS$ in the target dataset. Theorem \ref{thm:experimental-observational-bounds} shows that the lower bound on $PNS$ in our setting is identical to the lower bound in Eq. \ref{eq:pns-dawid-bounds-main}. However, the upper bound will be tighter whenever for any $C$, at least one $p^E_{1iC}$, but not all, is greater than $p_x + \delta^C_X$ and $1 - p_x - \delta^C_X$, or less than $p_x + \delta^C_X$ and $1 - p_x - \delta^C_X$. Theorem \ref{thm:experimental-observational-bounds} enables us to use genomics datasets that study the same outcome to tighten the bounds on PoC in the target dataset, without measuring or making restrictive assumptions on the treatment assignment mechanism for $X$ in the external genomics dataset. Now, using the Theorems presented in this paper, a variety of external datasets can be leveraged to tighten the bounds on the PoC.

\section{Discussion and Future Work}\label{sec:discussion}
Having presented various approaches to tightening the bounds on the Probabilities of Causation, we briefly highlight some key discussion points.

\textbf{Dealing with finite samples} Throughout this paper, we assume that having access to the dataset is equivalent to having access to the joint distribution over the variables contained in the dataset. In finite samples, this will not hold, and there will be additional statistical considerations. In this case, Maximum likelihood \cite{bickel2015mathematical} based approaches, as well as approximations may be used to ensure compatibility of target and external datasets.

\textbf{Assumption on prevalence parameter $\delta_X$} We provided theorems on bounds on PoC for the target dataset in terms of $\delta_X$. Even though we may not know the true value of $\delta_X$, we think of it like a sensitivity parameter that can be varied to understand how the bounds change. Ranges on $\delta_X$ can be imposed based on domain knowledge, or by utilizing information about the study design of the external dataset. Additionally, the prevalence parameter $\delta_X$ makes the assumptions on the treatment assignment mechanism of $X$ across datasets explicit, providing greater transparency in inference.

\textbf{Choice of $Y$} In the bounds we present, the prevalence of $Y$ is assumed to remained unchanged. As mentioned before, examples of such $Y$ include genetic mutations; according to Mendelian Randomization genes are randomized by nature, hence their prevalence is expected to be similar across populations with similar characteristics. This illustrates the useful role genetic mutations can play in tightening bounds on the PoC, and provides a way to leverage the growing number of genomics datasets. 
However, to relax the assumption on the unchanged prevalence of $Y$, a similar approach to the one used for $\delta_X$ could be employed to parameterize the difference in the treatment assignment mechanism of $Y$, and subsequently re-derive the bounds in this paper. We leave this to future work. 

\textbf{Invariance of Causal Mechanisms} We assume that $P(Z_{x, y, c})$ remains unchanged across datasets. This assumption is supported in the transportability, robustness and distribution shift literature \cite{pearl2011transportability, christiansen-generalization, buhlmann2020invariance}. However, if someone wishes to, following a similar approach used with $\delta_X$, this assumption can be further weakened, albeit at the cost of transferring less information across datasets.

\textbf{Data access and privacy} In the case of trials, we may not have access to individual-level records due to privacy or intellectual property concerns. The merit of our approach is that it only requires population-level summaries of the data, such as the adverse effect prevalence in the treated and untreated group, or these quantities within strata of the subject population. 

\textbf{Conclusion}
In this paper, we presented approaches to tighten the bounds on the Probabilities of Causation via merging external datasets studying the same outcome variable, but examining different treatments or covariates. To this end, we tightened existing bounds on the Probabilities of Causation by merging external datasets with the target dataset, allowing the external dataset to have a different treatment assignment mechanism. This is accomplished by parameterizing the difference in treatment mechanisms and providing bounds in terms of this parameter. Our approach could also be extended to derive bounds on counterfactual statements (rung 3 in Pearl’s ladder of causation) other than Probabilities of Causation.
\bibliography{aaai24}

\appendix

\maketitle
\onecolumn
\begin{center}
    {\LARGE\textbf{Appendix for Tightening Bounds on Probabilities of Causation By Merging Datasets}}
\end{center}
    
\section{Proof of Theorem 1}
To prove Theorem 1, we first state lemmas we utilize as part of our proof. The lemmas are organized in subsections as follows. First, Lemma 1 and 2 are needed to constrain the set of target distributions using the external dataset (\S \ref{subsec:constraint_set_target_distr}) and to express the set of target distributions compatible with the external distribution as a system of equations and its $N$ corresponding free parameters, where $N + 1$ is the cardinality of the support of $Y$. Furthermore, they place bounds on the free parameters in this system. Lemma 3 and 4 examine the conditions that govern the bounds on the free parameters. Next, Lemma 5 and 6 re-express the bounds on $PNS$ originally derived using the joint distributions $P(Z, X, Y)$ in terms of the free parameters derived in Lemma 1 and Lemma 2. Lemmas 7 through 20 examine the behavior of the maximum  operators in the bounds provided in Lemma 5 and 6 under various constraints on the free parameters. Finally, we use these lemmas to provide the proof for the lower bound and upper bound of the PNS. The proof for the lemmas can be found in (\S \ref{sec:lemma-proofs}). 

\subsection{Constraining the Set of Target Distributions Using the External Dataset}\label{subsec:constraint_set_target_distr}

\begin{lemma}\label{lemma:RREF-y}
Let $X$ and $Z$ be binary random variables, and let $Y$ be a discrete random variable taking on values in $0, \dots ,N$. Let $X$, $Y$ and $Z$ obey the causal structure in Fig. 3(b). Then, given marginals $P^T(Z, X)$ and $P^E(Z, Y)$, the conditional distributions $P(Z \mid X, Y)$ can be expressed in terms of free parameters $P(Z = 1 \mid X = 1, Y = i)$ for $i \in \{1, \dots, N\}$ as

{\small
\begin{align*}
\mathbb{P}(Z = 1 \mid X = 0, Y = 0) &=  \frac{p_x}{1 - p_x}\sum_{n = 1}^N\frac{p_{y_n}}{p_{y_0}}\mathbb{P}(Z = 1 \mid X = 1, Y = n) + \frac{p^E_{10}p_{y_0} - p^T_{11}p_x}{(1 - p_x)p_{y_0}}\\
\mathbb{P}(Z = 1 \mid X = 1, Y = 0)&= \frac{p^T_{11}}{p_{y_0}} - \sum_{n = 1}^N\frac{p_{y_n}}{p_{y_0}}\mathbb{P}(Z = 1 \mid X = 1, Y = n)\\
\mathbb{P}(Z = 1 \mid X = 0, Y = 1)&= \frac{p^E_{11}}{1 - p_x} - \frac{p_x}{1 - p_x}\mathbb{P}(Z = 1 \mid X = 1, Y = 1)\\
\mathbb{P}(Z = 1 \mid X = 0, Y = 2)&= \frac{p^E_{12}}{1 - p_x} - \frac{p_x}{1 - p_x}\mathbb{P}(Z = 1 \mid X = 1, Y = 2)\\
\vdots\\
\mathbb{P}(Z = 1 \mid X = 0, Y = N)&= \frac{p^E_{1N}}{1 - p_x} - \frac{p_x}{1 - p_x}\mathbb{P}(Z = 1 \mid X = 1, Y = N)\\
\end{align*}
}
Where $P(X = 1)$ is denoted as $p_x$, $P(Y = n)$ is denoted as $p_{y_n}$, $P^T(Z = 1 \mid X = 1)$ is denoted as $p^T_{11}$, $P^T(Z = 1 \mid X = 0)$ is denoted as $p^T_{10}$, and $P^E(Z = 1 \mid Y = n)$ is denoted as $p^E_{1n}$ for $n \in \{0, \dots, N\}$.
\end{lemma}

\begin{lemma}\label{lem:free-param-bounds}
To ensure the solution to the system of equations in Lemma \ref{lemma:RREF-y} form coherent probabilities, the following bounds on the free parameters must hold
{\small
\begin{align}\label{eq:sum-bounds}
\max\begin{bmatrix}
   \frac{p^T_{11}p_x - p^E_{10}p_{y_0}}{p_x} \\
   p_{11} - p_{y_0}
\end{bmatrix} \leq \sum_{n = 1}^N p_{y_n}\mathbb{P}(Z = 1 \mid X = 1, Y = n) 
\leq \min\begin{bmatrix}
    \frac{(1 - p_x)p_{y_0} + p^T_{11}p_x - p^E_{10}p_{y_0}}{p_x}\\
    p^T_{11}
\end{bmatrix}
\end{align}
}

And for each $i \in \{1, \dots, N\}$, we the bounds on the following free parameters $P(Z = 1\mid X = 1, Y = i)$
\begin{align}\label{eq:individual-bounds}
    \max\begin{bmatrix}
        0\\
        \frac{p^E_{1i} - (1 - p_x)}{p_x}
    \end{bmatrix}
        \leq P(Z = 1\mid X = 1, Y = i) \leq 
    \min\begin{bmatrix}
        1\\
        \frac{p^E_{1i}}{p_x}
    \end{bmatrix}
\end{align}
\end{lemma}

\subsection{Relation of Bounds on Free Parameters to the External Dataset}

Lemma \ref{lem:bounds-conditions} and Lemma \ref{lem:inidiv-bounds-conditions} establish the relationship between the external dataset and the upper and lower bounds in Lemma \ref{lem:free-param-bounds}.
\begin{lemma}\label{lem:bounds-conditions}
    The following implications hold:
    \begin{enumerate}
        \item 
        \begin{equation*}
        p^E_{10} < p_x \implies \frac{p^T_{11}p_x - p^E_{10}p_{y_0}}{p_x} > p^T_{11} - p_{y_0}   
        \end{equation*}
    
        \item 
        \begin{equation*}
        p^E_{10} \geq p_x \implies \frac{p^T_{11}p_x - p^E_{10}p_{y_0}}{p_x} \leq p^T_{11} - p_{y_0}   
        \end{equation*}
        \item 
        \begin{equation*}
        p^E_{10} < 1 - p_x \implies p^T_{11} < \frac{(1 - p_x)p_{y_0} + p^T_{11}p_x - p^E_{10}p_{y_0}}{p_x}
        \end{equation*}
        \item 
        \begin{equation*}
        p^E_{10} \geq 1 - p_x \implies p^T_{11} \geq \frac{(1 - p_x)p_{y_0} + p^T_{11}p_x - p^E_{10}p_{y_0}}{p_x}
        \end{equation*}
\end{enumerate}
\end{lemma}
\begin{lemma}\label{lem:inidiv-bounds-conditions}
For $i \in \{1, \dots, N\}$, when $p^E_{1i} \geq 1 - p_x \implies \frac{p^E_{11} - (1 - p_x)}{p_x} \geq 0$. When $p^E_{1i} < p_x \implies \frac{p^E_{11}}{p_x} \leq 1$    
\end{lemma}

\subsection{Bounds on PNS In Terms of Free Parameters}

In Lemma \ref{lem:joint-bounds}, we re-express bounds in \citet{dawid2017probability} on PC in terms of PNS under the assumption of exogenity. 
\begin{lemma}\label{lem:joint-bounds}
Let $X$, $Z$ be binary random variables, and let $Y$ be a discrete multivalued random variable, obeying the causal structure in Fig 2. We define the $PNS$ as
\begin{equation*}
    PNS \equiv P(Z_{\mathbf{x}} = 1, Z_{\mathbf{x^\prime}}  = 0)
\end{equation*}
Where $\mathbf{x}$ and $\mathbf{x^\prime}$ denote the setting of $X$ to $1$ and $0$ respectively. Bounds on the PNS under the causal graph in Fig. \ref{fig:single-rct-with-covariates} given the joint distribution $P(Z, X, Y)$ are given in \citet{dawid2017probability} as 
\begin{align*}
    \Delta \leq PNS \leq \mathbb{P}(Z = 1 \mid X = 1) - \Gamma
\end{align*}

Where
\begin{equation*}
    \Delta = \sum_y \mathbb{P}(Y = y)\max \{0, \mathbb{P}(Z = 1 \mid X = 1, Y = y) - \mathbb{P}(Z = 1 \mid X = 0, Y = y)\} 
\end{equation*}

And $\Gamma$ is 
\begin{align*}
    \Gamma = \sum_y \mathbb{P}(Y = y)\max \{0, \mathbb{P}(Z = 1 \mid X = 1, Y = y) - \mathbb{P}(Z = 0 \mid X = 0, Y = y)\} 
\end{align*}
\end{lemma}

At this point we remind to the reader that the above bounds require access to the joint distribution $P(Z, X, Y)$. Given the scenario described in the paper, we do not have access to this. Therefore, in Lemma \ref{lem:max-operator-expressions} we are expressing the joint $P(Z\mid  X, Y)$ in terms of the free parameters corresponding to the system of equations derived in Lemma \ref{lemma:RREF-y}. 
\begin{lemma}\label{lem:max-operator-expressions}
    Given $P^T(Z, X)$ and $P^E(Z, Y)$, using the solution to the system of equations in Lemma \ref{lemma:RREF-y}, each of the terms in $\Delta$ and $\Gamma$ defined in Lemma \ref{lem:joint-bounds} are expressed in terms of the free parameters below. First, for $\Gamma$, each of the terms can be written as
    \[
    \mathbb{P}(Z = 1 \mid X = 1, Y = 0) - \mathbb{P}(Z = 0 \mid X = 0, Y = 0)
    \]
    \begin{equation}\label{eq:max-operator-free-parameter-0}
    = \frac{p^E_{10} - (1 - p_x)}{1 - p_x} + \frac{\sum_{n = 1}^N p_{y_n}P(Z = 1 \mid X = 1, Y = n) - p^T_{11}}{p_{y_0}}\Bigg(\frac{p_x}{1 - p_x} - 1\Bigg)
    \end{equation}
    And for the terms in $\Gamma$ corresponding to $Y = i$, $i \in \{1, \dots , N\}$
    \[
    P(Z = 1 \mid X = 1, Y = i) - P(Z = 0 \mid X = 0, Y = i)
    \]
    \begin{equation}\label{eq:max-operator-free-parameter-i}
        P(Z = 1 \mid X = 1, Y = i)\Bigg(1 - \frac{p_x}{1 - p_x}\Bigg) + \frac{p^E_{11} - (1 - p_x)}{1 - p_x}
    \end{equation}
    Similarly, the terms in $\Delta$ can be expressed in terms of the free parameter as
    \[
    \mathbb{P}(Z = 1 \mid X = 1, Y = 0) - \mathbb{P}(Z = 1 \mid X = 0, Y = 0)
    \]
    \begin{equation}
       \frac{p^T_{11} - p^E_{10}p_{y_0} - \sum_{n = 1}^N p_{y_n}\mathbb{P}(Z = 1\mid X = 1, Y = n)}{(1 - p_x)p_{y_0}} 
    \end{equation}
    And for the terms in $\Delta$ corresponding to $Y = i$, $i \in \{1, \dots , N\}$
    \[
    \mathbb{P}(Z = 1 \mid X = 1, Y = 1) - \mathbb{P}(Z = 1 \mid X = 0, Y = 1)
    \]
    \begin{equation}
        \frac{\mathbb{P}(Z = 1 \mid X = 1, Y = i) - p^E_{1i}}{1 - p_x}
    \end{equation}
\end{lemma}

\subsection{Behavior of Max Operators in $\Gamma$}
The behavior of this function must be analyzed in two cases, one where $p_x > 1 - p_x$, and vice-versa. We present results for $p_x > 1 - p_x$, and analogous results can be derived for $p_x < 1 - p_x$. 

\begin{lemma}\label{lem:max-operator-gamma-0}
    Assume $p_x > 1 - p_x$. The function $\max \{0, \mathbb{P}(Z = 1 \mid X = 1, Y = 0) - \mathbb{P}(Z = 0 \mid X = 0, Y = 0)\}$ defined in Lemma \ref{lem:max-operator-expressions} will be greater than or equal to $0$ when
    \begin{equation}
            \sum_{n = 1}^N p_{y_n}P(Z = 1 \mid X = 1, Y = n) \geq \frac{p_{y_0}((1 - p_x) - p^E_{10})}{p_x - (1 - p_x)} + p^T_{11}
    \end{equation}
\end{lemma}

\begin{lemma}\label{lem:max-operator-gamma-0-positive}
    Assume $p_x > 1 - p_x$. If $p^E_{10} \geq p_x$, then 
    \[
    p^T_{11} - p_{y_0} \geq \frac{p_{y_0}((1 - p_x) - p^E_{10})}{p_x - (1 - p_x)} + p^T_{11}
    \]
\end{lemma}

\begin{remark}
    Lemma \ref{lem:max-operator-gamma-0} and Lemma \ref{lem:max-operator-gamma-0-positive} imply that when $p_x > 1 - p_x$, $p^E_{10} \geq \max\{p_x, 1 - p_x\}$ then 
\begin{align*}
    &\max\{0, \mathbb{P}(Z = 1 \mid X = 1, Y = 0) - \mathbb{P}(Z = 0 \mid X = 0, Y = 0)\}\\
    &= \mathbb{P}(Z = 1 \mid X = 1, Y = 0) - \mathbb{P}(Z = 0 \mid X = 0, Y = 0)
\end{align*}
\end{remark}
Analogously, when $p^E_{10} \leq \min\{p_x, 1 - p_x\}$, i.e. $p^E_{10} \leq 1 - p_x$, similar results are obtained, and to show this, we state the following Lemmas.
\begin{lemma}\label{lem:gamma-0-lower-side-px}
    Assume $p_x > 1 - p_x$. If $p^E_{10} < p_x$, then 
    \[
    \frac{p^T_{11}p_x - p^E_{10}p_{y_0}}{p_x} < \frac{p_{y_0}((1 - p_x) - p^E_{10})}{p_x - (1 - p_x)} + p^T_{11}
    \]
\end{lemma}

\begin{lemma}\label{lem:gamma-0-lower-side-1-minus-px}
    Assume $p_x > 1 - p_x$. If $p^E_{10} \leq 1 - p_x$, then 
    \[
    p^T_{11} \leq \frac{p_{y_0}((1 - p_x) - p^E_{10})}{p_x - (1 - p_x)} + p^T_{11}
    \]
\end{lemma}
\begin{remark}
Lemma \ref{lem:gamma-0-lower-side-px} and Lemma \ref{lem:gamma-0-lower-side-1-minus-px} imply that when $p_x > 1 - p_x$, $p^E_{10} \leq \min\{p_x, 1 - p_x\}$ then 
\begin{align*}
    &\max\{0, \mathbb{P}(Z = 1 \mid X = 1, Y = 0) - \mathbb{P}(Z = 0 \mid X = 0, Y = 0)\} = 0
\end{align*}    
\end{remark}

\begin{lemma}\label{lem:gamma-0-middle}
    Assume $p_x > 1 - p_x$. When $1 - p_x < p^E_{10}$, the following inequality holds.
    \[
    \frac{p_{y_0}((1 - p_x) - p^E_{10})}{p_x - (1 - p_x)} + p^T_{11} < \frac{(1 - p_x)p_{y_0} + p^T_{11}p_x - p^E_{10}p_{y_0}}{p_x}    
    \]
\end{lemma}

Now, we provide lemmas that similarly examine the behavior of $\max \{0, \mathbb{P}(Z = 1 \mid X = 1, Y = i) - \mathbb{P}(Z = 0 \mid X = 0, Y = i)\}$ for $i \in \{1, \dots, N\}$. As before, the behavior of this function needs to be analyzed in two cases, $p_x > 1 - p_x$ and one where $p_x < 1 - p_x$. We provide lemmas for $p_x > 1 - p_x$, and lemmas for $p_x < 1 - p_x$ can be analogously derived.

\begin{lemma}\label{lem:max-operator-gamma-i}
    Assume $p_x > 1 - p_x$. The function $\max \{0, \mathbb{P}(Z = 1 \mid X = 1, Y = i) - \mathbb{P}(Z = 0 \mid X = 0, Y = i)\} \geq 0$ for $i \in \{1, \dots, N\}$ when
    \begin{align*}
            P(Z = 1 \mid X = 1, Y = i) \leq \frac{p^E_{1i} - (1 - p_x)}{p_x - (1 - p_x)}    
    \end{align*}
\end{lemma}

\begin{lemma}\label{lem:max-operator-gamma-i-positive}
    Assume $p_x > 1 - p_x$. If $p^E_{1i} \geq p_x$ for $i \in \{1, \dots , N\}$, then 
    \begin{align*}
            1 \leq \frac{p^E_{1i} - (1 - p_x)}{p_x - (1 - p_x)}    
    \end{align*}
\end{lemma}

\begin{remark}
    Lemma \ref{lem:max-operator-gamma-i} and Lemma \ref{lem:max-operator-gamma-i-positive} imply that when $p^E_{1i} \geq \max \{p_x, 1 - p_x\}$, then
    \[
    \max \{0, \mathbb{P}(Z = 1 \mid X = 1, Y = i) - \mathbb{P}(Z = 0 \mid X = 0, Y = i)\}
        = \mathbb{P}(Z = 1 \mid X = 1, Y = i) - \mathbb{P}(Z = 0 \mid X = 0, Y = i)
    \]
\end{remark} 

\begin{lemma}\label{lem:max-operator-gamma-i-0-1}
    Assume $p_x > 1 - p_x$. If $p^E_{1i} < p_x$ for $i \in \{1, \dots , N\}$, then 
    \begin{align*}
            \frac{p^E_{11}}{p_x} > \frac{p^E_{1i} - (1 - p_x)}{p_x - (1 - p_x)}    
    \end{align*}
\end{lemma}
\begin{lemma}\label{lem:max-operator-gamma-i-0-2}
    Assume $p_x > 1 - p_x$. If $p^E_{1i} \leq 1 - p_x$ for $i \in \{1, \dots , N\}$, then 
    \begin{align*}
            0 \geq \frac{p^E_{1i} - (1 - p_x)}{p_x - (1 - p_x)}    
    \end{align*}
\end{lemma}
\begin{remark}
    Lemma \ref{lem:max-operator-gamma-i-0-1} and Lemma \ref{lem:max-operator-gamma-i-0-2} imply that when $p^E_{1i} \leq \min \{p_x, 1 - p_x\}$, then
    \begin{align*}
    \max \{0, P(Z = 1 \mid X = 1, Y = i) - P(Z = 0 \mid X = 0, Y = i)\} = 0
    \end{align*}
\end{remark}

And when $1 - p_x < p^E_{1i} < p_x$, then $\frac{p^E_{11}}{p_x} > \frac{p^E_{1i} - (1 - p_x)}{p_x - (1 - p_x)}$ and $\frac{p^E_{11} - (1 - p_x)}{p_x} < \frac{p^E_{1i} - (1 - p_x)}{p_x - (1 - p_x)}$. 

\begin{lemma}\label{lem:max-operarot-simultaneous-open}
Assume $p_x > 1 - p_x$, and for all $i \in \{0, \dots, N\}$, $1 - p_x < p^E_{1i} < p_x$. Then the space of the free parameters defined in Lemma \ref{lemma:RREF-y} will contain a value of the free parameters such that every max operator corresponding in $\Gamma$ defined in Lemma \ref{lem:joint-bounds} will simultaneously satisfy the following condition
\begin{equation}\label{eq:max-operator-opening-condition}
\max\{0, P(Z = 1 \mid X = 1, Y = i) - P(Z = 0 \mid X = 0, Y = i)\} = P(Z = 1 \mid X = 1, Y = i) - P(Z = 0 \mid X = 0, Y = i)    
\end{equation}
for all $i$ if and only if $p^T_{11} + p^T_{10} - 1 \geq 0$. 
\end{lemma}

\subsection{Behavior of Max Operators in $\Delta$}

\begin{lemma}\label{lem:delta-max-operator-0-non-0}
    The function $\max \{0, P(Z = 1 \mid X = 1, Y = 0) - P(Z = 1 \mid X = 0, Y = 0)\} \geq 0$ when
    \begin{align*}
        p^T_{11} - p^E_{10}p_{y_0} \geq  \sum_{n = 1}^N p_{y_n}\mathbb{P}(Z = 1\mid X = 1, Y = n)
    \end{align*}
\end{lemma}

\begin{lemma}\label{lem:delta-0-max-operator-bounds}
    The following inequalities hold
    \begin{enumerate}
        \item 
        \begin{equation}
            p^T_{11} \geq p^T_{11} - p^E_{10}p_{y_0}
        \end{equation}

    \item 
    \begin{equation}
        \frac{(1 - p_x)p_{y_0} + p^T_{11}p_x - p^E_{10}p_{y_0}}{p_x} \geq p^T_{11} - p^E_{10}p_{y_0}
    \end{equation}

    \item 
    \begin{equation}
        p^T_{11} - p_{y_0} \leq p^T_{11} - p^E_{10}y_0
    \end{equation}
    
    \item 
    \begin{equation}
        \frac{p^T_{11}p_x - p^E_{10}p_{y_0}}{p_x} \leq p^T_{11} - p^E_{10}p_{y_0}
    \end{equation}    
    \end{enumerate}
\end{lemma}

\begin{remark}
Lemma \ref{lem:delta-0-max-operator-bounds} implies that  $\max\{0, \mathbb{P}(Z = 1 \mid X = 1, Y = 0) - \mathbb{P}(Z = 1 \mid X = 0, Y = 0)\}$ can be either zero or non-zero over the range of the free parameters. 
\end{remark}

Now, we similarly examine the behavior of $\max \{0, \mathbb{P}(Z = 1 \mid X = 1, Y = i) - \mathbb{P}(Z = 1 \mid X = 0, Y = i)\}$.

\begin{lemma}\label{lem:delta-max-operator-i-non-0}
    The function $\max \{0, \mathbb{P}(Z = 1 \mid X = 1, Y = i) - \mathbb{P}(Z = 1 \mid X = 0, Y = i)\} \geq 0$ for $i \in \{1, \dots, N\}$ when
    \begin{equation*}
        p^E_{1i} \leq P(Z = 1 \mid X = 1, Y = 1)
    \end{equation*}
\end{lemma}

\begin{lemma}\label{lem:delta-max-operator-i-non-0-bounds-test}
    The following inequalities hold
    \begin{enumerate}
        \item 
        \begin{equation}
            1 \geq p^E_{1i}
        \end{equation}
        
        \item 
        \begin{equation}
            \frac{p^E_{1i}}{p_x} \geq p^E_{1i}
        \end{equation}
        
        \item 
        \begin{equation}
            p^E_{1i} \geq 0
        \end{equation}

        \item 
        \begin{equation}
            p^E_{1i} \geq \frac{p^E_{1i} - (1 - p_x)}{p_x}
        \end{equation}
    \end{enumerate}
\end{lemma}

\begin{remark}
Lemma \ref{lem:delta-max-operator-i-non-0} and Lemma \ref{lem:delta-max-operator-i-non-0-bounds-test} imply that $\max\{0, P(Z = 1 | X = 1, Y = i) - P(Z = 1 | X = 0, Y = i)\}$ can be either zero or non-zero over the range of the free parameters. 
\end{remark}

\begin{lemma}\label{lem:max-operator-simul-open-delta}
All max operators in $\Delta$ can simultaneously satisfy the following condition for all $i \in \{0, \dots, N\}$ if and only if
    \begin{equation}\label{eq:delta-simul-condition}
    \max\{0, P(Z = 1 \mid X = 1, Y = i\} - P(Z = 1\mid X = 0, Y = i)\} = P(Z = 1 \mid X = 1, Y = i\} - P(Z = 1\mid X = 0, Y = i)
    \end{equation}
if and only if $p^T_{11} - p^T_{10} \geq 0$. 
\end{lemma}
\subsection{Behavior of Upper Bound on PNS Under Special Conditions}

Here we analyze the behavior of the upper bound on the PNS in Theorem 1 and Theorem 2 when $p^E_{1i}$ is always greater than $\max\{p_x, 1 - p_x\}$ or always less than $\min\{p_x, 1 - p_x\}$.

\begin{lemma}\label{lem:pns-all-right-extereme}
    For the upper bound in Theorem 2, when $p_x > 1 - p_x$ and all $p^E_{1i} \geq p_x$ for $i \in \{0, \dots, N\}$, then 
    \[
    p^T_{00} < p^T_{11} - \sum_{i = 0}^N \Phi_i
    \]
\end{lemma}

\begin{proof}
    Since $p^E_{1i} \geq p_x$ for all $i$, then all the indicators in $\Phi_i$ will evaluate to $1$, as a result $p^T_{11} - \sum_{i = 0}^N \Phi_i$ will equal
    \begin{align*}
        &p^T_{11} - \sum_{i = 0}^N \frac{p_{y_i}(p^E_{1i} - p_x)}{1 - p_x}\\
        &\implies = \frac{p^T_{11}(1 - p_x) - \sum_{i = 0}^N p_{y_i}(p^E_{1i} - p_x)}{1 - p_x}\\
        &\implies = \frac{p^T_{11}(1 - p_x) - p^T_{11}p_x - p^T_{10}(1 - p_x) + p_x}{1 - p_x}\\
        &\implies = \frac{p^T_{11}(1 - p_x) - p^T_{11}p_x + p^T_{00}(1 - p_x) - (1 - p_x) + p_x}{1 - p_x}\\
        &\implies = p^T_{01}(\frac{p_x}{1 - p_x} - 1) + p^T_{00}\\
    \end{align*}
    And since $p_x > 1 - p_x$, this quantity will be greater than $p^T_{00}$.
\end{proof}

\begin{lemma}\label{lem:pns-all-left-extereme}
    For the upper bound in Theorem 2, when $p_x > 1 - p_x$ and $p^E_{1i} \leq 1 - p_x$ for $i \in \{0, \dots, N\}$ and , then 
    \[
    p^T_{00} - \sum_{i = 0}^N \Theta_i> p^T_{11}
    \]
\end{lemma}

\begin{proof}
Since $p^E_{1i} \leq 1 - p_x$ for all $i$, then all the indicators in $\Theta_i$ will evaluate to $1$, as a result $p^T_{00} - \sum_{i = 0}^N \Theta_i$ will equal
\begin{align*}
    &p^T_{00} - \sum_{i = 0}^N \frac{p_{y_i}((1 - p_x) - p^E_{1i})}{1 - p_x}\\
    &\implies = \frac{p^T_{00}(1 - p_x) - (1 - p_x) + \sum_{i = 0}^N p_{y_i}p^E_{1i}}{1 - p_x}\\
    &\implies = \frac{-p^T_{10}(1 - p_x) + \sum_{i = 0}^N p_{y_i}p^E_{1i}}{1 - p_x}\\
    &\implies = \frac{p^T_{11}p_x}{1 - p_x}
\end{align*}
And since $p_x > 1 - p_x$, this concludes the proof.
\end{proof}

Similar Lemmas can be provided for the case where $p_x < 1 - p_x$ as well. 

\subsection{Proof Of Upper Bound In Theorem 1}

The upper bound in Theorem 1 is given as
\begin{equation}\label{eq:thm-1-bound}
    PNS \leq \min\begin{bmatrix}
        p_{11} - \sum_{i = 0}^1\mathbb{I}(p^\prime_{1i} \geq \max\{1 - p_x, p_x\})\frac{p_{y_i}(p^\prime_{1i} - \max\{1 - p_x, p_x\})}{\min\{1 - p_x, p_x\}}\\
        p_{00} - \sum_{i = 0}^1\mathbb{I}(p^\prime_{1i} \leq \min\{1 - p_x, p_x\})\frac{p_{y_i}(\min\{1 - p_x, p_x\} - p^\prime_{1i})}{\min\{1 - p_x, p_x\}}\\
    \end{bmatrix}
\end{equation}
The upper bound is obtained by solving
\begin{equation}\label{eq:david-bound-link}
    PNS \leq  \mathbb{P}(Z = 1 \mid X = 1) - \min_{P_i} \Gamma(P_i)
\end{equation}
We follow a proof by cases approach, and demonstrate on a case-by-case basis that bounds obtained Eq. \ref{eq:thm-1-bound} will equal the bounds obtained from Eq.\ref{eq:david-bound-link}. 

First, since $Y$ is binary, we can use Lemma \ref{lemma:RREF-y} and Lemma \ref{lem:free-param-bounds} to obtain the following system of equations that constrain the set of possible choices for $P(Z = 1| X, Y)$ with respect to a single free parameter $P(Z = 1 \mid X = 1, Y = 1)$.

\begin{align*}
P(Z = 1 \mid X = 0, Y = 0) &=  \frac{p_x}{1 - p_x}\frac{p_{y_1}}{p_{y_0}}P(Z = 1 \mid X = 1, Y = 1) + \frac{p^E_{10}p_{y_0} - p^T_{11}p_x}{(1 - p_x)p_{y_0}}\\
P(Z = 1 \mid X = 1, Y = 0)&= \frac{p^T_{11}}{p_{y_0}} - \frac{p_{y_1}}{p_{y_0}}P(Z = 1 \mid X = 1, Y = 1)\\
P(Z = 1 \mid X = 0, Y = 1)&= \frac{p^E_{11}}{1 - p_x} - \frac{p_x}{1 - p_x}P(Z = 1 \mid X = 1, Y = 1)
\end{align*}

And the following bounds on the free parameter $P(Z = 1 \mid X = 1, Y = 1)$ must hold

\begin{equation}\label{eq:sum-bounds-thm1}
\max\begin{bmatrix}
   \frac{p^T_{11}p_x - p^E_{10}p_{y_0}}{p_x} \\
   p^T_{11} - p_{y_0}
\end{bmatrix} \leq p_{y_1}P(Z = 1 \mid X = 1, Y = 1) 
\leq \min\begin{bmatrix}
    \frac{(1 - p_x)p_{y_0} + p_{11}p_x - p^\prime_{10}p_{y_0}}{p_x}\\
    p_{11}
\end{bmatrix}
\end{equation}
Along with
\begin{equation}\label{eq:indiv-bounds-thm1}
    \max\begin{bmatrix}
        0\\
        \frac{p^\prime_{11} - (1 - p_x)}{p_x}
    \end{bmatrix}
        \leq P(Z = 1\mid X = 1, Y = 1) \leq 
    \min\begin{bmatrix}
        1\\
        \frac{p^\prime_{11}}{p_x}
    \end{bmatrix}
\end{equation}
In the binary case, from Lemma \ref{lem:joint-bounds}$, \Gamma$ is expressed as
\begin{align*}
    \Gamma &= P(Y = 0)\max \{0, P(Z = 1 \mid X = 1, Y = 0) - P(Z = 0 \mid X = 0, Y = 0)\}\\
    &+ P(Y = 1)\max \{0, P(Z = 1 \mid X = 1, Y = 1) - P(Z = 0 \mid X = 0, Y = 1)\} 
\end{align*}

And based on Lemma \ref{lem:max-operator-expressions}, each of these terms in $\Gamma$ can be written in terms of the free parameter $P(Z = 1 \mid X = 1, Y = 1)$ as
\begin{enumerate}
    \item 
    \[P(Z = 1 \mid X = 1, Y = 0) - P(Z = 0 \mid X = 0, Y = 0)\]
    \begin{equation}\label{eq:binary-gamma-y0-thm1}
        \frac{p^E_{10} - (1 - p_x)}{1 - p_x} + \frac{p_{y_1}P(Z = 1 \mid X = 1, Y = 1) - p^T_{11}}{p_{y_0}}\Bigg(\frac{p_x}{1 - p_x} - 1\Bigg) 
    \end{equation}
    \item 
    \[
    P(Z = 1 \mid X = 1, Y = 1) - P(Z = 0 \mid X = 0, Y = 1)
    \]
    \begin{equation}\label{eq:binary-gamma-y1-thm1}
        P(Z = 1 \mid X = 1, Y = 1)(1 - \frac{p_x}{1 - p_x}) + \frac{p^E_{11} - (1 - p_x)}{1 - p_x}\\
    \end{equation}
\end{enumerate}

For ease of presentation, we group the cases we consider in our proof into three broad categories. The first category groups cases where $p_x = 1 - p_x$, the second groups cases where $p_x > 1 - p _x$ and the third groups cases where $p_x < 1 - p_x$. We provide proofs for the first two, and analogous derivations can be performed for $p_x < 1 - p_x$.\\
\newline
\textbf{Category I} In Category I, we provide proofs for when $p_x = 1 - p_x$. 
In this setting, we will prove the following equations hold: 
{\small
\begin{equation}\label{eq:upper-bound-1-thm1}
p^T_{11} - \min_{P(Z = 1 \mid X = 1, Y = 1)} \Gamma(P(Z = 1 \mid X = 1, Y = 1)) = p^T_{11} - \sum_{i = 0}^1\mathbb{I}(p^E_{1i} \geq \max\{1 - p_x, p_x\})\frac{p_{y_i}(p^E_{1i} - \max\{1 - p_x, p_x\})}{\min\{1 - p_x, p_x\}}    
\end{equation}
}
And
{\small
\begin{equation}\label{eq:upper-bound-2-thm1}
p^T_{11} - \min_{P(Z = 1 \mid X = 1, Y = 1)} \Gamma(P(Z = 1 \mid X = 1, Y = 1)) = p^T_{00} - \sum_{i = 0}^N\mathbb{I}(p^E_{1i} \leq \min\{1 - p_x, p_x\})\frac{p_{y_i}(\min\{1 - p_x, p_x\} - p^E_{1i})}{\min\{1 - p_x, p_x\}}    
\end{equation}
}
Note that when $p_x = 1 - p_x$, $\Gamma$ is no longer a function of $P(Z = 1 \mid X = 1, Y = 1)$ since \ref{eq:binary-gamma-y0-thm1} and \ref{eq:binary-gamma-y1-thm1} are no longer a function of $P(Z = 1 \mid X = 1, Y = 1)$. $\Gamma$ will equal
\begin{align*}
    \Gamma = p_{y_0}\max \{0, \frac{p^E_{10} - (1 - p_x)}{1 - p_x} \}
    + p_{y_1}\max \{0,  \frac{p^E_{11} - (1 - p_x)}{1 - p_x}\} 
\end{align*}
And the upper bound on PNS will equal
\begin{align*}
    p^T_{11} - \Gamma
\end{align*}
Next, we can see on a case by case basis that $p^T_{11} - \Gamma$ will always equal both \ref{eq:upper-bound-1-thm1} and \ref{eq:upper-bound-2-thm1}. We provide the proof for one case, and the rest of the cases can be proved in a similar fashion.\\
\begin{enumerate}
    \item Consider the case where $p^\prime_{10} \geq p_x$ and $p^\prime_{11} \geq p_x$. Since $p_x = 1 - p_x$, $\Gamma$ will equal
        \begin{align*}
            \Gamma &= \frac{p_{y_0}p^E_{10} - p_{y_0}(1 - p_x)}{1 - p_x} + \frac{p_{y_1}p^E_{11} - p_{y_1}(1 - p_x)}{1 - p_x}\\
            \implies p^T_{11} - \Gamma &= p^T_{11} - \frac{p_{y_0}p^E_{10} - p_{y_0}(1 - p_x)}{1 - p_x} - \frac{p_{y_1}p^E_{11} - p_{y_1}(1 - p_x)}{1 - p_x}
        \end{align*}
    By direct comparison, this is equal to \ref{eq:upper-bound-1-thm1}. Next, to prove the equality to \ref{eq:upper-bound-2-thm1}, note that
    \begin{align*}
        &p^T_{11} - \frac{p_{y_0}p^E_{10} - p_{y_0}(1 - p_x)}{1 - p_x} - \frac{p_{y_1}p^E_{11} - p_{y_1}(1 - p_x)}{1 - p_x}\\
        \intertext{Since $p_x = 1 - p_x$ and $p^T_{11}p_x + p^T_{10}(1 - p_x) = p^E_{11}p_{y_1} + p^E_{10}p_{y_0}$}
        &\implies = \frac{p^T_{11}p_x - p_{y_0}p^E_{10} + p_{y_0}(1 - p_x) - p_{y_1}p^E_{11} + p_{y_1}(1 - p_x)}{1 - p_x}\\
        &\implies = \frac{-p^T_{10}(1 - p_x) + (1 - p_x)}{1 - p_x}\\
        &\implies = p^T_{00}
    \end{align*}
    And by comparison to \ref{eq:upper-bound-2-thm1}, we can see these are equal as well.
\end{enumerate}

This concludes the proof for when $p_x = 1 - x$. Next, we move on to the category of cases where $p_x > 1 - p_x$.\\
\newline
\textbf{Category II} In Category II, we provide proofs for when $p_x > 1 - p_x$. Here we follow a similar proof by cases as well, with each case being defined by relation of $p^E_{10}$ and $p^E_{11}$ to $p_x$ and $1 - p_x$. 
Note that since $p_x > 1 - p_x$, from Lemma \ref{lem:max-operator-expressions} we see that $\max \{0, \mathbb{P}(Z = 1 \mid X = 1, Y = 0) - \mathbb{P}(Z = 0 \mid X = 0, Y = 0)\}$ is an increasing function of the free parameter, while $\max \{0, \mathbb{P}(Z = 1 \mid X = 1, Y = i) - \mathbb{P}(Z = 0 \mid X = 0, Y = i)\}$ is a decreasing function of the free parameter. Now, we proceed to use a proof by cases. We provide proofs for a set of representative cases, and a similar approach can be used to derive the rest of the cases. 
\begin{enumerate}
    \item Case I: We prove the upper bound in Theorem 1 in the following case
        \begin{align*}
            p_x &\leq p^E_{10}\\
            p_x &\leq p^E_{11}\\
        \end{align*}
        Since $p^E_{10} \geq \max\{p_x, 1 - p_x\}$, and $p^E_{11}$, Lemma \ref{lem:max-operator-gamma-0} and Lemma \ref{lem:max-operator-gamma-0-positive} imply 
        \begin{equation*}
        \max \{0, \mathbb{P}(Z = 1 \mid X = 1, Y = 0) - \mathbb{P}(Z = 0 \mid X = 0, Y = 0)\} = \mathbb{P}(Z = 1 \mid X = 1, Y = 0) - \mathbb{P}(Z = 0 \mid X = 0, Y = 0)    
        \end{equation*}
        And Lemma \ref{lem:max-operator-gamma-i-0-1} and Lemma \ref{lem:max-operator-gamma-i-0-2} imply
        \begin{equation*}
            \max \{0, \mathbb{P}(Z = 1 \mid X = 1, Y = 1) - \mathbb{P}(Z = 0 \mid X = 0, Y = 1)\}
            = \mathbb{P}(Z = 1 \mid X = 1, Y = 1) - \mathbb{P}(Z = 0 \mid X = 0, Y = 1)
        \end{equation*}
        And from Lemma \ref{lem:joint-bounds}, $\Gamma$ is equal to the weighted sum of these two max functions, hence
        \begin{align}\label{eq:weighted-sum-binary}
            \Gamma &= P(Y = 0)(P(Z = 1 \mid X = 1, Y = 0) - P(Z = 0 \mid X = 0, Y = 0)\\
            &+ P(Y = 1)(P(Z = 1 \mid X = 1, Y = 1) - P(Z = 0 \mid X = 0, Y = 1))\\
            \intertext{Since $X \independent Y$}
            &= P(Z = 1 \mid X = 1) + P(Z = 1 \mid X = 0) - 1
        \end{align}
        And in notation introduced before, this is written as $p^T_{11} + p_{10} - 1$. Since the upper bound on $PNS$ is $p^T_{11} - \Gamma$, this will evaluate to $p^T_{00}$. And based on Lemma \ref{lem:pns-all-right-extereme}, this will equal the upper bound result in Theorem 1.
    \item Case II: We prove the upper bound in Theorem 1 in the following case
            \begin{align*}
            p^E_{10} \leq 1 - p_x \\
            p^E_{11} \leq 1 - p_x \\
            \end{align*}
        Lemma \ref{lem:gamma-0-lower-side-px} and Lemma \ref{lem:gamma-0-lower-side-1-minus-px} along with Lemma \ref{lem:max-operator-gamma-i-0-1} and Lemma \ref{lem:max-operator-gamma-i-0-2} imply that both max operators will equal $0$ over the entire range of the free parameter. Hence $\Gamma = 0$, hence $p_{11} - \Gamma = p_{11}$, and from Lemma \ref{lem:pns-all-left-extereme} this matches the bound in Theorem 1.
    \item Case III: We prove the upper bound in Theorem 1 in the following case
            \begin{align*}
            p_x &\leq p^E_{10}\\
            p^E_{11} &\leq 1 - p_x\\
            \end{align*}
            Since $p^E_{11} \leq \min\{p_x, 1 - p_x\}$, Lemma \ref{lem:max-operator-gamma-i-0-1} and Lemma \ref{lem:max-operator-gamma-i-0-2} imply that $\max \{0, \mathbb{P}(Z = 1 \mid X = 1, Y = 1) - \mathbb{P}(Z = 0 \mid X = 0, Y = 1)\} = 0$ over the entire range of the free parameter. 
            
            Since $p^E_{10} \geq \max\{p_x, 1 - p_x\}$, and $p^E_{11}$, Lemma \ref{lem:max-operator-gamma-0} and Lemma \ref{lem:max-operator-gamma-0-positive} imply 
            \begin{equation*}
                \max \{0, \mathbb{P}(Z = 1 \mid X = 1, Y = 0) - \mathbb{P}(Z = 0 \mid X = 0, Y = 0)\} = \mathbb{P}(Z = 1 \mid X = 1, Y = 0) - \mathbb{P}(Z = 0 \mid X = 0, Y = 0)    
            \end{equation*} 
            Consequently, from Lemma \ref{lem:max-operator-expressions}, $\mathbb{P}(Z = 1 \mid X = 1, Y = 0) - \mathbb{P}(Z = 0 \mid X = 0, Y = 0)$ can be expressed in terms of the free parameter, and $\min \Gamma$ is obtained by solving
            \begin{equation}\label{eq:cat2-case-3-term-min}
                \min_{P(Z = 1 \mid X = 1, Y = 1)}  \frac{p^E_{10} - (1 - p_x)}{1 - p_x} + \frac{p_{y_1}P(Z = 1 \mid X = 1, Y = 1) - p^T_{11}}{p_{y_0}}\Bigg(\frac{p_x}{1 - p_x} - 1\Bigg) 
            \end{equation}
            This is an increasing function of the free parameter, and the minimum will be attained at the lower bound of the free parameter. Note that from Eq. \ref{eq:sum-bounds-thm1} and Eq. \ref{eq:indiv-bounds-thm1}, the lower bound on $P(Z = 1 \mid X = 1, Y = 1)$ will equal $\max\{0, p_{11} - p_{y_0}\}$. Substituting both of these lower bounds into Eq. \ref{eq:cat2-case-3-term-min} and noting that $\Gamma$ is a linearly increasing function of the free parameter gives
            \begin{align*}
                \Gamma = \max\{\frac{p_{y_0}(p^E_{10} - p_x)}{1 - p_x}, (p_{11} + p_{10} - 1) + \frac{p_{y_1}((1 - p_x) - p^E_{11})}{1 - p_x} \}
            \end{align*}        
            And since the upper bound is $p^T_{11} - \Gamma$, then 
            \begin{align*}
                &p^T_{11} - \max\{\frac{p_{y_0}(p^E_{10} - p_x)}{1 - p_x}, (p_{11} + p_{10} - 1) + \frac{p_{y_1}((1 - p_x) - p^E_{11})}{1 - p_x} \}\\
                &= \min\{p^T_{11} - \frac{p_{y_0}(p^E_{10} - p_x)}{1 - p_x}, p^T_{00} - \frac{p_{y_1}((1 - p_x) - p^E_{11})}{1 - p_x}\} 
            \end{align*}
            This is equal to the upper bound in Theorem 1.
            \item Case IV:  We prove the upper bound in Theorem 1 in the following case
            \begin{align*}
                1 - p_x &< p_x \leq p^E_{10}\\
                1 - p_x &< p^E_{11} < p_x\\
            \end{align*}
           Since $p^E_{10} \geq \max\{p_x, 1 - p_x\}$, Lemma \ref{lem:max-operator-gamma-0} and Lemma \ref{lem:max-operator-gamma-0-positive} imply the following over the range of the free parameters:
            \begin{equation}\label{eq:thm-1-always-greater}
                \max \{0, \mathbb{P}(Z = 1 \mid X = 1, Y = 0) - \mathbb{P}(Z = 0 \mid X = 0, Y = 0)\} = \mathbb{P}(Z = 1 \mid X = 1, Y = 0) - \mathbb{P}(Z = 0 \mid X = 0, Y = 0)    
            \end{equation} 
            However, from Lemma \ref{lem:max-operator-gamma-i-0-1}, $\max \{0, P(Z = 1 \mid X = 1, Y = 1) - P(Z = 0 \mid X = 0, Y = 1) \}$ is not restricted in this way and can take on values of either $0$ or $P(Z = 1 \mid X = 1, Y = 0) - P(Z = 0 \mid X = 0, Y = 0)$ depending on the lower bounds on $P(Z = 1 \mid X = 1, Y = 1)$ in Eq. \ref{eq:sum-bounds-thm1} and Eq. \ref{eq:indiv-bounds-thm1}. 
            
            First, when the range on $P(Z = 1 \mid X = 1, Y = 1)$ allows $\max \{0, P(Z = 1 \mid X = 1, Y = 1) - P(Z = 0 \mid X = 0, Y = 1) \}$ to equal either $P(Z = 1 \mid X = 1, Y = 1) - P(Z = 0 \mid X = 0, Y = 1)$ or $0$, then $\min \Gamma = p_{11} + p_{10} - 1$. This is because any decrease in $P(Z = 1 \mid X = 1, Y = 1)$ would increase the value of $P(Z = 1 \mid X = 1, Y = 1) - P(Z = 0 \mid X = 0, Y = 1)$, but this would be offset by the decrease in the value of $P(Z = 1 \mid X = 1, Y = 0) - P(Z = 0 \mid X = 0, Y = 0)$, since their weighted sum always adds up to $p_{11} + p_{10} - 1$, as seen in Eq. \ref{eq:weighted-sum-binary}. Similarly, any increase in $P(Z = 1 \mid X = 1, Y = 1)$ would increase the value of $P(Z = 1 \mid X = 1, Y = 0) - P(Z = 0 \mid X = 0, Y = 0)$ while driving the value of $P(Z = 1 \mid X = 1, Y = 1) - P(Z = 0 \mid X = 0, Y = 1)$ lower until 
            \[
            \max\{0, P(Z = 1 \mid X = 1, Y = 1) - P(Z = 0 \mid X = 0, Y = 1)\} = 0
            \]
            at which point it can no longer counterbalance the increase in $P(Z = 1 \mid X = 1, Y = 0) - P(Z = 0 \mid X = 0, Y = 0)$, increasing the value of $\Gamma$, and hence cannot be a minimum either. So we have showed, $\min \Gamma = p_{11} + p_{10} - 1$ in this situation. 

            But, in the case where the lower bound on $P(Z = 1 \mid X = 1, Y = 1)$ is large enough to force $\max\{0, P(Z = 1 \mid X = 1, Y = 1) - P(Z = 0 \mid X = 0, Y = 1)\} = 0$ over the range of the free parameter, then $\min \Gamma$ will be achieved at the lower bound of $P(Z = 1 \mid X = 1, Y = 1)$ since this is an increasing function of $P(Z = 1 \mid X = 1, Y = 1)$. From Lemma \ref{lem:max-operator-gamma-i-0-1}, the only lower bound on $P(Z = 1 \mid X = 1, Y = 1)$ capable of doing this is $\frac{p^T_{11} - p_{y_0}}{p_{y_1}}$. And here, $\min \Gamma = \frac{p_{y_0}(p^E_{10} - p_x)}{1 - p_x}$. 
            
            So, the value of $\min \Gamma$ is decided by whether the lower bound on $P(Z = 1 \mid X = 1, Y = 1)$ is sufficiently large to force $\max\{0, P(Z = 1 \mid X = 1, Y = 1) - P(Z = 0 \mid X = 0, Y = 1)\} = 0$. This can be equivalently stated as $\min \Gamma = \max\{p^T_{11} + p^T_{10} - 1, \frac{p_{y_0}(p^E_{10} - p_x)}{1 - p_x}\}$. 
            \begin{align*}
                p^T_{11} - \min \Gamma &= p^T_{11} - \max\{p^T_{11} + p^T_{10} - 1, p^T_{11} - \frac{p_{y_0}(p^E_{10} - p_x)}{1 - p_x}\}\\
                &= \min\{ p^T_{11} - \frac{p_{y_0}(p^E_{10} - p_x)}{1 - p_x}, p_{00}\}
            \end{align*}
            And this matches the bounds proposed in Theorem 1.\\

\end{enumerate}

Similar approaches can be used for the rest of the cases, as well as for the case when $p_x < 1 - p_x$. 

\subsection{Proof Of Lower Bound In Theorem 1}

Based on Lemma \ref{lem:max-operator-expressions}, every term in $\Delta$ can be expressed in terms of the free parameter as

\begin{enumerate}
    \item $P(Z = 1 \mid X = 1, Y = 0) - P(Z = 1 \mid X = 0, Y = 0)$
    \begin{equation}\label{eq:delta-max-operator-y0}
    \frac{p^T_{11} - p^E_{10}p_{y_0} - p_{y_1}\mathbb{P}(Z = 1\mid X = 1, Y = 1)}{(1 - p_x)p_{y_0}}
\end{equation} 

\item $\mathbb{P}(Z = 1 \mid X = 1, Y = 1) - \mathbb{P}(Z = 1 \mid X = 0, Y = 1)$
    \begin{equation}\label{eq:delta-max-operator-y1}
        \frac{\mathbb{P}(Z = 1 \mid X = 1, Y = 1) - p^E_{11}}{1 - p_x}
    \end{equation}
\end{enumerate}

First, consider the case when the following conditions can be simultaneously satisfied. 
\begin{equation}\label{eq:delta-term-1-thm1-lb}
\max\{0, P(Z = 1 \mid X = 1, Y = 0) - P(Z = 1 \mid X = 0, Y = 0)\} = P(Z = 1 \mid X = 1, Y = 0) - P(Z = 1 \mid X = 0, Y = 0)    
\end{equation}

\begin{equation}\label{eq:delta-term-2-thm1-lb}
\max\{0, P(Z = 1 \mid X = 1, Y = 1) - P(Z = 1 \mid X = 0, Y = 1)\} = P(Z = 1 \mid X = 1, Y = 1) - P(Z = 1 \mid X = 0, Y = 1)
\end{equation}
This can happen when $p^E_{11} < P(Z = 1 \mid X = 1, Y = 1) \leq \frac{p_{11} - p^\prime_{10}y_0}{y_1}$, and as proved in Lemma \ref{lem:max-operator-simul-open-delta}, happens when $p_{11} - p_{10} \geq 0$. In this case, $\min \Delta = p^T_{11} - p^T_{10}$. A similar counterbalancing argument utilized in the proof for the upper bound is employed here. We first examine how to decrease Eq. \ref{eq:delta-max-operator-y0}. This would be accomplished by increasing the value of $P(Z = 1 \mid X = 1, Y = 1)$, however, this also increases the value of Eq. \ref{eq:delta-max-operator-y1}. Since their weighted sum is a constant, it means that the increase of one counterbalances the decrease in another, however, after a certain point since the max operator corresponding to Eq. \ref{eq:delta-max-operator-y0} starts evaluating to $0$, this increase is no longer counterbalanced, leading to Eq. \ref{eq:delta-max-operator-y1} taking on a value greater than $p^T_{11} - p^T_{10}$ (the value of the sum of both max operators). 

Next, consider the case when $p^T_{11} - p^T_{10} < 0$, based on Lemma \ref{lem:max-operator-simul-open-delta}, both Eq. \ref{eq:delta-term-1-thm1-lb} and Eq. \ref{eq:delta-term-2-thm1-lb} cannot simultaneously hold. First we check whether any lower bounds are large enough to make $P(Z = 1 \mid X = 1, Y = 1) - p^E_{11} > 0$. 

First, note that when the lower bound is $0$ or $\frac{p^E_{11} - (1 - p_x)}{p_x}$, from Lemma \ref{lem:delta-max-operator-i-non-0-bounds-test} we have seen these are not large enough to make $\max\{0, P(Z = 1 \mid X = 1, Y = 1) - P(Z = 1 \mid X = 0, Y = 1)\}$ equal to a non-zero value. We now check the remaining possibilities for lower bounds using a proof by contradiction. 
    \begin{enumerate}
        \item When the lower bound on $P(Z = 1 \mid X = 1, Y = 1)$ is equal to $\frac{p^T_{11} - p_{y_0}}{p_{y_1}}$: Assume
        \begin{align*}
            \frac{p^T_{11} - p_{y_0}}{p_{y_1}} &> p^E_{11}\\
            \implies (1 - p_x)(p^T_{11} - p^T_{10}) > p_{y_0}p^E_{00}
        \end{align*}
        This is a contradiction since $p^T_{11} - p^T_{10} \leq 0$, hence 
        \[
            \frac{p^T_{11} - p_{y_0}}{p_{y_1}} \leq p^E_{11}
        \]
        \item The lower bound is $\frac{p^T_{11}p_x - p^E_{10}p_{y_0}}{p_xp_{y_1}}$: Assume 
        \begin{align*}
            \frac{p^T_{11}p_x - p^E_{10}p_{y_0}}{p_xp_{y_1}} &> p^E_{11}\\
            \implies p^T_{11} - p^T_{10} > p^E_{10}p_{y_0}
        \end{align*}
        And this is a contradiction as well for the same reasons as before, hence
        \[
            \frac{p^T_{11}p_x - p^E_{10}p_{y_0}}{p_xp_{y_1}} \leq p^E_{11}\\
        \]
    \end{enumerate}
Hence, we have shown the lower bounds are always going to allow $\max\{0, P(Z = 1 \mid X = 1, Y = 1) - P(Z = 1 \mid X = 0, Y = 1)\}$ to take on the value of $0$. 
    
From Lemma \ref{lem:delta-0-max-operator-bounds}, and a similar proof by contradiction applied to the remaining bounds, we can show that the upper bounds on $P(Z = 1 \mid X = 1, Y = 1)$ will always be greater than $\frac{p^T_{11} - p^E_{10}y_0}{y_1}$.
Since we have shown that the lower bound of $P(Z = 1\mid X = 1, Y = 1)$ is always less than $p^E_{11}$ and the upper bound is always greater than $\frac{p^T_{11} - p^E_{10}p_{y_0}}{p_{y_1}}$ and $p^T_{11} - p^T_{10} < 0 \implies \frac{p^T_{11} - p^E_{10}y_0}{y_1} < p^E_{11}$, there will always be a value of $P(Z = 1\mid X = 1, Y = 1)$ allowed such that both max operators equal $0$. 

So we have shown that when $p_{11} - p_{10} > 0$, $\Delta$ will evaluate to $p_{11} - p_{10}$, and $0$ otherwise. This concludes the proof of the lower bound. 

\section{Proof of Theorem 2}

The proof for Theorem 2 follows a similar logic with the proof of Theorem 1. We prove the lower bounds and upper bounds separately for arbitrary cardinality of support of $Y$. 

Based on Lemma \ref{lem:free-param-bounds}, the following bounds hold for the free parameters. 
{\small
\begin{align}\label{eq:sum-bounds-thm-2-gamma}
\max\begin{bmatrix}
   \frac{p^T_{11}p_x - p^E_{10}p_{y_0}}{p_x} \\
   p^T_{11} - p_{y_0}
\end{bmatrix} \leq \sum_{n = 1}^N p_{y_n}\mathbb{P}(Z = 1 \mid X = 1, Y = n) 
\leq \min\begin{bmatrix}
    \frac{(1 - p_x)p_{y_0} + p^T_{11}p_x - p^E_{10}p_{y_0}}{p_x}\\
    p^T_{11}
\end{bmatrix}
\end{align}
}

And for each $i \in \{1, \dots, N\}$, we the bounds on the following free parameters $P(Z = 1\mid X = 1, Y = i)$
\begin{align}\label{eq:individual-bounds-thm2}
    \max\begin{bmatrix}
        0\\
        \frac{p^E_{1i} - (1 - p_x)}{p_x}
    \end{bmatrix}
        \leq P(Z = 1\mid X = 1, Y = i) \leq 
    \min\begin{bmatrix}
        1\\
        \frac{p^E_{1i}}{p_x}
    \end{bmatrix}
\end{align}

\subsection{Proof for Upper Bound in Theorem 2}
From Lemma \ref{lem:max-operator-expressions}, each of these terms in $\Gamma$ defined in Lemma \ref{lem:joint-bounds} can be written as
\begin{enumerate}
    \item 
    \[P(Z = 1 \mid X = 1, Y = 0) - P(Z = 0 \mid X = 0, Y = 0)\]
    \begin{equation}\label{equation:arbitrary-gamma-y0}
        \frac{p^E_{10} - (1 - p_x)}{1 - p_x} + \frac{\sum_{n = 1}^N p_{y_n}P(Z = 1 \mid X = 1, Y = 1) - p^T_{11}}{p_{y_0}}\Bigg(\frac{p_x}{1 - p_x} - 1\Bigg) 
    \end{equation}
    \item For all $i \in \{1, \dots, N\}$:
    \[
    P(Z = 1 \mid X = 1, Y = 1) - P(Z = 0 \mid X = 0, Y = 1)
    \]
    \begin{equation}\label{equation:arbitrary-gamma-yi}
        P(Z = 1 \mid X = 1, Y = i)(1 - \frac{p_x}{1 - p_x}) + \frac{p^E_{1i} - (1 - p_x)}{1 - p_x}\\
    \end{equation}
\end{enumerate}
As before, we split the proof into three categories of cases, and provide proofs for when $p_x = 1 - p_x$, and when $p_x > 1 - p_x$. Analogous proofs can be derived for when $p_x < 1 - p_x$.\\
\newline
\textbf{Category I} In Category I, we provide proofs for when $p_x = 1 - p_x$. In this setting, we will prove the following equations hold 
\begin{equation}\label{eq:arbitrary-y-upper-bound-1}
p^T_{11} - \min_{P_i} \Gamma(P_i) = p^T_{11} - \sum_{i = 0}^N\mathbb{I}(p^E_{1i} \geq \max\{1 - p_x, p_x\})\frac{p_{y_i}(p^E_{1i} - \max\{1 - p_x, p_x\})}{\min\{1 - p_x, p_x\}}    
\end{equation}
And
\begin{equation}\label{equation:arbitrary-y-upper-bound-2}
p^T_{11} - \min_{P_i} \Gamma(P_i) = p^T_{00} - \sum_{i = 0}^N\mathbb{I}(p^E_{1i} \leq \min\{1 - p_x, p_x\})\frac{p_{y_i}(\min\{1 - p_x, p_x\} - p^E_{1i})}{\min\{1 - p_x, p_x\}}    
\end{equation}
Note that when $p_x = 1 - p_x$, $\Gamma$ is no longer a function of the free parameters since \ref{equation:arbitrary-gamma-y0} and \ref{equation:arbitrary-gamma-yi} are no longer a function of the free parameters. $\Gamma$ will equal
\begin{align*}
    \Gamma = \sum_{i = 0}^N p_{y_i}\max \{0, \frac{p^E_{1i} - (1 - p_x)}{1 - p_x} \}
\end{align*}
Next, we show on a case by case basis that $p_{11} - \Gamma$ will always equal both \ref{eq:arbitrary-y-upper-bound-1} and \ref{equation:arbitrary-y-upper-bound-2}. Partition the set $\{0, \dots, N\}$ into two disjoint subsets $R$ and $L$ such that $\forall r \in R$, $p^E_{1r} \geq p_x$, $\forall l \in L$, $p^E_{1l} \leq p_x$. Then, $p_{11} - \Gamma$ will evaluate to
\begin{equation}\label{eq:equal-x-part-a}
    p^T_{11} - \Gamma = p^T_{11} - \sum_{r \in R} p_{y_r}\frac{p^E_{1r} - (1 - p_x)}{1 - p_x}
\end{equation}
By direct comparison, this is equal to Eq. \ref{eq:arbitrary-y-upper-bound-1}. Next, since $p_x = 1 - p_x$ Eq. \ref{eq:equal-x-part-a} can be written as
\begin{align*}
    &\frac{p^T_{11}p_x - \sum_{r \in R} p_{y_r}(p^E_{1r} - (1 - p_x))}{1 - p_x}
\intertext{Using the identity $p^T_{11}p_x + p^T_{10}(1 - p_x) = \sum_{i = 0}^N p^E_{1i}p_{y_i}$}
&\implies = \frac{- p^T_{10}(1 - p_x) + \sum_{l \in L} p_{y_l}p^E_{1l} + (1 - \sum_{l \in L}p_{y_l})(1 - p_x)}{1 - p_x}\\
&\implies = p^T_{00} - \sum_{l \in L} \frac{p_{y_l}((1 - p_x) - p^E_{1l})}{1 - p_x}
\end{align*}
And by comparison to Eq. \ref{equation:arbitrary-y-upper-bound-2}, we can see these are equal as well. This proves that in the case $x = 1 - x$, $p_{11} - \Gamma$ will equal \ref{eq:arbitrary-y-upper-bound-1} and \ref{equation:arbitrary-y-upper-bound-2}. Now, we move on to the category of cases where $p_x > 1 - p_x$.\\
\newline
\textbf{Category II} In Category II, we provide proofs for when $p_x > 1 - p_x$. Broadly, the proof utilizes the fact that Eq. \ref{eq:individual-bounds-thm2} provide bounds on the individual values of the free parameters, while Eq. \ref{eq:sum-bounds-thm-2-gamma} provide bounds on the weighted sum of the free parameters. Depending on which of these two bounds are more restrictive, we get two different upper bound on the $PNS$. And we show that choosing the more restrictive bounds on the free parameter is equivalent to evaluating the minimum operator in the upper bound. 
\begin{enumerate}
    \item Case I: We prove the upper bound in Theorem 2 in the following case. $p^E_{10} \leq \min{p_x, 1 - p_x}$. Partition $\{1, \dots N\}$ into three non-empty disjoint subsets $L$, $R$ and $S$ such that $\forall l \in L \; p^E_{1l} \leq \min\{p_x, 1 - p_x\}$,  $\forall r \in R \; p^E_{1r} \geq \max\{p_x, 1 - p_x\}$,  and $\forall s \in S \; 1 - p_x < p^E_{1s} < p_x$. 

    Then, fromLemma \ref{lem:gamma-0-lower-side-px} and Lemma \ref{lem:gamma-0-lower-side-1-minus-px}, $\max\{P(Z = 1\mid X = 1, Y = 0) - P(Z = 0\mid X = 0, Y = 0)\} = 0$ over the entire range of the free parameters. Similarly,from Lemma \ref{lem:max-operator-gamma-i-0-1} and Lemma \ref{lem:max-operator-gamma-i-0-2},  $\forall l \max\{P(Z = 1\mid X = 1, Y = l) - P(Z = 0\mid X = 0, Y = l)\} = 0$ over the entire range of free parameters.  

    For the remaining max operators in $\Gamma$ corresponding to $S$ and $R$, they will all vary independently and be decreasing functions of their corresponding free parameter. Hence, to minimize $\Gamma$ over the range of free parameters, we want to minimize each of the max operators in $\Gamma$ corresponding to $S$ and $R$ while obeying the bounds on the free parameters. 

    From Lemma \ref{lem:max-operator-gamma-i} and Lemma \ref{lem:max-operator-gamma-i-positive}, $\forall r \in R \max\{P(Z = 1\mid X = 1, Y = r) - P(Z = 0\mid X = 0, Y = r)\} = P(Z = 1\mid X = 1, Y = r) - P(Z = 0\mid X = 0, Y = r)$. Similarly, all the max operators corresponding to $S$ can be zero or non-zero.

    The following values of free parameters will minimize $\Gamma$.
    \begin{enumerate}
        \item $\forall l \in L \; P(Z = 1 \mid X = 1, Y = l) = 0$
        \item $\forall r \in R \; P(Z = 1 \mid X = 1, Y = l) = 1$
        \item $\forall s \in S \; P(Z = 1 \mid X = 1, Y = l) = \frac{p^E_{1s} - (1 - p_x)}{p_x - (1 - p_x)}$
    \end{enumerate}
    This setting of free parameters minimizes $\Gamma$ because it sets all the max operators capable of attaining $0$ equal to $0$, while minimizing the remaining max operators. This would result in $\Gamma$ evaluating as
    \[
    \Gamma = \sum_{r \in R} p_{y_r}\frac{p^E_{1r} - p_x}{1 - p_x}
    \]
    However, we must check whether this setting of the free parameters obeys the restrictions on the weighted sum of free parameters. To this end, we must check whether the weighted sum of the free parameters satisfies the bounds in Eq. \ref{eq:sum-bounds-thm-2-gamma}, which equals the bounds below from an application of Lemma \ref{lem:bounds-conditions} along with $p^E_{10} \leq \min\{p_x, 1 - p_x\}$.
    \begin{align}
        \frac{p^T_{11}p_x - p^E_{10}p_{y_0}}{p_x} 
         \leq \sum_{s \in S} y_s \frac{p^E_{1s} - (1 - p_x)}{p_x - (1 - p_x)} + \sum_{r \in R} p_{y_r}
        \leq p^T_{11}
    \end{align}
    With respect to the lower bound, even if $\sum_{s \in S} y_s \frac{p^E_{1s} - (1 - p_x)}{p_x - (1 - p_x)} + \sum_{r \in R} p_{y_r}$ is not sufficiently large enough to uphold the lower bound, any increase in the free parameters will not change the value of $\Gamma$ since all the max operators that are decreasing functions of the free parameter are either already minimized or set to value that makes them evaluate to $0$. This will not change the value of $\Gamma$ since the corresponding max operators will still remain $0$ even with the increase in the value of free parameters.     

    However, when this upper bound does not hold, then, we must minimize $\Gamma$ while obeying $\sum_{i = 1}^N p_{y_i}P(Z = 1 \mid X = 1, Y = i) \leq p^T_{11}$. To evaluate this, note that any solution to $\min \Gamma$ will set all the max operators $\forall s \in S\; P(Z = 1 \mid X = 1, Y = s) - P(Z = 0\mid X = 0, Y = s)$ to exactly $0$ or greater, since further reduction still gives a zero, while restricting the range of the other free parameters, thereby increasing the value of $\Gamma$ since other max operators cannot be minimized more by increasing the value of the free parameter. Therefore, $\forall s \in S\; \max\{0, P(Z = 1 \mid X = 1, Y = s) - P(Z = 0\mid X = 0, Y = s)\} = P(Z = 1 \mid X = 1, Y = s) - P(Z = 0\mid X = 0, Y = s) = 0$. Now, since the remaining max operators corresponding to $L$ along with $\max\{0, P(Z = 1 \mid X = 1, Y = 1) - P(Z = 0\mid X = 0, Y = 0)$ will equal $0$, $\Gamma$ will be evaluated as
    \begin{align}
        \Gamma &= \min_{P} \sum_{s \in S} P(Y = s)(P(Z = 1\mid X = 1, Y = s) - P(Z = 0\mid X = 1, Y = s))\notag\\ 
        &+ \sum_{r \in R} P(Y = r)(P(Z = 1\mid X = 1, Y = r) - P(Z = 0\mid X = 1, Y = r))    
    \end{align}
    And expressing this in terms of free parameters as 
    \begin{align*}
    \Gamma &= \min_P \sum_{i \in R, S}p_{y_i}(P(Z = 1 \mid X = 1, Y = i)\Bigg(1 - \frac{p_x}{1 - p_x}\Bigg) + \sum_{i \in R, S}p_{y_i} \frac{p^E_{11} - (1 - p_x)}{1 - p_x})\\
    &= \Bigg(1 - \frac{p_x}{1 - p_x}\Bigg)\sum_{i \in R, S}p_{y_i}P(Z = 1 \mid X = 1, Y = i) + \sum_{i \in R, S} \frac{p^E_{11} - (1 - p_x)}{1 - p_x}
    \end{align*}
    
    Since $\Gamma$ is a decreasing function of the weighted sum of free parameters, it will be minimized at the upper bound of $\sum_{i = 1}^N P(Z = 1 \mid X = 1, Y = i)$, which in this case equals $p_{11}$. And since the solution for $\min \Gamma$ will set $\forall l \in L\; P(Z = 1 \mid X = 1, L = l) = 0$, $\sum_{i \in R, S}p_{y_i}(P(Z = 1 \mid X = 1, Y = i) = p^T_{11}$. Plugging this into the expression of $\Gamma$ evaluates to 
    \begin{align*}
        &\Bigg(1 - \frac{p_x}{1 - p_x}\Bigg)p^T_{11} + \sum_{i \in R, S} p_{y_i}\frac{p^E_{11} - (1 - p_x)}{1 - p_x}\\
        &\implies = p^T_{11} +  \frac{-p^T_{11}p_x + \sum_{i \in R, S} p_{y_i}(p^E_{11} - (1 - p_x))}{1 - p_x}\\
        &\implies = p^T_{11} +  \frac{p_{10}(1 - p_x) - \sum_{l \in l} p_{y_l}p^E_{1l} - \sum_{i \in R, S} p_{y_i}(1 - p_x)}{1 - p_x}\\
        &\implies = p^T_{11} +  \frac{p_{10}(1 - p_x) - (1 - p_x) - \sum_{l \in l} p_{y_l}p^E_{1l} - \sum_{i \in R, S} p_{y_i}(1 - p_x) + (1 - p_x)}{1 - p_x}\\
        &\implies = p^T_{11} +  p^T_{10} - 1 + \frac{ - \sum_{l \in l} p_{y_l}p^E_{1l} + \sum_{l \in L} p_{y_l}(1 - p_x)}{1 - p_x}\\
        &\implies = p^T_{11} +  p^T_{10} - 1 + \sum_{l \in L}p_{y_l}\frac{(1 - p_x) - p^E_{1l}}{1 - p_x}
    \end{align*}
    
    Hence, $\min \Gamma$ will be decided by whether $p^T_{11} < \sum_{s \in S} y_s \frac{p^E_{1s} - (1 - p_x)}{p_x - (1 - p_x)} + \sum_{r \in R} p_{y_r}$ or not. And since $\Gamma$ is a decreasing function of the free parameters, this is equivalent to choosing $\max\{ \sum_{r \in R} p_{y_r}\frac{p^E_{1r} - p_x}{1 - p_x}, p^T_{11} +  p^T_{10} - 1 + \sum_{l \in L}p_{y_l}\frac{(1 - p_x) - p^E_{1l}}{1 - p_x}\}$. Since this this is subtracted from $p^T_{11}$, this can be written as
    \begin{align*}
        &p_{11} - \max\{ \sum_{r \in R} p_{y_r}\frac{p^E_{1r} - p_x}{1 - p_x}, p^T_{11} +  p^T_{10} - 1 + \sum_{l \in L}p_{y_l}\frac{(1 - p_x) - p^E_{1l}}{1 - p_x}\}\\ &\hspace{0.5in}= \min\{p^T_{11} -  \sum_{r \in R} p_{y_r}\frac{p^E_{1r} - p_x}{1 - p_x}, p^T_{00} - \sum_{l \in L}p_{y_l}\frac{(1 - p_x) - p^E_{1l}}{1 - p_x}\}
    \end{align*}
    This concludes the proof for this case. 
    \item Case II: We prove the upper bound in Theorem 2 in the following case. $p^E \geq \max{p_x, 1 - p_x}$. Partition $\{1, \dots N\}$ into three non-empty disjoint subsets $L$, $R$ and $S$ such that $\forall l \in L \; p^E_{1l} \leq \min\{p_x, 1 - p_x\}$,  $\forall r \in R \; p^E_{1r} \geq \max\{p_x, 1 - p_x\}$,  and $\forall s \in S \; 1 - p_x < p^E_{1s} < p_x$. 

    Then, from Lemma \ref{lem:max-operator-gamma-0} and Lemma \ref{lem:max-operator-gamma-0-positive}, $\max\{0, P(Z = 1\mid X = 1, Y = 0) - P(Z = 0\mid X = 0, Y = 0)\} = P(Z = 1\mid X = 1, Y = 0) - P(Z = 0\mid X = 0, Y = 0)$ over the entire range of the free parameters. Similarly, following a similar argument as the previous case, $\forall l \max\{P(Z = 1\mid X = 1, Y = l) - P(Z = 0\mid X = 0, Y = l)\} = 0$, and $\forall r \max\{P(Z = 1\mid X = 1, Y = r) - P(Z = 0\mid X = 0, Y = r)\} = P(Z = 1\mid X = 1, Y = r) - P(Z = 0\mid X = 0, Y = r)$ over the entire range of free parameters.  

    Consequently, $\Gamma$ will evaluate to
    \begin{align}
        \Gamma =& \sum_{i \in \{0, R\}} P(Y = i)(P(Z = 1 \mid X = 1, Y = i) - P(Z = 0 \mid X = 0, Y = i))\notag\\ 
        &+ \sum_{s \in S}P(Y = s) \max\{0, P(Z = 1 \mid X = 1, Y = s) - P(Z = 0\mid X = 0, Y = s)
    \end{align}
    Expressing this in terms of free parameters we get
    \begin{align}
        \Gamma =& p_{y_0}\frac{p^E_{10} - (1 - p_x)}{1 - p_x} + (\sum_{n = 1}^N p_{y_n}P(Z = 1 \mid X = 1, Y = n) - p^T_{11})\Bigg(\frac{p_x}{1 - p_x} - 1\Bigg)\notag\\
        &+  \sum_{r \in R}p_{y_i} \{P(Z = 1 \mid X = 1, Y = r)\Bigg(1 - \frac{p_x}{1 - p_x}\Bigg) + \frac{p^E_{11} - (1 - p_x)}{1 - p_x}\}\notag\\ 
        &+ \sum_{s \in S} P(Y = s)\max\{0, P(Z = 1 \mid X = 1, Y = s)\Bigg(1 - \frac{p_x}{1 - p_x}\Bigg) + \frac{p^E_{1s} - (1 - p_x)}{1 - p_x}\}
    \end{align}
    The above can be simplified to
    \begin{align*}
        \Gamma = & p_{y_0}\frac{p^E_{10} - (1 - p_x)}{1 - p_x} - p_{11}\Bigg(\frac{p_x}{1 - p_x} - 1\Bigg) + \Bigg(\frac{p_x}{1 - p_x} - 1\Bigg)\sum_{n = 1}^N p_{y_n}P(Z = 1 \mid X = 1, Y = n)\notag\\
        &+  \sum_{r \in R}\Bigg(1 - \frac{p_x}{1 - p_x}\Bigg)p_{y_i}P(Z = 1 \mid X = 1, Y = r) + \sum_{r \in R} p_{y_r}\frac{p^E_{1r} - (1 - p_x)}{1 - p_x}\notag\\ 
        &+ \sum_{s \in S} p_{y_s}\max\{0, P(Z = 1 \mid X = 1, Y = s)\Bigg(1 - \frac{p_x}{1 - p_x}\Bigg) + \frac{p^E_{1s} - (1 - p_x)}{1 - p_x}\}\\
        &\implies \Gamma  = p_{y_0}\frac{p^E_{10} - (1 - p_x)}{1 - p_x} - p_{11}\Bigg(\frac{p_x}{1 - p_x} - 1\Bigg)\\
        &+ \sum_{r \in R} p_{y_r}\frac{p^E_{1r} - (1 - p_x)}{1 - p_x} + \Bigg(\frac{p_x}{1 - p_x} - 1\Bigg)\sum_{n \in N \setminus R} p_{y_n}P(Z = 1 \mid X = 1, Y = n)\\ 
        &+ \sum_{s \in S} p_{y_s}\max\{0, P(Z = 1 \mid X = 1, Y = s)\Bigg(1 - \frac{p_x}{1 - p_x}\Bigg) + \frac{p^E_{1s} - (1 - p_x)}{1 - p_x}\}\notag\\
    \end{align*}
    This quantity will be minimized when Eq. \ref{eq:case-to-minimize} in minimized, since $p_x > 1 - p_x$ and probabilities are non-negative and $N \setminus R \equiv \{S, L\}$.
    \begin{align}\label{eq:case-to-minimize}
    &\Bigg(\frac{p_x}{1 - p_x} - 1\Bigg)\sum_{l \in L} p_{y_l}P(Z = 1 \mid X = 1, Y = l) + \Bigg(\frac{p_x}{1 - p_x} - 1\Bigg)\sum_{s \in S} p_{y_s}P(Z = 1 \mid X = 1, Y = s)\notag \\ 
    &+ \sum_{s \in S} p_{y_s}\max\{0, P(Z = 1 \mid X = 1, Y = s)\Bigg(1 - \frac{p_x}{1 - p_x}\Bigg) + \frac{p^E_{1s} - (1 - p_x)}{1 - p_x}\}
    \end{align}
    First, to minimize $\Gamma$ (in the later part of this proof we explore the behavior when the following does not hold) all the free parameters corresponding to $l$ will be minimized when set to $0$, which the individual bounds on the free parameters allow, based on our assumptions for this case. Note that the max operator equal to $\max\{0, P(Z = 1 \mid X = 1, Y = s)\Bigg(1 - \frac{p_x}{1 - p_x}\Bigg) + \frac{p^\prime_{1s} - (1 - p_x)}{1 - p_x}\}$, which has a counterbalancing effect (similar to that described in Case IV in Category II of the upper bound proof in Theorem 1) with $\Bigg(\frac{p_x}{1 - p_x} - 1\Bigg) p_{y_s}P(Z = 1 \mid X = 1, Y = s)$.

    Hence the minimum of Eq. \ref{eq:case-to-minimize} is attained when each max operator is exactly $P(Z = 1 \mid X = 1, Y = s)\Bigg(1 - \frac{p_x}{1 - p_x}\Bigg) + \frac{p^\prime_{1s} - (1 - p_x)}{1 - p_x}$, since any deviation away from this will only increase the value of the free parameters without counterbalancing their increase with the max operator since it would be $0$. Plugging this minimum quantity back into the equation for $\Gamma$ we get:
    \begin{align*}
        &\Gamma = p_{y_0}\frac{p^E_{10} - (1 - p_x)}{1 - p_x} - p_{11}\Bigg(\frac{p_x}{1 - p_x} - 1\Bigg) + \sum_{i \in R} p_{y_r}\frac{p^E_{11} - (1 - p_x)}{1 - p_x}\\ 
        &+ \sum_{s \in S} p_{y_s}\frac{p^E_{1s} - (1 - p_x)}{1 - p_x}\}\notag\\
    \end{align*}
    And following a similar calculation as the previous case, this can be simplified to
    \begin{align*}
        \Gamma = p^T_{11} + p^T_{10} - 1 + \sum_{l \in L}p_{y_l}\frac{(1 - p_x) - p^E_{1l}}{1 - p_x} 
    \end{align*}
    However, the above calculations assume that the constraints on the free parameters allow them to take on values such that the max operators behave as we describe. Specifically, this involves $P(Z = 1\mid X = 1, Y = r) = \frac{p^\prime_{1r}}{p_x}$, letting all the variables corresponding to $S$ equal $\frac{p^\prime_{1s} - (1 - p_x)}{p_x - (1 - p_x)}$, and all the variables in $L$ equal to $0$. Based on the individual bounds on the free parameter described in Eq. \ref{eq:individual-bounds-thm2}, this setting of free parameter obeys these bounds. However, in the case where Eq. \ref{eq:sum-bounds-thm-2-gamma} are the more restrictive bounds, i.e. $\sum_{i = 1}^N p_{y_N}P(Z = 1 \mid X = 1, Y = i) \geq p_{11} - p_{y_0}$, then this setting of free parameters may no longer hold. 
    
    In this case, as before, the max operator equal to $\max\{0, P(Z = 1 \mid X = 1, Y = s)\Bigg(1 - \frac{p_x}{1 - p_x}\Bigg) + \frac{p^\prime_{1s} - (1 - p_x)}{1 - p_x}\}$, which has a counterbalancing effect described in Theorem 1 with $\Bigg(\frac{p_x}{1 - p_x} - 1\Bigg) p_{y_s}P(Z = 1 \mid X = 1, Y = s)$, following a similar argument for counterbalancing the increase in the representation of max operators corresponding to $S$ as presented in Case I, and so, $\Gamma$ will evaluate to 
    \begin{align}
        \Gamma =& \sum_{i \in \{0, R\}} P(Y = i)(P(Z = 1 \mid X = 1, Y = i) - P(Z = 0 \mid X = 0, Y = i))\notag\\ 
        &+ \sum_{s \in S}P(Y = s) (P(Z = 1 \mid X = 1, Y = s) - P(Z = 0\mid X = 0, Y = s))
    \end{align}
    \begin{align*}
    &= p_{y_0}\frac{p^E_{10} - (1 - p_x)}{1 - p_x} - p^T_{11}\Bigg(\frac{p_x}{1 - p_x} - 1\Bigg) + \Bigg(\frac{p_x}{1 - p_x} - 1\Bigg)\sum_{n = 1}^N p_{y_n}P(Z = 1 \mid X = 1, Y = n)\notag\\
        &+  \sum_{i \in R}\Bigg(1 - \frac{p_x}{1 - p_x}\Bigg)p_{y_i}P(Z = 1 \mid X = 1, Y = i) + \sum_{r \in R} p_{y_r}\frac{p^E_{1r} - (1 - p_x)}{1 - p_x}\notag\\ 
        &+ \sum_{s \in S} p_{y_s}(P(Z = 1 \mid X = 1, Y = s)\Bigg(1 - \frac{p_x}{1 - p_x}\Bigg) + \frac{p^E_{1s} - (1 - p_x)}{1 - p_x})
    \end{align*}
    And canceling out relevant terms yields
    \begin{align}\label{eq:upper-bounds-thm2-case-II-to-min}
    &= p_{y_0}\frac{p^E_{10} - (1 - p_x)}{1 - p_x} - p^T_{11}\Bigg(\frac{p_x}{1 - p_x} - 1\Bigg) + \Bigg(\frac{p_x}{1 - p_x} - 1\Bigg)\sum_{l \in L} p_{y_l}P(Z = 1 \mid X = 1, Y = l)\notag\\
        &+ \sum_{r \in R} p_{y_r}\frac{p^E_{1r} - (1 - p_x)}{1 - p_x} + \sum_{s \in S} p_{y_s}\frac{p^E_{1s} - (1 - p_x)}{1 - p_x}
    \end{align}

    This is an increasing function of the free parameters, hence, to minimize this quantity, we must find the minimum allowed value of $\sum_{l \in L} p_{y_l}P(Z = 1 \mid X = 1, Y = l)$ while still respecting $\sum_{i = 1}^N p_{y_N}P(Z = 1 \mid X = 1, Y = i) \geq p^T_{11} - p_{y_0}$ and having the max operators behave the same way. To corresponds to setting $\forall r \in R$, set $P(Z = 1 \mid X = 1, Y =r) = 1$ to the highest possible value since these cancel out. Next, set $\forall s \in S$, set $P(Z = 1 \mid X = 1, S = s) = \frac{p^E_{1s} - (1 - p_x)}{p_x - (1 - p_x)}$, since any value greater than this loses the counterbalancing effect between $\max\{0, P(Z = 1 \mid X = 1, Y = s)\Bigg(1 - \frac{p_x}{1 - p_x}\Bigg) + \frac{p^E_{1s} - (1 - p_x)}{1 - p_x}\}$ and $\Bigg(\frac{p_x}{1 - p_x} - 1\Bigg) p_{y_s}P(Z = 1 \mid X = 1, Y = s)$, increasing the value of $\Gamma$. This entails the following bounds on the weighted sum of the free parameters $\sum_{i = 1}^N p_{y_n}P(Z = 1 \mid X = 1, Y = n)$.
    \begin{align*}
        p^T_{11} - p_{y_0} \leq \sum_{l \in L}p_{y_l} P(Z = 1 \mid X = 1, Y = l) + \sum_{s \in S} p_{y_s} \frac{p^E_{1s} - (1 - p_x)}{p_x - (1 - p_x)} + \sum_{r \in R}p_{y_r}
    \end{align*}
    Substituting the lowest possible value for $\sum_{l \in L}p_{y_l}P(Z = 1 \mid X = 1, Y = l)$ into Eq. \ref{eq:upper-bounds-thm2-case-II-to-min}
    \begin{align*}
    &= p_{y_0}\frac{p^E_{10} - (1 - p_x)}{1 - p_x} - p_{11}\Bigg(\frac{p_x}{1 - p_x} - 1\Bigg)\\
    &+ \Bigg(\frac{p_x}{1 - p_x} - 1\Bigg)\Bigg(p_{11} - p_{y_0} - \sum_{s \in S} p_{y_s} \frac{p^E_{1s} - (1 - p_x)}{p_x - (1 - p_x)} - \sum_{r \in R}p_{y_r} \Bigg)\notag\\
    &+ \sum_{r \in R} p_{y_r}\frac{p^E_{1r} - (1 - p_x)}{1 - p_x} + \sum_{s \in S} p_{y_s}\frac{p^E_{1s} - (1 - p_x)}{1 - p_x}
    \end{align*}
    
    And this can be simplified as
    \begin{align*}
    &= p_{y_0}\frac{p^E_{10} - p_x}{1 - p_x} 
    + \Bigg(\frac{p_x}{1 - p_x} - 1\Bigg)\Bigg( - \sum_{s \in S} p_{y_s} \frac{p^E_{1s} - (1 - p_x)}{p_x - (1 - p_x)} - \sum_{r \in R}p_{y_r} \Bigg)\notag\\
    &+ \sum_{r \in R} p_{y_r}\frac{p^E_{1r} - (1 - p_x)}{1 - p_x} + \sum_{s \in S} p_{y_s}\frac{p^E_{1s} - (1 - p_x)}{1 - p_x}\\
    &\implies = p_{y_0}\frac{p^E_{10} - p_x}{1 - p_x} 
    + \sum_{r \in R} p_{y_r}\frac{p^E_{1r} - p_x}{1 - p_x} 
    \end{align*}
    
    And subtracting this from $p^T_{11}$, and applying a similar argument as presented in Case I between the correspondence of the  bounds on the free parameter with the bounds can be used to match the bounds in Theorem 2.
    \item Case III:  We prove the upper bound in Theorem 2 in the following case.  $1 - p_x < p^E_{10} < p_x$. Partition $\{1, \dots N\}$ into non-empty three disjoint subsets $L$, $R$ and $S$ such that $\forall l \in L \; p^E_{1l} \leq \min\{p_x, 1 - p_x\}$,  $\forall r \in R \; p^E_{1r} \geq \max\{p_x, 1 - p_x\}$,  and $\forall s \in S \; 1 - p_x < p^E_{1s} < p_x$. Following the argument used in previous cases, $\Gamma$ will equal:
    \begin{align}
        \Gamma =& P(Y = 0)\max\{0, P(Z = 1 \mid X = 1, Y = 0) - P(Z = 0\mid X = 0, Y = 0)\}\notag\\
        &+ \sum_{r \in R} P(Y = i)(P(Z = 1 \mid X = 1, Y = r) - P(Z = 0 \mid X = 0, Y = r))\notag\\ 
        &+ \sum_{s \in S}P(Y = s) \max\{0, P(Z = 1 \mid X = 1, Y = s) - P(Z = 0\mid X = 0, Y = s)\}
    \end{align}
    The first approach to minimize $\Gamma$ is to find values of the free parameters such that $\forall s \in S \max\{0, P(Z = 1 \mid X = 1, Y = s) - P(Z = 0 \mid X = 0, Y = s) \} = 0 $ and similarly, every max operator corresponding to $R$ is minimized. The following setting of the free parameters accomplish this:
    \begin{enumerate}
        \item $\forall r \in R \; P(Z = 1 \mid X = 1, Y = l) = 1$
        \item $\forall s \in S \; P(Z = 1 \mid X = 1, Y = l) = \frac{p^E_{1s} - (1 - p_x)}{p_x - (1 - p_x)}$
    \end{enumerate}
    Additionally, we must choose values of the free parameters such that $\max\{0, P(Z = 1 \mid X = 1, Y = 0) - P(Z = 0\mid X = 0, Y = 0)\}$ is still $0$, which from Lemma \ref{lem:max-operator-gamma-0} will happen when
    \[
    \sum_{n = 1}^N p_{y_n}P(Z = 1 \mid X = 1, Y = n) \leq \frac{p_{y_0}((1 - p_x) - p^E_{10})}{p_x - (1 - p_x)} + p^T_{11}
    \]
    Since $\forall l \in L \; \max\{0, P(Z = 1 \mid X = 1, Y = l) - P(Z = 0 \mid X = 0, Y = 0) = 0$, checking whether the above equation holds for $\forall l \in L \; P(Z = 1 \mid X = 1, Y = l) = 0$ holds is sufficient, since any increase in the value of the free parameters can only increase the value $P(Z = 1 \mid X = 1, Y = 0) - P(Z = 0\mid X = 0, Y = 0)$. This is equivalent to checking the following inequality:
    \begin{equation}\label{eq:pivotal-eq-case-3}
        \sum_{s \in S} y_s \frac{p^E_{1s} - (1 - p_x)}{p_x - (1 - p_x)} + \sum_{r \in R} p_{y_r} \leq \frac{p_{y_0}((1 - p_x) - p^E_{10})}{p_x - (1 - p_x)} + p^T_{11}
    \end{equation}
    This is inequality may or may not hold, depending on the values of $P^T(Z, X)$ and $P^E(Z, Y)$. When it does hold, $\min \Gamma$ will have all max operators except those corresponding to $R$ equal $0$, and hence $\min \Gamma$ will equal:
    \begin{align*}
        \min \Gamma & = \sum_{r \in R}p_{y_r}\frac{p^E_{1r} - (1 - p_x)}{p_x}\\
        p^T_{11} - \min \Gamma &= p^T_{11} - \sum_{r \in R}p_{y_r}\frac{p^E_{1r} - (1 - p_x)}{p_x}
    \end{align*}
    However, when the inequality in Eq. \ref{eq:pivotal-eq-case-3} does not hold, we must use the following approach.

    When the max operators corresponding to $S$, along with $\max\{0, P(Z = 1 \mid X = 1, Y = 0) - P(Z = 0\mid X = 0, Y = 0)\}$ cannot simultaneously be set to $0$ as above, we utilize the property that $\max\{0, P(Z = 1 \mid X = 1, Y = 0) - P(Z = 0\mid X = 0, Y = 0)\}$, $\forall r \in R \max\{0, P(Z = 1 \mid X = 1, Y = r) - P(Z = 0\mid X = 0, Y = r)\}$ and $\forall l \in L \max\{0, P(Z = 1 \mid X = 1, Y = l) - P(Z = 0\mid X = 0, Y = l)\}$ counterbalance each other, i.e. increasing the value of the free parameter in $P(Z = 1 \mid X = 1, Y = i) - P(Z = 0\mid X = 0, Y = i)$ for $i \in {S, R}$ equally decreases the value of that free parameter in $ P(Z = 1 \mid X = 1, Y = 0) - P(Z = 0\mid X = 0, Y = 0$, and a similar behavior is seen when decreasing the value of the free parameter. So, $\Gamma$ will be minimized with respect to $P(Z = 1 \mid X = 1, Y = r) \forall r \in R$ and similarly for $S$ when $\max\{0, P(Z = 1 \mid X = 1, Y = 0) - P(Z = 0\mid X = 0, Y = 0)\} \geq 0$, $\forall r \in R \max\{0, P(Z = 1 \mid X = 1, Y = r) - P(Z = 0\mid X = 0, Y = r)\} \geq 0$ and $\forall l \in L \max\{0, P(Z = 1 \mid X = 1, Y = l) - P(Z = 0\mid X = 0, Y = l)\} \geq 0$. And in this case $\Gamma$ will equal
    \begin{align*}
        \Gamma =& P(Y = 0)(P(Z = 1 \mid X = 1, Y = 0) - P(Z = 0\mid X = 0, Y = 0))\notag\\
        &+ \sum_{r \in R} P(Y = i)(P(Z = 1 \mid X = 1, Y = r) - P(Z = 0 \mid X = 0, Y = r))\notag\\ 
        &+ \sum_{s \in S}P(Y = s) (0, P(Z = 1 \mid X = 1, Y = s) - P(Z = 0\mid X = 0, Y = s))
    \end{align*}
    Expressing this in terms of free parameters equals
    \begin{align*}
        \Gamma &= p_{y_0}\frac{p^E_{10} - (1 - p_x)}{1 - p_x} - p^T_{11}\Bigg(\frac{p_x}{1 - p_x} - 1\Bigg) + \Bigg(\frac{p_x}{1 - p_x} - 1\Bigg)\sum_{l \in L} p_{y_l}P(Z = 1 \mid X = 1, Y = l)\notag\\
        &+ \sum_{r \in R} p_{y_r}\frac{p^E_{1r} - (1 - p_x)}{1 - p_x} + \sum_{s \in S} p_{y_s}\frac{p^E_{1s} - (1 - p_x)}{1 - p_x}
    \end{align*}
    And following a similar calculation as Case II, this will be minimized when $P(Z = 1 \mid X = 1, Y = l) = 0 \forall l \in L$, and this simplifies to  
    \[
    \min \Gamma = p^T_{11} + p^T_{10} - 1 + \sum_{l \in L}p_{y_l}\frac{(1 - p_x) - p^E_{1l}}{1 - p_x} 
    \]
    So, when Eq. \ref{eq:pivotal-eq-case-3} holds, we get $\min \Gamma = \sum_{r \in R}p_{y_r}\frac{p^E_{1r} - (1 - p_x)}{p_x}$ and $\min \Gamma = p^T_{11} + p^T_{10} - 1 + \sum_{l \in L}p_{y_l}\frac{(1 - p_x) - p^E_{1l}}{1 - p_x}$ otherwise. A similar argument to that presented in Case I can be used to obtain the following bound:
    \begin{align*}
        p^T_{11} - \min \Gamma = \min\{p^T_{11} - \sum_{r \in R}p_{y_r}\frac{p^E_{1r} - (1 - p_x)}{p_x}, p^T_{00} - \sum_{l \in L}p_{y_l}\frac{(1 - p_x) - p^E_{1l}}{1 - p_x}\}
    \end{align*}
    \item Case IV: Next, we consider the case where for all $i \in \{0, \dots, N\}$, $1 - p_x < p^E_{1i} < p_x$. Then, from Lemma \ref{lem:max-operarot-simultaneous-open}, for all max operators to be greater than or equal to $0$ simultaneously, $p^T_{11} + p^T_{10} - 1 \geq 0$. Here, by a similar argument as Theorem 1, if $p^T_{11} + p^T_{10} - 1 \geq 0$, then all max operators will cancel out the values for the free parameter, and any shift from this will be greater that $p^T_{11} + p^T_{10} - 1$ since max operators will not be able to offset the increase in each other. 

    \item Case V: Next, consider the case where $p^T_{11} + p^T_{10} - 1 < 0$. Here all max operators cannot simultaneously be greater than or equal to $0$ from \ref{lem:max-operarot-simultaneous-open}, and we prove that there is always a valid setting of the free parameters such that all max operators evaluate to $0$. 

    When $P(Z = 1 \mid X = 1, Y = i) = \frac{p^E_{1i} - (1 - p_x)}{p_x - (1 - p_x)}$, then the $\max\{0, P(Z = 1 \mid X = 1, Y = i) - P(Z = 0 \mid X = 0, Y = i\} = 0$. However, we still need to examine the behavior of $\max\{0, P(Z = 1 \mid X = 1, Y = 0) - P(Z = 0\mid X = 0, Y = 0\}$ at this value of the free parameters. Using a proof by contra positive, we can show that $p^T_{11} + p^T_{10} - 1 < 0 \implies \sum_{i = 1}^N p_{y_i}\frac{p^E_{1i} - (1 - p_x)}{p_x - (1 - p_x)} < \frac{p_{y_0}((1 - p_x) - p^E_{10})}{p_x - (1 - p_x)} + p_{11}$, implying this setting of the free parameters $\frac{p^E_{1i} - (1 - p_x)}{p_x - (1 - p_x)}$, all the max operators to evaluate to $0$. 
    
    Since Lemma \ref{lem:gamma-0-middle} shows the upper bound is greater than $\frac{p_{y_0}((1 - p_x) - p^E_{10})}{p_x - (1 - p_x)} + p_{11}$, only the lower bound needs to be checked against the presented value of the free parameters. However, even if the lower bound on the weighted sum of the free parameters is greater than $\sum_{i = 1}^N p_{y_i}\frac{p^\prime_{1i} - (1 - p_x)}{p_x - (1 - p_x)}$, this still allows for $\max\{0, P(Z = 1 \mid X = 1, Y = 0) - P(Z = 0\mid X = 0, Y = 0)\} = 0$ from Lemma \ref{lem:gamma-0-lower-side-px}, and the remaining max operators are will continue to evaluate to $0$ since they are decreasing functions of the max operators. This demonstrates the bounds on the $PNS$ will equal $\min\{p^T_{11}, p^T_{00}\}$ in this case. 
\end{enumerate}

\subsection{Proof for Lower Bound in Theorem 2}

As seen in Lemma \ref{lem:max-operator-simul-open-delta}, when $p^T_{11} - p^T_{10} \geq 0$, for all $i \in \{0, \dots, N\}$, the following conditions can simultaneously hold:
\begin{equation}\label{eq:delta-max-simul-thm2}
\max\{0, P(Z = 1 \mid X = 1, Y = 1) - P(Z = 1 \mid X = 0, Y = 1)\} = P(Z = 1 \mid X = 1, Y = i) - P(Z = 1 \mid X = 0, Y = i)    
\end{equation}

In this case, $p_{11} - p_{10}$ will be the minimum since
\[
\sum_Y P(Y) (P(Z = 1 \mid X = 1, Y ) - P(Z = 1 \mid X = 0, Y)) = p^T_{11} - p^T_{10}
\]
and any change to this requires changing the value of the free parameters such that a  max operator evaluates to $0$. However, similar to the argument present in the proof for the lower bound of Theorem 1, this change to the value of the free parameter results in the loss of the the counterbalancing effect that max operators have on each other, leading to an increase in the overall value of $\Delta$. 

Next, consider the case where $p^T_{11} - p^T_{10} < 0$, then all $\forall i \in \{0, \dots , N\}$ Eq. \ref{eq:delta-max-simul-thm2} cannot simultaneously hold, as seen in Lemma \ref{lem:max-operator-simul-open-delta}. Here we show that a setting of the free parameters such that all max operators in equal $0$ will be a valid solution to the system of equations. For $i \in \{1, \dots , N\}$, each of their corresponding max operators will equal $0$ when $P(Z = 1\mid X = 1, Y = i) = p^E_{1i}$, as seen from Lemma \ref{lem:max-operator-expressions}. As seen in Lemma \ref{lem:delta-max-operator-i-non-0-bounds-test}, the individual bounds on $P(Z = 1 \mid X = 1, Y = 1)$ allow this point to exist.  Next, we check whether setting of $P(Z = 1 \mid X = 1, Y = i) = p^E_{1i}$ makes the following hold
\begin{equation}\label{eq:thm-2-prood-delta-00}
\max\{0, P(Z = 1 \mid X = 1, Y = 0) - P(Z = 1 \mid X = 1, Y =0)\} = 0    
\end{equation}
From Lemma \ref{lem:delta-max-operator-0-non-0}, this is equivalent to checking  $\sum_{i = 1}^N p_{y_i}p^E_{1i} \geq p_{11} - p^E_{10}p_{y_0}$. And since $p_{11} - p_{10} < 0$, $\sum_{i = 1}^N p_{y_i}p^E_{1i} > p^T_{11} - p^E_{10}p_{y_0}$. Hence for this setting of the free parameters, Eq. \ref{eq:thm-2-prood-delta-00} would equal $0$.

However, this still requires us to ensure $\sum_{i = 1}^N p_{y_i}p^E_{1i}$ is a valid setting for the free parameter, and therefore we must check whether this is less than both the upper bounds on the weighted sum of the free parameter stated in Lemma \ref{lem:free-param-bounds}, i.e.

\begin{equation}
    \sum_{i = 1}^N p_{y_i}p^E_{1i} \leq \min\begin{bmatrix}
                                                p^T_{11}\\
                                                \frac{(1 - p_x)p_{y_0} + p^T_{11}p_x - p^E_{10}p_{y_0}}{p_x}
                                            \end{bmatrix}
\end{equation}
        
When either of the values are the upper bound, and are less than $\sum_{i = 1}^N p_{y_i}p^E_{1i}$,
Lemma \ref{lem:delta-0-max-operator-bounds} shows that there exists a setting for the free parameters such that $\max\{P(Z = 1 \mid X = 1, Y = 0) - P(Z = 1 \mid X = 0, Y = 0) = 0$. So, we set $\sum_{i = 1}^N p_{y_i}P(Z = 1 \mid X = 1, Y = i) = p^T_{11} - p^E_{10}p_{y_0}$. Since $p_{11} - p_{10} < 0$, then $p^T_{11} - p^E_{10}p_{y_0} < \sum_{i = 1}^N p^E_{1i}p_{y_i}$, this means the values of the individual free parameters can be lowered even more than $p^E_{1i}$, allowing the remaining max operators $\forall i \in \{1, \dots, N\}\; \max\{P(Z = 1 \mid X = 1, Y = i) - P(Z = 1\mid X = 0, Y = i)$ to still equal $0$ as well since they are increasing functions of $P(Z = 1 \mid X = 1, Y = i)$. This shows that in all cases, a setting of the free parameters such that all max operators evaluate to $0$ will be always be allowed. 

So, when $p^T_{11} - p^T_{10} \geq 0$, $\Delta = p^T_{11} - p^T_{10}$ and $0$ otherwise. This shows $\Delta = \max\{0, p^T_{11} - p^T_{10}\}$. 

\section{Proof of Theorem 3}
Theorem 3 provides bounds on $PNS$ for arbitrary cardinality of the support of $Y$, while parameterizing the difference in treatment assignment mechanism of $X$ by $\delta_X$. To prove Theorem 3, we first state lemmas we utilize in the proof. Proofs can be found in (\S \ref{sec:lemma-proofs}). 

\begin{lemma}\label{lemma:RREF-y-delta}
Let $X$ and $Z$ be binary random variables, and let $Y$ be a discrete random variable taking on values in $0, \dots ,N$. Assume access to two distributions $P^T(Z, X)$ and $P^E(Z, Y)$, where $P^T(Y) = P^E(Y)$, $P^E(X) = P^T(X) + \delta_X$, $X \independent Y$ in both $P^E$ and $P^T$ and $P^T(Z = 1 \mid X, Y) = P^E(Z = 1\mid X, Y)$. The conditional distributions $P(Z \mid X, Y)$ can be expressed as a system of equations with free parameters $P(Z = 1 \mid X = 1, Y = i)$ for $i \in \{1, \dots, N\}$ as

{\small
\begin{align*}
P(Z = 1 \mid X = 0, Y = 0) &=  \frac{p_x + \delta_X}{1 - p_x - \delta_X}\sum_{n = 1}^N\frac{p_{y_n}}{p_{y_0}}P^T(Z = 1 \mid X = 1, Y = n) + \frac{p^E_{10}p_{y_0} - p^T_{11}(p_x + \delta_X)}{(1 - p_x - \delta_X)p_{y_0}}\\
P(Z = 1 \mid X = 1, Y = 0)&= \frac{p^T_{11}}{p_{y_0}} - \sum_{n = 1}^N\frac{p_{y_n}}{p_{y_0}}P^T(Z = 1 \mid X = 1, Y = n)\\
P(Z = 1 \mid X = 0, Y = 1)&= \frac{p^E_{11}}{1 - p_x - \delta_X} - \frac{p_x + \delta_X}{1 - p_x - \delta_X}P^T(Z = 1 \mid X = 1, Y = 1)\\
P(Z = 1 \mid X = 0, Y = 2)&= \frac{p^E_{12}}{1 - p_x - \delta_X} - \frac{p_x + \delta_X}{1 - p_x - \delta_X}P^T(Z = 1 \mid X = 1, Y = 2)\\
\vdots\\
P(Z = 1 \mid X = 0, Y = N)&= \frac{p^E_{1N}}{1 - p_x - \delta_X} - \frac{p_x + \delta_X}{1 - p_x - \delta_X}P^T(Z = 1 \mid X = 1, Y = N)
\end{align*}
}
Where $P^T(X = 1)$ is denoted as $p_x$ , $P^T(Y = n)= P^E(Y = n)$ is denoted as $p_{y_n}$, $P^T(Z = 1 \mid X = 1)$ is denoted as $p^T_{11}$, $P^T(Z = 1 \mid X = 0)$ is denoted as $p^T_{10}$, and $P^T(Z = 1 \mid Y = n)$ is denoted as $p^E_{1n}$ for $n \in \{0, \dots, N\}$.
\end{lemma}

\begin{lemma}\label{lem:free-param-bounds-delta}
To ensure the solution to the system of equations in Lemma \ref{lemma:RREF-y-delta} form coherent probabilities, the following bounds on the free parameters must hold
{\small
\begin{align}
\max\begin{bmatrix}
   \frac{p^T_{11}(p_x + \delta_X) - p^E_{10}p_{y_0}}{p_x + \delta_X} \\
   p^T_{11} - p_{y_0}
\end{bmatrix} \leq \sum_{n = 1}^N y_nP^T(Z = 1 \mid X = 1, Y = n) 
\leq \min\begin{bmatrix}
    \frac{(1 - p_x - \delta_X)p_{y_0} + p^T_{11}(p_x + \delta_X) - p^E_{10}y_0}{p_x + \delta_X}\\
    p^T_{11}
\end{bmatrix}
\end{align}
}

And for each $i \in \{1, \dots, N\}$, we the bounds on the following free parameters $P(Z = 1\mid X = 1, Y = i)$
\begin{align}
    \max\begin{bmatrix}
        0\\
        \frac{p^E_{1i} - (1 - p_x - \delta_X)}{p_x + \delta_X}
    \end{bmatrix}
        \leq P^T(Z = 1\mid X = 1, Y = i) \leq 
    \min\begin{bmatrix}
        1\\
        \frac{p^E_{1i}}{p_x + \delta_X}
    \end{bmatrix}
\end{align}
\end{lemma}
Lemma \ref{lem:free-param-bounds-delta} can be proved using a similar approach as Lemma \ref{lem:free-param-bounds}. Similar lemmas to Lemma \ref{lem:bounds-conditions} through Lemma \ref{lem:pns-all-left-extereme} can be derived for this case, but $p_x$ will be replaced with $p_x + \delta_X$ and $1 - p_x$ with $1 - p_x - \delta_X$. Then, a similar approach to the one utilized in Theorem 2, combined with the identity $\sum_{j = 0}^{N} p^E_{1j}p_{y_j} - p^T_{10}(1 - p_x - \delta_X) - p^T_{11}(p_x + \delta_X) = 0$ (proved in Lemma \ref{lemma:RREF-y-delta}) can be used to prove this Theorem. The cases to consider will be when $p_x + \delta_X = 1 - p_x - \delta_X$, $p_x + \delta_X > 1 - p_x - \delta_X$ and $p_x + \delta_X < 1 - p_x - \delta_X$. Additionally, the indicators and bounds will be updated as well, replacing $p_x + \delta_X$ and $1 - p_x$ with $1 - p_x - \delta_X$.

\section{Proof of Theorem 4}
Theorem 4 provides bounds for the causal graph in Fig. 3(a), given access to a joint distribution over $P(Z, X< Y, C)$. 
\begin{proof}\phantom{...}{\newline}
We start by expressing bounds on the joint probability $P(Z_{\mathbf{x}} = 1, Z_{x^\prime} = 0, C, Y)$ as
{\small
\begin{equation*}
    \max\begin{bmatrix}
        0\\
        P(Z_{x} = 1, C, Y) - P(Z_{x^\prime} = 1, C, Y)
        \end{bmatrix} \leq P(Z_{x} = 1, Z_{x^\prime} = 0, C, Y)
        \leq \min \begin{bmatrix}
            P(Z_{x} = 1, C, Y)\\
            P(Z_{x^\prime} = 0, C, Y)
        \end{bmatrix}   
\end{equation*}
}
Where the upper bound follows from $P(A, B) \leq \min\{P(A), P(B)\}$ for random variables $A$ and $B$. The lower bound is proved below. 
\begin{align*}
    P(Z_{x} = 1, C, Y) - P(Z_{x^\prime} = 1, C, Y)  &= 
    P(Z_{x} = 1, Z_{\mathbf{x'}} = 0, C, Y) + P(Z_{x} = 1, Z_{\mathbf{x'}} = 1, C, Y)\\
    &- P(Z_{x} = 1, Z_{\mathbf{x'}} = 1, C, Y) - P(Z_{x} = 0, Z_{\mathbf{x'}} = 1, C, Y)\\
    &= P(Z_{x} = 1, Z_{\mathbf{x'}} = 0, C, Y) - P(Z_{x} = 0, Z_{\mathbf{x'}} = 1, C, Y)
\end{align*}

And since probabilities are non-negative, we obtain the lower bound
\[
\max \begin{bmatrix}
    0\\
    P(Z_{x} = 1, C, Y) - P(Z_{x^\prime} = 1, C, Y) 
\end{bmatrix} \leq P(Z_{x} = 1, Z_{\mathbf{x'}} = 0, C, Y)
\]
Next, the $PNS$ is obtained by marginalizing out $C$ and $Y$, giving a lower bound of
\begin{align*}
    &\sum_{C, Y}\max\begin{bmatrix}
                        0\\
                        P(Z_{x} = 1, C, Y) - P(Z_{x^\prime} = 1, C, Y)
                    \end{bmatrix} \leq P(Z_{x} = 1, Z_{x^\prime} = 0)
\end{align*}
And the upper bound is given as 
\begin{align*}
    & P(Z_{x} = 1, Z_{x^\prime} = 0) \leq \sum_{C, Y}\min \begin{bmatrix}
        P(Z_{x} = 1, C, Y)\\
        P(Z_{x^\prime} = 0, C, Y)
    \end{bmatrix}
\intertext{Since $Z_{x} \independent X$, an application of consistency allows the above bounds to be re-written as}
    &\sum_{C, Y}P(C, Y)\max\begin{bmatrix}
                                0\\
                                P(Z = 1 \mid X = 1, C, Y) - P(Z = 1 \mid X = 0, C, Y)
                            \end{bmatrix} \leq P(Z_{x} = 1, Z_{x^\prime} = 0)\\
    &P(Z_{x} = 1, Z_{x^\prime} = 0) \leq \sum_{C, Y} P(C, Y)\min
    \begin{bmatrix}
        P(Z = 1 \mid X = 1, C, Y)\\
        P(Z = 0 \mid X = 0, C, Y)
    \end{bmatrix}
\end{align*}
Exploiting the independence of $C$ and $Y$, the lower bound is written as
 \begin{align*}
    \sum_{C}P(C) \Delta_C \leq &P(Z_{x} = 1, Z_{x^\prime} = 0)\\
    \intertext{Where}
    \Delta_C = \sum_Y P(Y)\max\{0, P(Z = 1 \mid X = 1, C, Y)& - P(Z = 1 \mid X = 0, C, Y)\}\\ 
\end{align*}   
Next, the minimum operator in the upper bound can be re-written as
\begin{align*}
    &\min
    \begin{bmatrix}
        P(Z = 1 \mid X = 1, C, Y)\\
        P(Z = 0 \mid X = 0, C, Y)
    \end{bmatrix}\\
    &= P(Z = 1 \mid X = 1, C, Y) 
    - \max\begin{bmatrix}
          0\\
          P(Z = 1 \mid X = 1, C, Y) - P(Z = 0 \mid X = 0, C, Y)
          \end{bmatrix}
\end{align*}
Consequently, the upper bound is expressed as
\begin{align*}
   &P(Z_{x} = 1, Z_{x^\prime} = 0) \leq P(Z = 1\mid X = 1) - \sum_C P(C)\Gamma_C\\
   \intertext{Where}
   &\Gamma_C = \sum P(Y)\max \{0, P(Z = 1 \mid X = 1, C, Y) - P(Z = 0 \mid X = 0, C, Y) \}
\end{align*}
This concludes the proof of Theorem 4. 
\end{proof}

\section{Proof of Lemma 5}
\begin{proof}\phantom{...}{\newline}
Lemma 5 provides the conditions under which the bounds in Theorem 4 are tighter than the bounds provided in \cite{dawid2017probability}. The  bounds on PNS in Equation 2 are given as
\begin{equation}\label{eq:pns-dawid-bounds}
    \Delta \leq PNS \leq P(Z = 1 \mid X = 1) - \Gamma
\end{equation}
Where
\begin{align*}
    \Delta &= \sum_c \mathbb{P}(C = c)\max \{0, \mathbb{P}(Z = 1 \mid X = 1, C = c) - \mathbb{P}(Z = 1 \mid X = 0, C = c)\} \\
    \Gamma &= \sum_c \mathbb{P}(C = c)\max \{0, \mathbb{P}(Z = 1 \mid X = 1, C = c)
     - \mathbb{P}(Z = 0 \mid X = 0, C = c)\} 
\end{align*}

And the bounds in Lemma 5 are given as
    \begin{equation}\label{eq:trident-bounds}
        \sum_{C}P(C)\Delta_C \leq PNS \leq P(Z = 1 \mid X = 1) - \sum_{C}P(C)\Gamma_C
    \end{equation}
Where
\begin{align*}
    \Delta_C &= \sum_y \mathbb{P}(Y = y)\max \{0, \mathbb{P}(Z = 1 \mid X = 1, Y = y, C) - \mathbb{P}(Z = 1 \mid X = 0, Y = y, C)\} \\
    \Gamma_C &= \sum_y \mathbb{P}(Y = y)\max \{0, \mathbb{P}(Z = 1 \mid X = 1, Y = y, C) - \mathbb{P}(Z = 0 \mid X = 0, Y = y, C)\} 
\end{align*}

Starting with the lower bound, note that both lower bounds consist of an identical weighted sum $P(C)$, hence understanding the difference of these bounds comes down to examining $\max \{0, \mathbb{P}(Z = 1 \mid X = 1, C = c) - \mathbb{P}(Z = 1 \mid X = 0, C = c)\}$ versus $\sum_y \mathbb{P}(Y = y)\max \{0, \mathbb{P}(Z = 1 \mid X = 1, Y = y, C) - \mathbb{P}(Z = 1 \mid X = 0, Y = y, C)\}$. First, note that
\begin{align*}
&\mathbb{P}(Z = 1 \mid X = 1, C = c) - \mathbb{P}(Z = 1 \mid X = 0, C = c) =\\ &\sum_y P(Y = y)\Big(\mathbb{P}(Z = 1 \mid X = 1, Y = y, C) - \mathbb{P}(Z = 1 \mid X = 0, Y = y, C) \Big)
\end{align*}

Consequently, $\max \{0, \mathbb{P}(Z = 1 \mid X = 1, C = c) - \mathbb{P}(Z = 1 \mid X = 0, C = c)\}$ can be written as
\begin{align*}
    &\max \{0, \mathbb{P}(Z = 1 \mid X = 1, C = c) - \mathbb{P}(Z = 1 \mid X = 0, C = c)\}\\
    &= \max \Bigg\{0, \sum_y P(Y = y)\Big(\mathbb{P}(Z = 1 \mid X = 1, Y = y, C) - \mathbb{P}(Z = 1 \mid X = 0, Y = y, C) \Big)\Bigg\}
\end{align*}

And $\sum_y \mathbb{P}(Y = y)\max \{0, \mathbb{P}(Z = 1 \mid X = 1, Y = y, C) - \mathbb{P}(Z = 1 \mid X = 0, Y = y, C)\}$ is written as 
\begin{align*}
    \sum_y \max \{0, \mathbb{P}(Y = y) \Big(\mathbb{P}(Z = 1 \mid X = 1, Y = y, C) - \mathbb{P}(Z = 1 \mid X = 0, Y = y, C)\Big)\}
\end{align*}
Since the maximum of a sum is less than or equal to the sum of maximums, the lower bound in Equation \ref{eq:trident-bounds} will be greater than or equal to the lower bound in Equation \ref{eq:pns-dawid-bounds}. And when for any $C$, if for some values of $Y$, $\mathbb{P}(Z = 1 \mid X = 1, Y = y, C)$ is greater than $\mathbb{P}(Z = 1 \mid X = 0, Y = y, C)$, and form some values $\mathbb{P}(Z = 1 \mid X = 1, Y = y, C)$ is less than $\mathbb{P}(Z = 1 \mid X = 0, Y = y, C)$, than the lower bound in Equation \ref{eq:trident-bounds} will be greater than the lower bound in \ref{eq:pns-dawid-bounds}. 

A similar approach can be utilized for the upper bound as well, thereby completing the proof.
\end{proof}
\section{Proof of Theorem 6}
\begin{proof}\phantom{...}{\newline}
Theorem 6 provides bounds on the PNS when the treatment assignment mechanism for $X$ in the external dataset is confounded by a set of observed covariates $C$ or arbitrary dimensionality, and the target dataset records this $C$ as well. 
\begin{lemma}\label{lem:RREF-y-c-delta}
    Let $X$ and $Z$ be binary random variables, and let $C$ and $Y$ be discrete random variables taking values in $\{0, \dots, M\}$ and $\{0, \dots , N\}$ respectively. Let $\delta^C_X$ represent the link between the treatment assignment mechaism between the $P^E$ and $P^T$ as follows
    \[
    P^E(X = 1\mid C) = P^T(X = 1) + \delta^C_X
    \]
    For notational ease, we introduce the following terms. Let $P^T(Z = 1 \mid X = j, C = i)$ be denoted by $p^T{1ji}$ and let $P^E(Z = 1 \mid Y = j, C = i)$ be denoted as $p^E_{1ji}$. Given distributions $P^T(Z, X, C)$ and $P^E(Z, Y, C)$ obeying the following contraints
    \begin{align*}
        p^T_{10i} &= \sum_{j = 0}^N P(Z = 1\mid X = 0, Y = j, C = i)p_{y_j}\\
        p^T_{11i} &= \sum_{j = 0}^N P(Z = 1\mid X = 1, Y = j, C = i)p_{y_j}\\
        p^E_{10i} &= P(Z = 1\mid X = 0, Y = 0, C = i)(1 - p_x - \delta^i_X)\\
        &\hspace{0.2in} + P(Z = 1\mid X = 1, Y = 0, C = i)(p_x + \delta^i_X)\\
        \vdots\\
        p^E_{1Ni} &= P(Z = 1\mid X = 0, Y = N, C = i)(1 - p_x - \delta^i_X)\\
        &\hspace{0.2in} + P(Z = 1\mid X = 1, Y = N, C = i)(p_x + \delta^i_X)\\
    \end{align*}
    Then this system of equations can be expressed in terms of free parameters $P(Z = 1 \mid X = 1, Y = i, C)$ for $i \in \{1, \dots, N\}$ and all $C$ as
    {\small
    \begin{align*}
    \mathbb{P}(Z = 1 \mid X = 0, Y = 0, C) &=  \frac{p_x + \delta^C_X}{1 - p_x - \delta^C_X}\sum_{n = 1}^N\frac{p_{y_n}}{p_{y_0}}\mathbb{P}(Z = 1 \mid X = 1, Y = n, C) + \frac{p^E_{10C}p_{y_0} - p^T_{11C}(p_x +\delta^C_X)}{(1 - p_x - \delta^C_X)p_{y_0}}\\
    \mathbb{P}(Z = 1 \mid X = 1, Y = 0, C)&= \frac{p^T_{11C}}{p_{y_0}} - \sum_{n = 1}^N\frac{p_{y_n}}{p_{y_0}}\mathbb{P}(Z = 1 \mid X = 1, Y = n, C)\\
    \mathbb{P}(Z = 1 \mid X = 0, Y = 1, C)&= \frac{p^E_{11C}}{1 - p_x - \delta^C_X} - \frac{p_x + \delta^C_X}{1 - p_x - \delta^C_X}\mathbb{P}(Z = 1 \mid X = 1, Y = 1, C)\\
    \mathbb{P}(Z = 1 \mid X = 0, Y = 2, C)&= \frac{p^E_{12C}}{1 - p_x - \delta^C_X} - \frac{p_x + \delta^C_X}{1 - p_x - \delta^C_X}\mathbb{P}(Z = 1 \mid X = 1, Y = 2, C)\\
    \vdots\\
    \mathbb{P}(Z = 1 \mid X = 0, Y = N, C)&= \frac{p^E_{1N}}{1 - p_x - \delta^C_X} - \frac{p_x + \delta^C_X}{1 - p_x - \delta^C_X}\mathbb{P}(Z = 1 \mid X = 1, Y = N, C)\\
    \end{align*}
    }
\end{lemma}

To prove this Lemma, we apply a similar proof to the proof for Theorem 3, but apply it for every level of $C$, noting that in the case we consider, the minimum of the sum of independently varying quantities equals the sum of the minimum, and a similar result holds for the maximum. Specifically, following a similar proof as Theorem 3,
\[
    \sum_{Y}P(Y)\max\{0, P(Z = 1 \mid X = 1, C, Y) - P(Z = 1 \mid X = 0, C, Y)\} = \max\{0, P(Z = 1 \mid X = 1, C) - P(Z = 1 \mid X = 0, C)\}
\]

Similarly, the upper bound in Theorem 4 can be similarly derived by writing the upper bound as
\[
\sum_C P(C)(P(Z = 1\mid X = 1, C) - \Gamma_C)
\]
And then applying a similar proof used in Theorem 2 for each level of $C$ to obtain the upper bound. 
\end{proof}

\section{Proofs For Auxiliary Lemmas}\label{sec:lemma-proofs}

\begin{proof}[\textbf{Proof for Lemma \ref{lemma:RREF-y}}] This lemma is proved using mathematical induction performed on the the Reduced Row Echelon Form (RREF) of the matrix representation of the system of equations that in Equation X that impose constraints on $P(Z \mid X , Y)$ using $P^T(Z, X)$ and $P^E(Z, Y)$. 

Using the invariances defined in Section X, we obtain the following equations for $P(Z = 1 \mid X, Y)$ in terms of $P^T(Z, X)$ and $P^T(Z, Y)$. For ease of notation, we introduce the following terms. $P^T(X = 1)$ is denoted as $p_x$ , $P^T(Y = n)$ is denoted as $p_{y_n}$, $P^T(Z = 1 \mid X = 1)$ is denoted as $p^T_{11}$, $P^T(Z = 1 \mid X = 0)$ is denoted as $p^T_{10}$, and $P^E(Z = 1 \mid Y = n)$ is denoted as $p^E_{1n}$ for $n \in \{0, \dots, N\}$.

\begin{align*}
    p^T_{10} &= \sum_{i = 0}^N P(Z = 1\mid X = 0, Y = n)p_{y_n}\\
    p^T_{11} &= \sum_{n = 0}^N P(Z = 1\mid X = 1, Y = n)p_{y_n}\\
    p^E_{10} &= P(Z = 1\mid X = 0, Y = 0)(1 - p_x) + P(Z = 1\mid X = 1, Y = 0)p_x\\
    \vdots\\
    p^E_{1N} &= P(Z = 1\mid X = 0, Y = N)(1 - p_x) + P(Z = 1\mid X = 1, Y = N)p_x\\
\end{align*}

The above equations can be represented in matrix form $A\mathbf{x} = b$ as
{\scriptsize
\begin{align*}
A &= \begin{bmatrix}
    p_{y_0} & 0 & p_{y_1} & 0 & \dots & p_{y_{N - 1}} & 0 & p_{y_N} & 0\\
    0 & p_{y_0} & 0 & p_{y_1} & \dots & 0 & p_{y_{N - 1}}& 0 & p_{y_N}\\
    1 - p_{x} & p_{x} & 0 & 0 & \dots & 0 & 0 & 0 & 0\\
    0 & 0 & 1 - p_x & p_x & \dots & 0 & 0 & 0 & 0\\
    \vdots & \ddots & & & & & & & \vdots\\
    0 & 0 & \dots & 0 & 0 & 1 - p_x & p_x & 0 & 0\\
    0 & 0 & \dots & 0 & & 0 & 0 & 1 - p_x & p_x\\
\end{bmatrix}\\
b &= \begin{bmatrix}
    p^T_{10}\\
    p^T_{11}\\
    p^E_{10}\\
    p^E_{11}\\
    p^E_{12}
    \vdots\\
    p^E_{1(N - 1)}\\
    p^E_{1N}\\
\end{bmatrix}
\hspace{0.5in}\mathbf{x} 
= \begin{bmatrix}
P(Z = 1 \mid X = 0, Y = 0)\\    
P(Z = 1 \mid X = 1, Y = 0)\\
P(Z = 1 \mid X = 0, Y = 1)\\    
P(Z = 1 \mid X = 1, Y = 1)\\
\vdots\\
P(Z = 1 \mid X = 0, Y = N)\\    
P(Z = 1 \mid X = 1, Y = N)\\
\end{bmatrix}
\end{align*}
}
Next, denote the augmented matrix $A | b$ as 
\begin{align*}
    \begin{bmatrix}[ccccccccc|c]
    p_{y_0} & 0 & p_{y_1} & 0 & \dots & p_{y_{N - 1}} & 0 & p_{y_N} & 0 & p^T_{10}\\
    0 & p_{y_0} & 0 & p_{y_1} & \dots & 0 & p_{y_{N - 1}}& 0 & p_{y_N} & p^T_{11}\\
    1 - p_x & p_x & 0 & 0 & \dots & 0 & 0 & 0 & 0 & p^E_{10}\\
    0 & 0 & 1 - p_x & p_x & \dots & 0 & 0 & 0 & 0 & p^E_{11}\\
    \vdots & \ddots & & & & & & & \vdots & \vdots \\
    0 & 0 & \dots & 0 & 0 & 1 - p_x & p_x & 0 & 0 & p^E_{1(N - 1)}\\
    0 & 0 & \dots & 0 & & 0 & 0 & 1 - p_x & p_x & p^E_{1N}\\
    \end{bmatrix}
\end{align*}

The augmented matrix $A | b$ when expressed in RREF will express the system of equations in terms of basic variables and free parameters. Then, this can be compared to the system of equations presented in the Lemma to complete the proof.  

First, we utilize mathematical induction to provide a general form for the RREF of $A|b$ for any $N$. Formally, let $P(N)$ denote the following proposition. \\
\textbf{Proposition}: Let $P(N)$ denote the proposition that when $X$ and $Z$ are binary, and $Y$ has support in $\{ 0, \dots, N\}$, then $A|b$ will be a matrix with $N + 3$ rows and $2(N + 1) + 1$ columns. The following steps when executed will put $A|b$ in RREF, where $R_i$ denotes the $i$-th row of $A|b$

\begin{enumerate}
    \item 
        \begin{align}
        R_1 &= R_1/p_{y_0}\\
        R_3 &= R_3 - (1 - p_x)R_1
        \end{align}\label{step-1}
    \item
        \begin{align}
        R_2 &= R_2/p_{y_0}\\
        R_3 &= R_3 - p_xR_2
        \end{align}\label{step-2}
    \item 
        \begin{align}
        R_3 &= R3/(-\frac{(1 - p_x)p_{y_1}}{p_{y_0}})\\
        R_1 &= R_1 - \frac{p_{y_1}}{p_{y_0}}R3\\
        R_4 &= R4 - (1 - p_x)R3
        \end{align}\label{step-3}
    \item If $N \geq 2$, then loop through the following steps for $i \in \{2, \dots, N\}$:
    \begin{align}
        R_{2 + i} = R_{2 + i}/(-\frac{p_{y_i}(1 - p_x)}{p_{y_{i - 1}}})\\
        R_{1 + i} = R_{1 + i} - \frac{p_{y_{i}}}{p_{y_{i - 1}}}R_{2 + i}\\
        R_{3 + i} = R_{3 + i} - (1 - p_x)R_{2 + i}
    \end{align}\label{step-n}
\end{enumerate}

And the corresponding matrix will evaluate to
\begin{align}\label{matrix:rref} 
    \begin{bmatrix}
    1 &  0 &  0 &  -\frac{p_xp_{y_1}}{(1 - p_x)p_{y_0}} & \dots & 0 & -\frac{p_xp_{y_{N - 1}}}{(1 - p_x)p_{y_0}} & 0 & -\frac{p_xp_{y_N}}{(1 - p_x)p_{y_0}} & \frac{p^E_{10}p_{y_0} - p^T_{11}p_x}{(1 - p_x)p_{y_0}}\\
    0 &  1 &  0 & \frac{p_{y_1}}{p_{y_0}} & \dots & 0 & \frac{p_{y_{N - 1}}}{p_{y_0}} & 0 & \frac{p_{y_N}}{p_{y_0}} & \frac{p^T_{11}}{p_{y_0}}\\
    0 &  0 &  1 & \frac{p_x}{1 - p_x} &  \dots & 0 & 0 & 0 &  0 & \frac{p^E_{11}}{1 - p_x}\\
    \vdots &&&&&&& \ddots \\
    0 &  0 &  0 & 0 & \dots & 0 & 0 & 1 & \frac{p_x}{1 - p_x} & \frac{p^T_{10}(1 - p_x) + p^T_{11}p_x - \sum_{i = 0}^{N - 1} p^E_{1i}p_{y_i}}{p_{y_N}(1 - p_x)}\\
    0 &  0 &  0 & 0 & \dots & 0 & 0 & 0 & 0 & \frac{\sum_{i = 0}^{N} p^E_{1i}p_{y_i} - p^T_{10}(1 - p_x) - p^T_{11}p_x}{p_{y_N}} 
    \end{bmatrix}
\end{align}
\newline
\textbf{Base Case}: Starting with the base case of $N = 1$, i.e. $Y$ is binary. For the case when $Y$ is binary, we can enumerate the following equations for the $P(Z, X, Y)$
{\small
\begin{align*}
    \mathbb{P}(Z = 1 \mid X = 1) &= \mathbb{P}(Z = 1 \mid X = 1, Y = 1)\mathbb{P}(Y = 1) + \mathbb{P}(Z = 1 \mid X = 1, Y = 0)\mathbb{P}(Y = 0)\\
    \mathbb{P}(Z = 1 \mid X = 0) &= \mathbb{P}(Z = 1 \mid X = 0, Y = 1)\mathbb{P}(Y = 1) + \mathbb{P}(Z = 1 \mid X = 0, Y = 0)\mathbb{P}(Y = 0)\\
    \mathbb{P}(Z = 1 \mid Y = 1) &= \mathbb{P}(Z = 1 \mid X = 0, Y = 1)\mathbb{P}(X = 0) + \mathbb{P}(Z = 1 \mid X = 1, Y = 1)\mathbb{P}(X = 1)\\
    \mathbb{P}(Z = 1 \mid Y = 0) &= \mathbb{P}(Z = 1 \mid X = 0, Y = 0)\mathbb{P}(X = 0) + \mathbb{P}(Z = 1 \mid X = 1, Y = 0)\mathbb{P}(X = 1)
\end{align*}
}
And the augmented matrix $A|b$ will be written as
\begin{align*}
A|b &= \begin{bmatrix}
    p_{y_0} & 0 & p_{y_1} & 0 & p^T_{10}\\
    0 & p_{y_0} & 0 & p_{y_1} & p^T_{11}\\
    1 - p_x & p_x & 0 & 0 & p^E_{10}\\
    0 & 0 & 1 - p_x & p_x & p^E_{11}
\end{bmatrix}
\hspace{0.25in}\mathbf{x} 
= \begin{bmatrix}
P(Z = 1 \mid X = 0, Y = 0)\\    
P(Z = 1 \mid X = 1, Y = 0)\\
P(Z = 1 \mid X = 0, Y = 1)\\    
P(Z = 1 \mid X = 1, Y = 1)\\
\end{bmatrix}
\end{align*}

Executing the row operations on $A|b$ gives the following matrix
\begin{align*}
\begin{bmatrix}
    1 & 0 & 0 & -\frac{p_xp_{y_1}}{(1 - p_x)p_{y_0}} & \frac{p^E_{10}p_{y_0} - p^T_{11}p_x}{(1 - p_x)p_{y_0}}\\
    0 & 1 & 0 & \frac{p_{y_1}}{p_{y_0}} & \frac{p^T_{11}}{p_{y_0}}\\
    0 & 0 & 1 & \frac{p_x}{1 - p_x} & a\\ 
    0 & 0 & 0 & 0 & b
\end{bmatrix}    
\end{align*}
Where $a$ is:
\begin{align*}
    a &= \frac{p^T_{10}(1 - p_x) + p^T_{11}p_x - p^E_{10}p_{y_0}}{p_{y_1}(1 - p_x)}\\
\end{align*}

And $b$ is 
\begin{align*}
    b &= \frac{-p^T_{10}(1 - p_x) - p^T_{11}p_x + p^E_{10}p_{y_0} + p^E_{11}p_{y_1}}{p_{y_1}}\\
\end{align*}
This proves that the $A|b$ for $N = 1$ satisfies $P(1)$, proving the base case. \\
\newline
\textbf{Induction Step}: Next, we prove that if $P(N - 1)$ holds, then $P(N)$ holds as well. For $P(N)$, $A|b$ will look like
{\scriptsize
\begin{align*}
\begin{bmatrix}
    p_{y_0} & 0 & p_{y_1} & 0 & \dots & p_{y_{N - 1}} & 0 & y_N & 0 & p^T_{10}\\
    0 & p_{y_0} & 0 & p_{y_1} & \dots & 0 & p_{y_{N - 1}}& 0 & p_{y_N} & p^T_{11}\\
    1 - p_x & p_x & 0 & 0 & \dots & 0 & 0 & 0 & 0 & p^E_{10}\\
    0 & 0 & 1 - p_x & p_x & \dots & 0 & 0 & 0 & 0 & p^E_{11}\\
    \vdots & \ddots & & & & & & & & \vdots \\
    0 & 0 & 0 & 0 & \dots & 1 - p_x & p_x & 0 & 0 & p^E_{1(N - 1)}\\
    0 & 0 & 0 & 0 & \dots & 0 & 0 & 1 - p_x & p_x & p^E_{1N}\\
\end{bmatrix}    
\end{align*}
}

Now, since we assume $P(N - 1)$ holds, the submatrix of $A|b$ corresponding to $P(N - 1)$ can be put into the assumed RREF form. However, since we added two extra columns to $A \mid b$ the $P(N)$ compared to $a \mid b$ for $P(N- 1)$,  all of the row operations performed on the submatrix corresponding to $P(N - 1)$ must be performed on these added columns as well. None of these row operations apply to the final row, since it is newly added and therefore not touched by RREF computations for $P(N - 1)$. 

The row operations must be applied to the newly added columns, and we calculate these now. We use $c_N$ to denote the matrix of the newly added columns, and track the results on $c_N$ of the row operations applied to $P(N - 1)$.\\
Applying the row operations in \ref{step-1} to $c_N$
\begin{align*}
    c_N = \begin{bmatrix}
        \frac{p_{y_N}}{p_{y_0}} & 0\\
        0 & p_{y_N}\\
        -\frac{(1 - p_x)p_{y_N}}{p_{y_0}} & 0\\
        0 & 0\\
        \vdots \\
        1 - p_x & p_x\\
    \end{bmatrix}
\end{align*}
Next, applying the operations in \ref{step-2} to $c_N$
\begin{align*}
    c_N = \begin{bmatrix}
        \frac{p_{y_N}}{p_{y_0}} & 0\\
        0 & \frac{p_{y_N}}{p_{y_0}}\\
        -\frac{(1 - x)p_{y_N}}{p_{y_0}} & -\frac{p_x p_{y_N}}{p_{y_0}}\\
        0 & 0\\
        \vdots \\
        1 - p_x & p_x\\
    \end{bmatrix}
\end{align*}

Next, applying the operations in \ref{step-3} to $c_N$
\begin{align*}
    c_N = \begin{bmatrix}
        0 & -\frac{p_x p_{y_N}}{(1 - p_x)p_{y_0}}\\
        0 & \frac{p_{y_N}}{p_{y_0}}\\
        \frac{p_{y_N}}{p_{y_1}} & \frac{p_x p_{y_N}}{(1 - p_x)p_{y_1}}\\
        -\frac{p_{y_N}(1 - p_x)}{p_{y_1}} & -\frac{p_xp_{y_N}}{p_{y_1}}\\
        \vdots \\
        1 -p_ x & p_x\\
    \end{bmatrix}
\end{align*}

Next, applying the operations in \ref{step-n} to $c_N$ for $i = 2$, we get
\begin{align*}
    c_N = \begin{bmatrix}
        0 & -\frac{p_xp_{y_N}}{(1 - p_x)p_{y_0}}\\
        0 & \frac{p_{y_N}}{p_{y_0}}\\
        0 & 0\\
        \frac{p_{y_N}}{p_{y_2}} & \frac{p_x p_{y_N}}{(1 - p_x)p_{y_2}}\\
        -\frac{(1 - p_x)p_{y_N}}{p_{y_2}} & -\frac{p_x p_{y_N}}{p_{y_2}}\\
        0 & 0\\
        \vdots \\
        1 - p_x & p_x\\
    \end{bmatrix}
\end{align*}

As we repeat the row operations in \ref{step-n} for  $i \in \{3, \dots, N - 1\}$, $c_N$ will ultimately become
\begin{align*}
    c_N = \begin{bmatrix}
                0 & -\frac{xp_{y_N}}{(1 - p_x)p_{y_0}}\\
                0 & \frac{p_{y_N}}{p_{y_0}}\\
                0 & 0\\
                0 & 0\\
                \vdots \\
                0 & 0\\
                \frac{p_{y_N}}{p_{y_{N - 1}}} & \frac{p_x p_{y_N}}{p_{y_{N - 1}}(1 - p_x)}\\
                -\frac{(1 - p_x)p_{y_N}}{p_{y_{N - 1}}} & -\frac{p_x p_{y_N}}{p_{y_{N - 1}}}\\
                1 - p_x & p_x\\
    \end{bmatrix}
\end{align*}

Consequently, the matrix $A \mid b$ corresponding to $P(N)$ will contain a submatrix corresponding to the RREF in $P(N - 1)$, with the resulting $c_N$ and extra row appended. Formally, it will look like the matrix presented below. 
{\tiny
\begin{align}\label{rref}
    \begin{bmatrix}[ccccccccc|c]
    1 &  0 &  0 &  -\frac{p_xp_{y_1}}{(1 - p_x)p_{y_0}} & \dots & 0 & -\frac{p_xp_{y_{N - 1}}}{(1 - p_x)p_{y_0}} & 0 & -\frac{p_xp_{y_N}}{(1 - p_x)p_{y_0}} & \frac{p^E_{10}p_{y_0} - p^T_{11}p_x}{(1 - p_x)p_{y_0}}\\
    0 &  1 &  0 & \frac{p_{y_1}}{p_{y_0}} & \dots & 0 & \frac{p_{y_{N - 1}}}{p_{y_0}} & 0 & \frac{p_{y_N}}{p_{y_0}} & \frac{p^T_{11}}{p_{y_0}}\\
    0 &  0 &  1 & \frac{p_x}{1 - p_x} &  \dots & 0 & 0 & 0 &  0 & \frac{p^E_{11}}{1 - p_x}\\
    \vdots &&&&&&& \ddots & &\vdots\\
    0 &  0 &  0 & 0 & \dots & 1 & \frac{p_x}{1 - p_x} & \frac{p_{y_N}}{p_{y_{N - 1}}}& \frac{p_xp_{y_N}}{p_{y_{N - 1}}(1 - p_x)} & \frac{p^T_{10}(1 - p_x) + p^T_{11}p_x - \sum_{i = 0}^{N - 2} p^E_{1i}p_{y_i}}{p_{y_{N - 1}}(1 - p_x)}\\
    0 &  0 &  0 & 0 & \dots & 0 & 0 & -\frac{(1 - p_x)p_{y_N}}{p_{y_{N - 1}}} & -\frac{p_x p_{y_N}}{p_{y_{N - 1}}} & \frac{\sum_{i = 0}^{N - 1} p^E_{1i}p_{y_i} - p^T_{10}(1 - p_x) - p^T_{11}p_x}{p_{y_{N - 1}}}\\
    0 &  0 &  0 & 0 & \dots & 0 & 0 & 1 - x & x & p^E_{1N}
    \end{bmatrix}
\end{align}
}
This matrix is not in RREF yet, so we perform the following row operations to put the matrix in RREF
\begin{align*}
    R_{N + 2} = \frac{R_{N + 2}}{-\frac{(1 - p_x)p_{y_N}}{p_{y_{N - 1}}}}\\
    R_{N + 1} = R_{N + 1} - \frac{p_{y_N}}{p_{y_{N - 1}}}R_{N + 2}\\
    R_{N + 3} = R_{N + 3} - (1 - p_x)R_{N + 2}
\end{align*}

Resulting in the following matrix:
{\tiny
\begin{align}\label{matrix:induction-case-rref}
    \begin{bmatrix}[ccccccccc|c]
    1 &  0 &  0 &  -\frac{p_xp_{y_1}}{(1 - p_x)p_{y_0}} & \dots & 0 & -\frac{p_xp_{y_{N - 1}}}{(1 - p_x)p_{y_0}} & 0 & -\frac{[_xp_{y_N}}{(1 - p_x)p_{y_0}} & \frac{p^E_{10}p_{y_0} - p^T_{11}p_x}{(1 - p_x)p_{y_0}}\\
    0 &  1 &  0 & \frac{p_{y_1}}{p_{y_0}} & \dots & 0 & \frac{p_{y_{N - 1}}}{p_{y_0}} & 0 & \frac{p_{y_N}}{p_{y_0}} & \frac{p^T_{11}}{p_{y_0}}\\
    0 &  0 &  1 & \frac{p_x}{1 - p_x} &  \dots & 0 & 0 & 0 &  0 & \frac{p^E_{11}}{1 - p_x}\\
    \vdots &&&&&&& \ddots & &\vdots\\
    0 &  0 &  0 & 0 & \dots & 1 & \frac{p_x}{1 - p_x} & 0 & 0 & \frac{p^E_{1(N - 1)}}{1 - p_x}\\
    0 &  0 &  0 & 0 & \dots & 0 & 0 & 1 & \frac{p_x}{1 - p_x} & \frac{p^T_{10}(1 - p_x) + p^T_{11}p_x - \sum_{i = 0}^{N - 1} p^E_{1i}p_{y_i}}{p_{y_N}(1 - p_x)}\\
    0 &  0 &  0 & 0 & \dots & 0 & 0 & 0 & 0 & p^E_{1N} - \frac{p^T_{10}(1 - p_x) + p^T_{11}p_x - \sum_{i = 0}^{N - 1}p^E_{1i}p_{y_i}}{p_{y_N}}
    \end{bmatrix}
\end{align}
}

And by comparison, this matrix is equal to the matrix stated in our proposition. So, if $P(N - 1)$, then $P(N)$ holds, and this concludes the induction proof. 

Finally, since $P(Z, X)$ and $P(Z, Y)$ are marginalized distributions obtained from marginalizing out $Y$ and $X$ from $P(Z, X, Y)$ respectively, they must agree on $P(Z)$. This gives the identity
\[
p^T_{11}p_x + p^T_{10}(1 - p_x) - \sum_{i = 0}^N p^E_{1i}p_{y_i} = 0
\]
And substituting this identity into the RREF form of $A | b$, we obtain
{\tiny
\begin{align}\label{rref}
    \begin{bmatrix}[ccccccccc|c]
    1 &  0 &  0 &  -\frac{p_xp_{y_1}}{(1 - p_x)p_{y_0}} & \dots & 0 & -\frac{p_xp_{y_{N - 1}}}{(1 - p_x)p_{y_0}} & 0 & -\frac{p_xp_{y_N}}{(1 - p_x)p_{y_0}} & \frac{p^E_{10}p_{y_0} - p^T_{11}p_x}{(1 - p_x)p_{y_0}}\\
    0 &  1 &  0 & \frac{p_{y_1}}{p_{y_0}} & \dots & 0 & \frac{p_{y_{N - 1}}}{p_{y_0}} & 0 & \frac{p_{y_N}}{p_{y_0}} & \frac{p^T_{11}}{p_{y_0}}\\
    0 &  0 &  1 & \frac{p_x}{1 - p_x} &  \dots & 0 & 0 & 0 &  0 & \frac{p^E_{11}}{1 - p_x}\\
    \vdots &&&&&&& \ddots & &\vdots\\
    0 &  0 &  0 & 0 & \dots & 1 & \frac{p_x}{1 - p_x} & 0 & 0 & \frac{p^E_{1(N - 1)}}{1 - p_x}\\
    0 &  0 &  0 & 0 & \dots & 0 & 0 & 1 & \frac{p_x}{1 - p_x} & \frac{p^E_{1N}}{1 - p_x}\\
    0 &  0 &  0 & 0 & \dots & 0 & 0 & 0 & 0 & 0
    \end{bmatrix}
\end{align}
}
Expressing this matrix in the form of its corresponding equations completes the proof for this Lemma.
\end{proof}

\begin{proof}[\textbf{Proof of Lemma \ref{lem:free-param-bounds}}]
Using the result presented in Lemma \ref{lemma:RREF-y}, when $Y$ has support $\{0, \dots, N\}$ and we are given marginals $P^T(Z, X)$ and $P^T(Z, Y)$, we can write the system of equations using free parameters as   

{\small
\begin{align*}
\mathbb{P}(Z = 1 \mid X = 0, Y = 0) &=  \frac{p_x}{1 - p_x}\sum_{n = 1}^N\frac{p_{y_n}}{p_{y_0}}\mathbb{P}(Z = 1 \mid X = 1, Y = n) + \frac{p^E_{10}p_{y_0} - p^T_{11}p_x}{(1 - p_x)p_{y_0}}\\
\mathbb{P}(Z = 1 \mid X = 1, Y = 0)&= \frac{p^T_{11}}{p_{y_0}} - \sum_{n = 1}^N\frac{p_{y_n}}{p_{y_0}}\mathbb{P}(Z = 1 \mid X = 1, Y = n)\\
\mathbb{P}(Z = 1 \mid X = 0, Y = 1)&= \frac{p^E_{11}}{1 - p_x} - \frac{p_x}{1 - p_x}\mathbb{P}(Z = 1 \mid X = 1, Y = 1)\\
\mathbb{P}(Z = 1 \mid X = 0, Y = 2)&= \frac{p^E_{12}}{ 1 - p_x} - \frac{p_x}{1 - p_x}\mathbb{P}(Z = 1 \mid X = 1, Y = 2)\\
\vdots\\
\mathbb{P}(Z = 1 \mid X = 0, Y = N)&= \frac{p^E_{1N}}{ 1 - p_x} - \frac{p_x}{1 - p_x}\mathbb{P}(Z = 1 \mid X = 1, Y = N)\\
\end{align*}
}

Now, each of the basic variables in the system above must be a valid probability, i.e. between zero and one, inclusive. Employing these constraints on the free parameters gives
\begin{align*}
    0 \leq P(Z = 1 \mid X = 0, Y = 0) \leq 1\\
    \implies 0 \leq  \frac{p_x}{1 - p_x}\sum_{n = 1}^N\frac{p_{y_n}}{p_{y_0}}\mathbb{P}(Z = 1 \mid X = 1, Y = n) + \frac{p^E_{10}p_{y_0} - p^T_{11}p_x}{(1 - p_x)p_{y_0}} \leq 1
\end{align*}

Solving for the lower bound first, we have
\begin{align*}
    0 \leq  \frac{p_x}{1 - p_x}\sum_{n = 1}^N\frac{p_{y_n}}{p_{y_0}}\mathbb{P}(Z = 1 \mid X = 1, Y = n) + \frac{p^E_{10}p_{y_0} - p^T_{11}p_x}{(1 - p_x)p_{y_0}}\\
    \implies \frac{p^T_{11}p_x - p^E_{10}p_{y_0}}{p_x} \leq \sum_{n = 1}^N P(Z = 1\mid X = 1, Y = n)
\end{align*}
Next, with the upper bound
\begin{align*}
    \frac{p_x}{1 - p_x}\sum_{n = 1}^N\frac{p_{y_n}}{p_{y_0}}\mathbb{P}(Z = 1 \mid X = 1, Y = n) + \frac{p^E_{10}p_{y_0} - p^T_{11}p_x}{(1 - p_x)p_{y_0}} \leq 1\\
    \implies x\sum_{n = 1}^N p_{y_n}\mathbb{P}(Z = 1 \mid X = 1, Y = n) + p^E_{10}p_{y_0} - p^T_{11}p_x \leq (1 - p_x)p_{y_0}\\
    \implies \sum_{n = 1}^Ny_n\mathbb{P}(Z = 1 \mid X = 1, Y = n) \leq \frac{(1 - p_x)p_{y_0} + p^T_{11}p_x - p^E_{10}p_{y_0}}{p_x}
\end{align*}

Following a similar calculation for $P(Z = 1\mid X = 1, Y = 0)$, we have
\begin{align*}
    p^T_{11} - p_{y_0} \leq \sum_{n = 1}^Ny_n\mathbb{P}(Z = 1 \mid X = 1, Y = n) \leq p^T_{11}
\end{align*}

Since all of the bounds need to hold simultaneously, we can re-express them using using maximum and minimum operators as
\begin{align*}
\max\begin{bmatrix}
   \frac{p^T_{11}p_x - p^E_{10}p_{y_0}}{p_x} \\
   p^T_{11} - p_{y_0}
\end{bmatrix} \leq \sum_{n = 1}^N y_n\mathbb{P}(Z = 1 \mid X = 1, Y = n) 
\leq \min\begin{bmatrix}
    \frac{(1 - p_x)p_{y_0} + p^T_{11}p_x - p^E_{10}p_{y_0}}{p_x}\\
    p^T_{11}
\end{bmatrix}
\end{align*}

Next, for $i \in \{1, \dots, N\}$, we have the following constraints on the basic variables $P(Z = 1\mid X = 0, Y = i)$
\begin{align*}
    0 \leq P(Z = 1 \mid X = 0, Y = i) \leq 1\\
    0 \leq \frac{p^E_{11}}{ 1 - p_x} - \frac{p_x}{1 - p_x}\mathbb{P}(Z = 1 \mid X = 1, Y = 1) \leq 1\\
    \implies \frac{p^E_{1i} - (1 - p_x)}{p_x} \leq P(Z = 1\mid X = 1, Y = i) \leq \frac{p^E_{1i}}{p_x} 
\end{align*}

And since the free parameters are probabilities themselves, they must be between $0$ and $1$ and hence the bounds for $P(Z = 1 \mid X = 1, Y = i)$ become
\begin{align*}
    \max\begin{bmatrix}
        0\\
        \frac{p^E_{1i} - (1 - p_x)}{p_x}
    \end{bmatrix}
        \leq P(Z = 1\mid X = 1, Y = i) \leq 
    \min\begin{bmatrix}
        1\\
        \frac{p^E_{1i}}{p_x}
    \end{bmatrix}
\end{align*}
\end{proof}

\begin{proof}[\textbf{Proof for Lemma \ref{lem:bounds-conditions}}]
Each of the implications is proved using a proof by contra-positive. Starting with $p^E_{10} < p_x \implies \frac{p^T_{11}p_x - p^E_{10}p_{y_0}}{p_x} > p^T_{11} - p_{y_0}$, the contra-positive of this statement is
    \[
    \frac{p^T_{11}p_x - p^E_{10}p_{y_0}}{p_x} \leq p^T_{11} - p_{y_0} \implies p^E_{10} \geq p_x 
    \]
    To see this holds, note that
    \begin{align*}
        \frac{p^T_{11}p_x - p^E_{10}p_{y_0}}{p_x} \leq p^T_{11} - p_{y_0}\\
        \implies -\frac{p^E_{10}p_{y_0}}{p_x} \leq  - p_{y_0}\\
        \implies p^E_{10} \leq p_x
    \end{align*}
This proves $p^E_{10} < p_x \implies \frac{p^T_{11}p_x - p^E_{10}p_{y_0}}{p_x} > p^T_{11} - p_{y_0}$.\\
Next, for $p^E_{10} \geq p_x \implies \frac{p^T_{11}p_x - p^E_{10}p_{y_0}}{p_x} \leq p^T_{11} - p_{y_0}$, following a similar proof strategy, the contra-positive can be written as
\[
\frac{p^T_{11}p_x - p^E_{10}p_{y_0}}{p_x} > p^T_{11} - p_{y_0} \implies p^E_{10} < p_x
\]
To see this holds, note that
\begin{align*}
    \frac{p^T_{11}p_x - p^E_{10}p_{y_0}}{p_x} > p^T_{11} - p_{y_0}\\
    \implies -\frac{p^E_{10}p_{y_0}}{p_x} > - p_{y_0}\\
    \implies p^E_{10} < p_x
\end{align*}
This proves $p^E_{10} \geq p_x \implies \frac{p^T_{11}p_x - p^E_{10}p_{y_0}}{p_x} \leq p^T_{11} - p_{y_0}$.\\
Next, for $p^E_{10} < 1 - p_x \implies p^T_{11} < \frac{(1 - p_x)p_{y_0} + p^T_{11}p_x - p^E_{10}p_{y_0}}{p_x}$, the contra-positive is written as
\[
p^T_{11} \geq \frac{(1 - p_x)p_{y_0} + p^T_{11}p_x - p^E_{10}p_{y_0}}{p_x}  \implies p^E_{10} \geq 1 - p_x
\]
To see this holds, note that
\begin{align*}
    &p^T_{11} \geq \frac{(1 - p_x)p_{y_0} + p^T_{11}p_x - p^E_{10}p_{y_0}}{p_x}\\
    &\implies 0 \geq (1 - p_x)p_{y_0} - p^E_{10}p_{y_0}\\
    &\implies  p^E_{10} \geq 1 - p_x
\end{align*}
This proves $p^E_{10} < 1 - p_x \implies p^T_{11} < \frac{(1 - p_x)p_{y_0} + p^T_{11}p_x - p^E_{10}p_{y_0}}{p_x}$.\\
Next, for $p^E_{10} \geq 1 - p_x \implies p^T_{11} \geq \frac{(1 - p_x)p_{y_0} + p^T_{11}p_x - p^E_{10}p_{y_0}}{p_x}$ the contra-positive is written as
\[
p^T_{11} < \frac{(1 - p_x)p_{y_0} + p^T_{11}p_x - p^E_{10}p_{y_0}}{p_x}  \implies p^E_{10} < 1 - p_x
\]
To see this holds, note that
\begin{align*}
    &p^T_{11} < \frac{(1 - p_x)p_{y_0} + p^T_{11}p_x - p^E_{10}p_{y_0}}{p_x}\\
    &\implies 0 < (1 - p_x)p_{y_0} - p^E_{10}p_{y_0}\\
    &\implies p^E_{10} < 1 - p_x
\end{align*}
This concludes the proof.
\end{proof}

The proofs for Lemma \ref{lem:inidiv-bounds-conditions} follow from the properties of probability and Lemma \ref{lem:joint-bounds} can be found in \citet{dawid2021bounding}.

\begin{proof}[\textbf{Proof for Lemma \ref{lem:max-operator-expressions}}] Using the system of equations obtained in Lemma \ref{lemma:RREF-y}, each term in $\Gamma$ can be expressed using these free parameters. Starting with
    $\max \{0, \mathbb{P}(Z = 1 \mid X = 1, Y = 0) - \mathbb{P}(Z = 0 \mid X = 0, Y = 0)\}$. First, expressing each of these terms using the free parameters, we get
    \begin{align*}
        P(Z = 1 \mid X = 1, Y = 0) &= \frac{p^T_{11}}{p_{y_0}} - \sum_{n = 1}^N\frac{p_{y_n}}{p_{y_0}}\mathbb{P}(Z = 1 \mid X = 1, Y = n)\\
        P(Z = 0 \mid X = 0, Y = 0) &= 1 - \Bigg(\frac{p_x}{1 - p_x}\sum_{n = 1}^N\frac{p_{y_n}}{p_{y_0}}\mathbb{P}(Z = 1 \mid X = 1, Y = n) + \frac{p^E_{10}p_{y_0} - p^T_{11}p_x}{(1 - p_x)p_{y_0}}\Bigg)
    \end{align*}
    Subtracting these two, 
    \begin{align*}
        &= \frac{p^T_{11}}{p_{y_0}} - \sum_{n = 1}^N\frac{p_{y_n}}{p_{y_0}}\mathbb{P}(Z = 1 \mid X = 1, Y = n) - 1\\
        &+ \frac{p_x}{1 - p_x}\sum_{n = 1}^N\frac{p_{y_n}}{p_{y_0}}\mathbb{P}(Z = 1 \mid X = 1, Y = n) + \frac{p^E_{10}p_{y_0} - p^T_{11}p_x}{(1 - p_x)p_{y_0}}\\
        &= \frac{p^E_{10}p_{y_0} - p^T_{11}p_x}{(1 - p_x)p_{y_0}} + \frac{p^T_{11}}{p_{y_0}} - 1 + \sum_{n = 1}^N\frac{p_{y_n}}{p_{y_0}}P(Z = 1 \mid X = 1, Y = n)\Bigg(\frac{p_x}{1 - p_x} - 1\Bigg)\\
        &= \frac{p^E_{10}p_{y_0} - p^T_{11}p_x + p^T_{11}(1 - p_x) - (1 - p_x)p_{y_0}}{(1 - p_x)p_{y_0}}   + \sum_{n = 1}^N\frac{p_{y_n}}{p_{y_0}}P(Z = 1 \mid X = 1, Y = n)\Bigg(\frac{p_x}{1 - p_x} - 1\Bigg)\\
        &= \frac{p^E_{10} - (1 - p_x)}{1 - p_x} + \frac{p^T_{11}}{p_{y_0}}\Big(1 - \frac{p_x}{1 - p_x}\Big)  + \sum_{n = 1}^N\frac{p_{y_n}}{p_{y_0}}P(Z = 1 \mid X = 1, Y = n)\Bigg(\frac{p_x}{1 - p_x} - 1\Bigg)\\
        &= \frac{p^E_{10} - (1 - p_x)}{1 - p_x} + \frac{\sum_{n = 1}^N p_{y_n}P(Z = 1 \mid X = 1, Y = n) - p^T_{11}}{p_{y_0}}\Bigg(\frac{p_x}{1 - p_x} - 1\Bigg) 
    \end{align*}

    Moving on to $\max\{0, P(Z = 1 \mid X = 1, Y = i) - P(Z = 0 \mid X = 0, Y = i)\}$, the probability $P(Z = 0 \mid X = 0, Y = i)$ for $i \in \{1, \dots ,N\}$ can be expressed in terms of the free parameters as
    \begin{align*}
        &P(Z = 0 \mid X = 0, Y = i) = 1 -\Bigg( \frac{p^E_{1i}}{1 - p_x} - \frac{p_x}{1 - p_x}\mathbb{P}(Z = 1 \mid X = 1, Y = i)\Bigg)\\
        &\implies P(Z = 1 \mid X = 1, Y = i) - P(Z = 0 \mid X = 0, Y = i) \\
        &= P(Z = 1 \mid X = 1, Y = i) - 1 + \frac{p^E_{1i}}{1 - p_x} - \frac{p_x}{1 - p_x}\mathbb{P}(Z = 1 \mid X = 1, Y = i)\\
        &= P(Z = 1 \mid X = 1, Y = i)\Bigg(1 - \frac{p_x}{1 - p_x}\Bigg) + \frac{p^E_{11} - (1 - p_x)}{1 - p_x}
    \end{align*}
    This concludes the proof for the terms in $\Gamma$. Following a similar approach for the terms in $\Delta$, first consider
    \begin{align*}
        &\mathbb{P}(Z = 1 \mid X = 1, Y = 0) - \mathbb{P}(Z = 1 \mid X = 0, Y = 0)\\
        &= \frac{p^T_{11}}{p_{y_0}} - \sum_{n = 1}^N\frac{p_{y_n}}{p_{y_0}}\mathbb{P}(Z = 1 \mid X = 1, Y = n) - \Bigg(\frac{p_x}{1 - p_x}\sum_{n = 1}^N\frac{p_{y_n}}{p_{y_0}}\mathbb{P}(Z = 1 \mid X = 1, Y = n)\\
        &+ \frac{p^E_{10}p_{y_0} - p^T_{11}p_x}{(1 - p_x)p_{y_0}}\Bigg)\\
        &= \frac{p^T_{11} - p^E_{10}p_{y_0} - \sum_{n = 1}^N p_{y_n}\mathbb{P}(Z = 1\mid X = 1, Y = n)}{(1 - p_x)p_{y_0}}
    \end{align*}

    And for the terms of the form $\max\{0, \mathbb{P}(Z = 1 \mid X = 1, Y = i) - \mathbb{P}(Z = 1 \mid X = 0, Y = i)\}$ for $i \in \{1, \dots, N\}$
    \begin{align*}
        &\mathbb{P}(Z = 1 \mid X = 0, Y = i) = \frac{p^E_{1i}}{1 - p_x} - \frac{p_x}{1 - p_x}\mathbb{P}(Z = 1 \mid X = 1, Y = i)\\
        &\implies \mathbb{P}(Z = 1 \mid X = 1, Y = i) - \mathbb{P}(Z = 1 \mid X = 0, Y = i) =\\
        &\mathbb{P}(Z = 1 \mid X = 1, Y = i) - \frac{p^E_{1i}}{1 - p_x} + \frac{p_x}{1 - p_x}\mathbb{P}(Z = 1 \mid X = 1, Y = i)\\
        &= \frac{\mathbb{P}(Z = 1 \mid X = 1, Y = i) - p^E_{1i}}{1 - p_x}
    \end{align*}
    This concludes the proof of the Lemma.
\end{proof}
\begin{proof}[\textbf{Proof for Lemma \ref{lem:max-operator-gamma-0}}]
From the result of Lemma \ref{lemma:RREF-y}, $\mathbb{P}(Z = 1 \mid X = 1, Y = 0) - \mathbb{P}(Z = 0 \mid X = 0, Y = 0)$ can be expressed in terms of the free parameter as
    \begin{align*}
        \frac{p^E_{10} - (1 - p_x)}{1 - p_x} + \frac{\sum_{n = 1}^N p_{y_n}P(Z = 1 \mid X = 1, Y = n) - p^T_{11}}{p_{y_0}}\Bigg(\frac{p_x}{1 - p_x} - 1\Bigg)
    \end{align*}
Now, we use a proof by contra-positive to first show that
    \begin{align*}
        &\frac{p^E_{10} - (1 - p_x)}{1 - p_x} + \frac{\sum_{n = 1}^N p_{y_n}P(Z = 1 \mid X = 1, Y = n) - p^T_{11}}{p_{y_0}}\Bigg(\frac{p_x}{1 - p_x} - 1\Bigg) < 0\\
        &\implies \sum_{n = 1}^N p_{y_n}P(Z = 1 \mid X = 1, Y = n) < \frac{p_{y_0}((1 - p_x) - p^E_{10})}{p_x - (1 - p_x)} + p^T_{11}
    \end{align*}
    To see this, note that
    \begin{align*}
        &\frac{p^E_{10} - (1 - p_x)}{1 - p_x} + \frac{\sum_{n = 1}^N p_{y_n}P(Z = 1 \mid X = 1, Y = n) - p^T_{11}}{p_{y_0}}\Bigg(\frac{p_x}{1 - p_x} - 1\Bigg) < 0\\
        &\implies p_{y_0}\frac{p^E_{10} - (1 - p_x)}{p_{y_0}(1 - p_x)} + \frac{\sum_{n = 1}^N p_{y_n}P(Z = 1 \mid X = 1, Y = n) - p^T_{11}}{p_{y_0}}\Bigg(\frac{p_x - (1 - p_x)}{1 - p_x} \Bigg) < 0\\
        &\implies p_{y_0}(p^E_{10} - (1 - p_x)) + (\sum_{n = 1}^N p_{y_n}P(Z = 1 \mid X = 1, Y = n) - p^T_{11})\Bigg(p_x - (1 - p_x)\Bigg) < 0\\
        &\implies \sum_{n = 1}^N p_{y_n}P(Z = 1 \mid X = 1, Y = n) < \frac{p_{y_0}((1 - p_x) - p^E_{10})}{p_x - (1 - p_x)} + p^T_{11}\\
    \end{align*}
    This concludes the proof.
\end{proof}

\begin{proof}[\textbf{Proof for Lemma \ref{lem:max-operator-gamma-0-positive}}]
Following a proof by contra-positive, the contra-positive is given as
    \begin{align*}
        p^T_{11} - y_0 < \frac{p_{y_0}((1 - p_x) - p^E_{10})}{p_x - (1 - p_x)} + p^T_{11} \implies p^E_{10} < p_x
    \end{align*}
    To see this, note that
    \begin{align*}
        &p^T_{11} - p_{y_0} < \frac{p_{y_0}((1 - p_x) - p^E_{10})}{p_x - (1 - p_x)} + p^T_{11}\\
        &\implies - 1 < \frac{((1 - p_x) - p^E_{10})}{p_x - (1 - p_x)}\\
        &\implies  p_x  >  p^E_{10}
    \end{align*}
    This concludes the proof.
\end{proof}

\begin{proof}[\textbf{Proof for Lemma \ref{lem:gamma-0-lower-side-px}}] Following a proof by contra-positive, the contra-positive is given as
    \[
    \frac{p^T_{11}p_x - p^E_{10}p_{y_0}}{p_x} \geq \frac{p_{y_0}((1 - p_x) - p^E_{10})}{p_x - (1 - p_x)} + p^T_{11} \implies p^E_{10} \geq p_x
    \]
    To see this, note that
    \begin{align*}
        &\frac{p^T_{11}p_x - p^E_{10}p_{y_0}}{p_x} \geq \frac{p_{y_0}((1 - p_x) - p^E_{10})}{p_x - (1 - p_x)} + p^T_{11}\\
        &\implies \frac{- p^E_{10}}{p_x} \geq \frac{((1 - p_x) - p^E_{10})}{p_x - (1 - p_x)}\\
        &\implies - p^E_{10}p_x + p^E_{10}(1 - p_x) \geq (1 - p_x)p_x - p^E_{10}p_x\\
        &\implies p^E_{10} \geq p_x
    \end{align*}
    This concludes the proof.
\end{proof}
\begin{proof}[\textbf{Proof for Lemma \ref{lem:gamma-0-lower-side-1-minus-px} and Lemma \ref{lem:gamma-0-middle}}]
A similar proof by contra positive approach can be used to prove this Lemma.\\
\end{proof}
\begin{proof}[\textbf{Proof for Lemma \ref{lem:max-operator-gamma-i}}] From the result of Lemma \ref{lem:max-operator-expressions}, $\mathbb{P}(Z = 1 \mid X = 1, Y = i) - \mathbb{P}(Z = 0 \mid X = 0, Y = i)\}$ can be expressed in terms of the free parameter as 
\[
P(Z = 1 \mid X = 1, Y = i)\Bigg(1 - \frac{p_x}{1 - p_x}\Bigg) + \frac{p^E_{11} - (1 - p_x)}{1 - p_x}
\]
Following a proof by contra-positive, we write the contra-positive as
\begin{align*}
&P(Z = 1 \mid X = 1, Y = i)\Bigg(1 - \frac{p_x}{1 - p_x}\Bigg) + \frac{p^E_{11} - (1 - p_x)}{1 - p_x} < 0\\
&\implies P(Z = 1 \mid X = 1, Y = i) > \frac{p^E_{1i} - (1 - p_x)}{p_x - (1 - p_x)}    
\end{align*}

To see this, note that
\begin{align*}
    &P(Z = 1 \mid X = 1, Y = i)\Bigg(1 - \frac{p_x}{1 - p_x}\Bigg) + \frac{p^E_{11} - (1 - p_x)}{1 - p_x} < 0\\
    &\implies P(Z = 1 \mid X = 1, Y = i)\Bigg(\frac{(1 - p_x) - p_x}{1 - p_x}\Bigg)  < \frac{(1 - p_x) - p^E_{11}}{1 - p_x}\\
    &\implies P(Z = 1 \mid X = 1, Y = i)\Bigg((1 - p_x) - p_x\Bigg)  < (1 - p_x) - p^E_{11}\\
    &\implies P(Z = 1 \mid X = 1, Y = i)  >  \frac{p^E_{11} - (1 - p_x)}{p_x - (1 - p_x)}
\end{align*}
This concludes the proof.
\end{proof}
\begin{proof}[\textbf{Proof for Lemma \ref{lem:max-operator-gamma-i-positive}}] A similar proof by contra positive approach as before can be carried out to see this. 
\end{proof}

\begin{proof}[\textbf{Proof for Lemma \ref{lem:max-operator-gamma-i-0-1}}]
Following a proof by contra-positive, the contra-positive is given as
    \[
    \frac{p^E_{11}}{p_x} \leq \frac{p^E_{1i} - (1 - p_x)}{p_x - (1 - p_x)} \implies p^E_{1i} \geq p_x
    \]
    To see this, note that
    \begin{align*}
        &\frac{p^E_{11}}{p_x} \leq \frac{p^E_{1i} - (1 - p_x)}{p_x - (1 - p_x)}\\
        &\implies p^E_{11}p_x - p^E_{11}(1 - p_x) \leq p^E_{1i}p_x - p_x(1 - p_x)\\
        &\implies p^E_{11} \geq p_x
    \end{align*}
    This concludes the proof.
\end{proof}
\begin{proof}[\textbf{Proof for Lemma 18}]
A similar proof by contra positive approach can be applied here.
\end{proof} 

\begin{proof}[\textbf{Proof for Lemma \ref{lem:max-operarot-simultaneous-open}}]
To prove an if and only if, we first prove the forward implication, i.e. the existence of a set of values for the free parameter such that every max operator satisfies Eq. \ref{eq:max-operator-opening-condition} implies $p_{11} + p_{10} - 1 \geq 0$, and then we prove the reverse implication, thereby establishing the if and only if. 

First, from Lemma \ref{lem:max-operator-gamma-0} and Lemma \ref{lem:max-operator-gamma-i}, note that the free parameters must satisfy the following conditions simultaneously to satisfy Eq. \ref{eq:max-operator-opening-condition}. 
\begin{equation}\label{eq:eq-1}
    \sum_{n = 1}^N p_{y_n}P(Z = 1 \mid X = 1, Y = n) \geq \frac{p_{y_0}((1 - p_x) - p^E_{10})}{p_x - (1 - p_x)} + p_{11}
\end{equation}
And for all $i \in \{1, \dots N\}$
\begin{equation}\label{eq:eq-2}
    P(Z = 1 \mid X = 1, Y = i) \leq \frac{p^E_{1i} - (1 - p_x)}{p_x - (1 - p_x)}
\end{equation}
For the bounds in Eq. \ref{eq:eq-1} and Eq. \ref{eq:eq-2} to hold simultaneously, the following must hold
\begin{align*}
    &\frac{p_{y_0}((1 - p_x) - p^E_{10})}{p_x - (1 - p_x)} + p^T_{11} \leq \sum_{i = 1}^N p_{y_i}\frac{p^E_{1i} - (1 - p_x)}{p_x - (1 - p_x)}\\
    &\implies p_{y_0}((1 - p_x) - p^E_{10}) + p^T_{11}(p_x - (1 - p_x)) +  \leq \sum_{i = 1}^N p_{y_i}(p^E_{1i} - (1 - p_x))\\
    &\implies (1 - p_x) + p^T_{11}(p_x - (1 - p_x)) +  \leq \sum_{i = 1}^N p_{y_i}p^E_{1i} + p^E_{10}p_{y_0}\\
    &\implies (1 - p_x) - p^T_{11}(1 - p_x) +  \leq p^T_{10}(1 - p_x)\\
    &\implies 0 \leq p^T_{11} + p^T_{10} - 1
\end{align*}

Next, to prove the backward implication, we utilize a proof by contra positive, i.e. we show that 
\[
\frac{p_{y_0}((1 - p_x) - p^E_{10})}{p_x - (1 - p_x)} + p^T_{11} > \sum_{i = 1}^N p_{y_i}\frac{p^E_{1i} - (1 - p_x)}{p_x - (1 - p_x)} \implies p^T_{11} + p^T_{10} - 1 < 0
\]

To see this, note that
\begin{align*}
    &\frac{p_{y_0}((1 - p_x) - p^E_{10})}{p_x - (1 - p_x)} + p^T_{11} > \sum_{i = 1}^N p_{y_i}\frac{p^E_{1i} - (1 - p_x)}{p_x - (1 - p_x)}\\
    &\implies p_{y_0}((1 - p_x) - p^E_{10}) + p^T_{11}(p_x - (1 - p_x)) > \sum_{i = 1}^N p_{y_i}(p^E_{1i} - (1 - p_x))\\
    &\implies (1 - p_x)  - p^T_{11}(1 - p_x) > p^T_{10}p_x\\
    &\implies 0  > p^T_{11} + p^T_{10} - 1
\end{align*}
This concludes the proof for the lemma.
\end{proof}

Lemmas \ref{lem:delta-max-operator-0-non-0} through \ref{lem:delta-max-operator-i-non-0-bounds-test} can be proved using a similar approach as presented for operators in $\Gamma$. 

\begin{proof}[\textbf{Proof for Lemma \ref{lem:max-operator-simul-open-delta}}]
To prove an \emph{if and only if}, we first show that all max operators simultaneously satisfying Eq. \ref{eq:delta-simul-condition} implies $p^T_{11} - p^T_{10} \geq 0$, and then we prove the backward implication.

Starting with the forward implication, note that for all max operators to simultaneously satisfy Eq. \ref{eq:delta-simul-condition} the following conditions (obtained from Lemma \ref{lem:delta-max-operator-0-non-0} and Lemma \ref{lem:delta-max-operator-i-non-0}) must hold
\[
    p^T_{11} - p^E_{10}p_{y_0} \geq  \sum_{n = 1}^N p_{y_n}P(Z = 1\mid X = 1, Y = n)
\]
And for all $i \in \{1, \dots N\}$
\[
    P(Z = 1 \mid X = 1, Y = i) \geq p^E_{1i}
\]
This is equivalent to
\begin{align*}
    &\sum_{i = 1}^N p_{y_i}p^E_{1i} \leq p^T_{11} - p^E_{10}p_{y_0}\\
    &\implies \sum_{i = 0}^N p_{y_i}p^E_{1i} \leq p^T_{11} \\
    &\implies p^T_{11}p_x + p^T_{10}(1 - p_x) \leq p^T_{11} \\
    &\implies 0 \leq (1 - p_x)(p^T_{11} - p^T_{10})\\
\end{align*}

This proves the forward implication. A similar proof for the backward implication can be provided using a proof by contra positive as well, concluding the proof for this lemma. 
\end{proof}

\begin{proof}[\textbf{Proof for Lemma \ref{lem:pns-all-right-extereme}}]
Since $p^E_{1i} \geq p_x$ for all $i$, then all the indicators in $\Phi_i$ will evaluate to $1$, as a result $p^T_{11} - \sum_{i = 0}^N \Phi_i$ will equal
    \begin{align*}
        &p^T_{11} - \sum_{i = 0}^N \frac{p_{y_i}(p^E_{1i} - p_x)}{1 - p_x}\\
        &\implies = \frac{p^T_{11}(1 - p_x) - \sum_{i = 0}^N p_{y_i}(p^E_{1i} - p_x)}{1 - p_x}\\
        &\implies = \frac{p^T_{11}(1 - p_x) - p^T_{11}p_x - p^T_{10}(1 - p_x) + p_x}{1 - p_x}\\
        &\implies = \frac{p^T_{11}(1 - p_x) - p^T_{11}p_x + p^T_{00}(1 - p_x) - (1 - p_x) + p_x}{1 - p_x}\\
        &\implies = p^T_{01}(\frac{p_x}{1 - p_x} - 1) + p^T_{00}\\
    \end{align*}
    And since $p_x > 1 - p_x$, this quantity will be greater than $p^T_{00}$.
\end{proof}

\begin{proof}[\textbf{Proof for Lemma \ref{lem:pns-all-left-extereme}}]
Since $p^E_{1i} \leq 1 - p_x$ for all $i$, then all the indicators in $\Theta_i$ will evaluate to $1$, as a result $p^T_{00} - \sum_{i = 0}^N \Theta_i$ will equal
\begin{align*}
    &p^T_{00} - \sum_{i = 0}^N \frac{p_{y_i}((1 - p_x) - p^E_{1i})}{1 - p_x}\\
    &\implies = \frac{p^T_{00}(1 - p_x) - (1 - p_x) + \sum_{i = 0}^N p_{y_i}p^E_{1i}}{1 - p_x}\\
    &\implies = \frac{-p^T_{10}(1 - p_x) + \sum_{i = 0}^N p_{y_i}p^E_{1i}}{1 - p_x}\\
    &\implies = \frac{p^T_{11}p_x}{1 - p_x}
\end{align*}
And since $p_x > 1 - p_x$, this concludes the proof.
\end{proof}

\begin{proof}[\textbf{Proof for Lemma \ref{lemma:RREF-y-delta}}]
This lemma is proven in two steps, first using mathematical induction, and then simplifying the final RREF based on properties of probability distributions. 

First, given the following constraints, 
\begin{align*}
        p^T_{10} &= \sum_{j = 0}^N P^T(Z = 1\mid X = 0, Y = j)p_{y_j}\\
        p^T_{11} &= \sum_{j = 0}^N P^T(Z = 1\mid X = 1, Y = j)p_{y_j}\\
        p^E_{10} &= P^T(Z = 1\mid X = 0, Y = 0)(1 - p_x - \delta_X)\\
        &\hspace{0.2in} + P^T(Z = 1\mid X = 1, Y = 0)(p_x + \delta_X)\\
        \vdots\\
        p^E_{1N} &= P^T(Z = 1\mid X = 0, Y = N)(1 - p_x - \delta_X)\\
        &\hspace{0.2in} + P^T(Z = 1\mid X = 1, Y = N)(p_x + \delta_X)\\
\end{align*}

The above equations can be represented in matrix form $A\mathbf{x} = b$ as
{\scriptsize
\begin{align*}
A &= \begin{bmatrix}
    p_{y_0} & 0 & p_{y_1} & 0 & \dots & p_{y_{N - 1}} & 0 & p_{y_N} & 0\\
    0 & p_{y_0} & 0 & p_{y_1} & \dots & 0 & p_{y_{N - 1}}& 0 & p_{y_N}\\
    1 - p_{x} - \delta_X & p_{x} + \delta_X & 0 & 0 & \dots & 0 & 0 & 0 & 0\\
    0 & 0 & 1 - p_x - \delta_X & p_x + \delta_X & \dots & 0 & 0 & 0 & 0\\
    \vdots & \ddots & & & & & & & \vdots\\
    0 & 0 & \dots & 0 & 0 & 1 - p_x - \delta_X & p_x + \delta_X & 0 & 0\\
    0 & 0 & \dots & 0 & & 0 & 0 & 1 - p_x - \delta_X & p_x + \delta_X\\
\end{bmatrix}\\
b &= \begin{bmatrix}
    p^T_{10}\\
    p^T_{11}\\
    p^E_{10}\\
    p^E_{11}\\
    p^E_{12}
    \vdots\\
    p^E_{1(N - 1)}\\
    p^E_{1N}\\
\end{bmatrix}
\hspace{1in}\mathbf{x} 
= \begin{bmatrix}
P(Z = 1 \mid X = 0, Y = 0)\\    
P(Z = 1 \mid X = 1, Y = 0)\\
P(Z = 1 \mid X = 0, Y = 1)\\    
P(Z = 1 \mid X = 1, Y = 1)\\
\vdots\\
P(Z = 1 \mid X = 0, Y = N)\\    
P(Z = 1 \mid X = 1, Y = N)\\
\end{bmatrix}
\end{align*}
}
Next, denote the augmented matrix $A | b$ as 
{\small
\begin{align*}
    \begin{bmatrix}[ccccccccc|c]
    p_{y_0} & 0 & p_{y_1} & 0 & \dots & p_{y_{N - 1}} & 0 & p_{y_N} & 0 & p_{10}\\
    0 & p_{y_0} & 0 & p_{y_1} & \dots & 0 & p_{y_{N - 1}}& 0 & p_{y_N} & p_{11}\\
    1 - p_x - \delta_X & p_x + \delta_X & 0 & 0 & \dots & 0 & 0 & 0 & 0 & p^E_{10}\\
    0 & 0 & 1 - p_x - \delta_X & p_x + \delta_X & \dots & 0 & 0 & 0 & 0 & p^E_{11}\\
    \vdots & \ddots & & & & & & & \vdots & \vdots \\
    0 & 0 & \dots & 0 & 0 & 1 - p_x - \delta_X & p_x + \delta_X & 0 & 0 & p^E_{1(N - 1)}\\
    0 & 0 & \dots & 0 & & 0 & 0 & 1 - p_x - \delta_X & p_x + \delta_X & p^E_{1N}\\
    \end{bmatrix}
\end{align*}
}


An identical induction approach to Lemma \ref{lemma:RREF-y} can be used to prove the RREF of $A|b$ will equal
{\small
\begin{align}\label{matrix:rref-delta-x} 
    \begin{bmatrix}
    1 &  0 &  0 &  -\frac{(p_x + \delta_X) p_{y_1}}{(1 - p_x - \delta_X)p_{y_0}} & \dots & 0 & -\frac{(p_x + \delta_X)p_{y_{N - 1}}}{(1 - p_x - \delta_X)p_{y_0}} & 0 & -\frac{(p_x + \delta_X)p_{y_N}}{(1 - p_x - \delta_X)p_{y_0}} & \frac{p^E_{10}p_{y_0} - p^T_{11}(p_x + \delta_X)}{(1 - p_x - \delta_X)p_{y_0}}\\
    0 &  1 &  0 & \frac{p_{y_1}}{p_{y_0}} & \dots & 0 & \frac{p_{y_{N - 1}}}{p_{y_0}} & 0 & \frac{p_{y_N}}{p_{y_0}} & \frac{p^T_{11}}{p_{y_0}}\\
    0 &  0 &  1 & \frac{p_x + \delta_X}{1 - p_x - \delta_X} &  \dots & 0 & 0 & 0 &  0 & \frac{p^E_{11}}{1 - p_x - \delta_X}\\
    \vdots &&&&&&& \ddots \\
    0 &  0 &  0 & 0 & \dots & 0 & 0 & 1 & \frac{p_x + \delta_X}{1 - p_x - \delta_X} & \frac{p^T_{10}(1 - p_x - \delta_X) + p^T_{11}(p_x + \delta_X) - \sum_{i = 0}^{N - 1} p^E_{1i}p_{y_i}}{p_{y_N}(1 - p_x - \delta_X)}\\
    0 &  0 &  0 & 0 & \dots & 0 & 0 & 0 & 0 & \frac{\sum_{i = 0}^{N} p^E_{1i}p_{y_i} - p^T_{10}(1 - p_x - \delta_X) - p^T_{11}(p_x + \delta_X)}{p_{y_N}} 
    \end{bmatrix}
\end{align}
}
Next, to complete this proof for this lemma, we prove the following identities. 
\[\frac{\sum_{j = 0}^{N} p^E_{1j}p_{y_j} - p^T_{10}(1 - p_x - \delta_X) - p^T_{11}(p_x + \delta_X)}{p_{y_N}} = 0\]

To prove this, note that
\[ p^T_{10}(1 - p_x - \delta_X) = \sum_{j = 0}^N P^T(Z = 1\mid X = 0, Y = j)p_{y_j}(1 - p_x - \delta_X)\]
Along with
\[
p^T_{11}(p_x + \delta_X) = \sum_{j = 0}^N P^T(Z = 1\mid X = 1, Y = j)p_{y_j}(p_x + \delta^i_X)
\]
Consequently, 
\begin{align*}
    &p^T_{10}(1 - p_x - \delta^0_X) + p^T_{11}(p_x + \delta^i_X) = \\
    & \sum_{j = 0}^N p_{y_j}\Bigg(P^T(Z = 1\mid X = 0, Y = j)(1 - p_x - \delta^0_X) + P^T(Z = 1\mid X = 1, Y = j)(p_x + \delta^i_X)\Bigg)\\
    &= \sum_{j = 0}^N p_{y_j}\Bigg(P^E(Z = 1\mid X = 0, Y = j)P^E(X = 0) + P^E(Z = 1\mid X = 1, Y = j)P^E(X = 1)\Big)\\
    &= \sum_{j = 0}^N p_{y_j}p^E_{1j}
\end{align*}

And this will be subtracted from $\sum_{j = 0}^N p_{y_j}p^E_{1ji}$, hence it equals $0$. 
\[
\frac{p^T_{10i}(1 - p_x- \delta^i_X) + p^T_{11i}(p_x + \delta^i_X) - \sum_{j = 0}^{N - 1} p^E_{1ji}p_{y_j}}{p_{y_N}(1 - p_x - \delta^i_X)} = 
\]

Following a similar approach as before, this will simplify to
\begin{align*}
    \frac{\sum_{j = 0}^N p_{y_j}p^E_{1j} - \sum_{j = 0}^{N - 1} p_{y_j}p^E_{1j}}{p_{y_N}(1 - p_x - \delta^i_X)} = \frac{p^E_{1N}}{1 - p_x - \delta^i_X}
\end{align*}
This concludes the proof of the Lemma.
\end{proof}

\begin{proof}[\textbf{Proof for Lemma \ref{lem:RREF-y-c-delta}}]
The proof for this Lemma follows two major steps, the first being a proof by double mathematical induction to represent the Reduced Row Echelon Form (RREF) for the system of equations for arbitrary dimensionality of $Y$ and $C$. The second part of the proof applies properties of probability distributions to express the system of equations in the form given in the Lemma.    

We start start with the proof by double mathematical induction, and state the proposition we prove. 

\textbf{Proposition}: Let $P(m, n)$ denote the proposition that given the system of equations corresponding to the constraints provided in Lemma, represented in matrix form $A_{m, n}\mathbf{x}_{m ,n } = \mathbf{b}_{m , n}$ provided below.
{\scriptsize
\begin{align*}
A_{m, n} &= \begin{bmatrix}
            A_{0, n} & [\mathbf{0}] & \dots & [\mathbf{0}]\\
            [\mathbf{0}] & A_{1, n} & \dots & [\mathbf{0}]\\
            [\mathbf{0}] & [\mathbf{0}] & \ddots & [\mathbf{0}]\\
            [\mathbf{0}] & [\mathbf{0}] & \dots & A_{m, n}
            \end{bmatrix}
\hspace{0.2in} b_{m, n} = \begin{bmatrix}
                            b_{0, n}\\
                            \vdots\\
                            b_{m, n}
                          \end{bmatrix}
\hspace{0.2in} \mathbf{x}_{m , n} = \begin{bmatrix}
                                    \mathbf{x}_{0, n}\\
                                    \vdots\\
                                    \mathbf{x}_{m, n}
                                    \end{bmatrix}
\intertext{Where $A_{i, n}$, $b_{i, n}$ and $\mathbf{x}_{i, n}$ will be of the form}
A_{i, n} &= \begin{bmatrix}
    p_{y_0} & 0 & p_{y_1} & 0 & \dots & p_{y_{N - 1}} & 0 & p_{y_N} & 0\\
    0 & p_{y_0} & 0 & p_{y_1} & \dots & 0 & p_{y_{N - 1}}& 0 & p_{y_N}\\
    1 - p_{x} - \delta^i_X & p_{x} + \delta^i_X & 0 & 0 & \dots & 0 & 0 & 0 & 0\\
    0 & 0 & 1 - p_x - \delta^i_X & p_x + \delta^i_X & \dots & 0 & 0 & 0 & 0\\
    \vdots & \ddots & & & & & & & \vdots\\
    0 & 0 & \dots & 0 & 0 & 1 - p_x - \delta^i_X & p_x + \delta^i_X & 0 & 0\\
    0 & 0 & \dots & 0 & & 0 & 0 & 1 - p_x - \delta^i_X & p_x - \delta^i_X\\
\end{bmatrix}\\
b_{i, n} &= \begin{bmatrix}
             p^T_{10i}\\
             p^T_{11i}\\
             p^E_{10i}\\
             p^E_{11i}\\
             p^E_{12i}
             \vdots\\
             p^E_{1(n - 1)i}\\
             p^E_{1ni}\\
            \end{bmatrix}
\hspace{1in} \mathbf{x}_{i, n} = \begin{bmatrix}
                                P(Z = 1 \mid X = 0, Y = 0, C = i)\\    
                                P(Z = 1 \mid X = 1, Y = 0, C = i)\\
                                P(Z = 1 \mid X = 0, Y = 1, C = i)\\    
                                P(Z = 1 \mid X = 1, Y = 1, C = i)\\
                                \vdots\\
                                P(Z = 1 \mid X = 0, Y = N, C = i)\\    
                                P(Z = 1 \mid X = 1, Y = N, C = i)\\
                                \end{bmatrix}
\end{align*}
}
Given this system of equations, the RREF of the augmented matrix ${A \mid b}_{m, n}$ corresponding to this system of equations will be of the form
\begin{align*}
    {A \mid b}^R_{m, n} = \begin{bmatrix}
                        A^R_{0, n} & [\mathbf{0}] & [\mathbf{0}] & b^R_{0, n} \\
                        [\mathbf{0}] & A^R_{1, n} & [\mathbf{0}] & b^R_{1, n}\\
                        [\mathbf{0}] & [\mathbf{0}] & \ddots & \vdots \\
                        [\mathbf{0}] & \dots & A^R_{m, n} & b^R_{m, n} \\
                        0 & \dots & 0 & \frac{\sum_{j = 0}^{N} p^E_{1j0}p_{y_j} - p^T_{10}(1 - p_x - \delta^0_X) - p^T_{110}(p_x + \delta^0_X)}{p_{y_N}}\\
                        \vdots & \dots & 0 & \vdots\\
                        0 & \dots & 0 & \frac{\sum_{j = 0}^{N} p^E_{1j0}p_{y_j} - p^T_{10}(1 - p_x - \delta^0_X) - p^T_{110}(p_x + \delta^0_X)}{p_{y_N}}\\
                \end{bmatrix}
\end{align*}
Where $A^R_{i, n}$ and $b^R_{i, n}$ will be of the form
\begin{align*}
    \label{matrix:rref-with-covariates} 
    A^R_{i, n} = \begin{bmatrix}
    1 &  0 &  0 &  -\frac{(p_x + \delta^i_X)p_{y_1}}{(1 - p_x - \delta^i_X)p_{y_0}} & \dots & 0 & -\frac{(p_x + \delta^i_X) p_{y_{N - 1}}}{(1 - p_x - \delta^i_X)p_{y_0}} & 0 & -\frac{(p_x + \delta^i_X)p_{y_N}}{(1 - p_x- \delta^i_X)p_{y_0}}\\
    0 &  1 &  0 & \frac{p_{y_1}}{p_{y_0}} & \dots & 0 & \frac{p_{y_{N - 1}}}{p_{y_0}} & 0 & \frac{p_{y_N}}{p_{y_0}}\\
    0 &  0 &  1 & \frac{p_x + \delta^i_X}{1 - p_x - \delta^i_X} &  \dots & 0 & 0 & 0 &  0\\
    \vdots &&&&&&& \ddots\\
    0 &  0 &  0 & 0 & \dots & 0 & 0 & 1 & \frac{p_x + \delta^i_X}{1 - p_x - \delta^i_X} \\
    \end{bmatrix}
\end{align*}

And $b^R_{i, n}$ will be of the form
\begin{align*}
\begin{bmatrix}
    \frac{p^E_{10i}p_{y_0} - p^T_{11i}p_x}{(1 - p_x - \delta^i_X)p_{y_0}}\\
    \frac{p^T_{11i}}{p_{y_0}}\\
    \vdots\\
    \frac{p^T_{10i}(1 - p_x- \delta^i_X) + p^T_{11i}(p_x + \delta^i_X) - \sum_{j = 0}^{N - 1} p^E_{1ji}p_{y_j}}{p_{y_N}(1 - p_x - \delta^i_X)}
\end{bmatrix}    
\end{align*}

To prove $P(m, n)$ using double mathematical induction, we first prove the base case with $m = n = 1$.\\
\newline
\textbf{Base Case}: Using a symbolic solver to put the matrix in RREF, we obtain the RREF form of $A \mid b$ as
{\scriptsize
\begin{align*}
\begin{bmatrix}
1 &  0 &  0 &  -\frac{p_{y_1}(\delta^0_X + p_x)}{p_{y_0}(1 - p_x - \delta^0_X)} &  0 &  0 &  0 & 0 & \frac{p^E_{100}p_{y_0} - p^T_{110}(\delta^0_X + p_x)}{p_{y_0}(1 - p_x - \delta^0_X)}\\
0 &  1 &  0 & \frac{p_{y_1}}{p_{y_0}} &  0 &  0 &  0 & 0 & \frac{p^T_{110}}{p_{y_0}}\\
0 &  0 &  1 & \frac{\delta^0_X + x}{1 - x - \delta^0_X} &  0 &  0 &  0 & 0 & \frac{p^T_{100}(1 - p_x - \delta^0_X) + p^T_{110}(\delta^0_X + x) - p^E_{100}p_{y_0}}{p_{y_1}(1 - p_x - \delta^0_X)} \\
0 &  0 &  0 & 0 &  1 &  0 &  0 &  -\frac{p_{y_1}(\delta^1_X + x)}{p_{y_0}(1 - p_x - \delta^1_X)} & \frac{p^E_{101}p_{y_0} - p^T_{111}(\delta^1_X + p_x)}{p_{y_0}(1 - p_x - \delta^1_X)}\\
0 &  0 &  0 & 0 &  0 &  1 &  0 & \frac{p_{y_1}}{p_{y_0}} & \frac{p^T_{111}}{p_{y_0}} \\
0 &  0 &  0 & 0 &  0 &  0 &  1 & \frac{p_x + \delta^1_X}{1 - x - \delta^1_X} &  \frac{p^T_{101}(1 - p_x - \delta^1_X) + p^T_{111}(\delta^1_X + x) - p^E_{101}p_{y_0}}{p_{y_1}(1 - p_x - \delta^1_X)}\\
0 &  0 &  0 & 0 &  0 &  0 &  0 & 0 & \frac{p^E_{110}p_{y_1} + p^E_{100}p_{y_0} - p_{100}(1 - p_x - \delta^0_X) - p_{110}(\delta^0_X + p_x)}{p_{y_1}} \\
0 &  0 &  0 & 0 &  0 &  0 &  0 & 0 &  \frac{p^E_{111}p_{y_1} + p^E_{101}p_{y_0} - p^T_{101}(1 - p_x - \delta^1_X) - p_{111}(\delta^1_X + p_x)}{p_{y_1}}
\end{bmatrix}
\end{align*}
}
And by comparison, this matches the base case. Hence the base case holds.\\ 
\newline
\textbf{Induction Step I}: Here we prove that if $P(1, n - 1)$ holds, then $P(1, n)$ holds as well. For $P(1, n)$, $A|b$ will look like

\begin{align*}
{A\mid b}_{1, n} &= \begin{bmatrix}
                    A_{0, n} & [\mathbf{0}] & b_{0, n} \\
                    [\mathbf{0}] & A_{1, n} & b_{1, n}
            \end{bmatrix}
\end{align*}

Since $A_{0, n}$ and $A_{1, n}$ are non-overlapping diagional sub-matrices of ${A \mid b}_{1, n}$, a similar proof to Lemma X for each level of of $C$ can be applied, i.e. for each pair $A_{0, n}$, $b_{0,n}$ and $A_{1, n}$, $b_{1,n}$ to put each of them in their RREF matrix separately. Then the last rows correspoding to the RREF of $A_{0, n}$, $b_{0,n}$ and $A_{1, n}$, $b_{1,n}$ can be shuffled as needed to put ${A \mid b}_{1, n}$ in RREF, proving the first step of the induction.\\
\newline
\textbf{Induction Step II:} Next, we must prove that for all $n \geq 1$, $P(m, n)$ is true then $P(m + 1, n)$ is true. For $P(m + 1, n)$, the system of equations is represented in matrix form as
{\scriptsize
\begin{align*}
A_{m, n} &= \begin{bmatrix}
            A_{0, n} & [\mathbf{0}] & \dots & [\mathbf{0}] & [\mathbf{0}]\\
            [\mathbf{0}] & A_{1, n} & \dots & [\mathbf{0}] & [\mathbf{0}]\\
            [\mathbf{0}] & [\mathbf{0}] & \ddots & [\mathbf{0}] &  [\mathbf{0}]\\
            [\mathbf{0}] & [\mathbf{0}] & \dots & A_{m, n} & [\mathbf{0}]\\
            [\mathbf{0}] & [\mathbf{0}] & \dots & [\mathbf{0}] & A_{m + 1, n}
            \end{bmatrix}
\hspace{0.2in} b_{m, n} = \begin{bmatrix}
                            b_{0, n}\\
                            \vdots\\
                            b_{m, n}\\
                            b_{m + 1, n}
                          \end{bmatrix}
\hspace{0.2in} \mathbf{x}_{m , n} = \begin{bmatrix}
                                    \mathbf{x}_{0, n}\\
                                    \vdots\\
                                    \mathbf{x}_{m, n}\\
                                    \mathbf{x}_{m + 1, n}
                                    \end{bmatrix}
\intertext{Where $A_{i, n}$, $b_{i, n}$ and $\mathbf{x}_{i, n}$ will be of the form}
A_{i, n} &= \begin{bmatrix}
    p_{y_0} & 0 & p_{y_1} & 0 & \dots & p_{y_{N - 1}} & 0 & p_{y_N} & 0\\
    0 & p_{y_0} & 0 & p_{y_1} & \dots & 0 & p_{y_{N - 1}}& 0 & p_{y_N}\\
    1 - p_{x} - \delta^i_X & p_{x} + \delta^i_X & 0 & 0 & \dots & 0 & 0 & 0 & 0\\
    0 & 0 & 1 - p_x - \delta^i_X & p_x + \delta^i_X & \dots & 0 & 0 & 0 & 0\\
    \vdots & \ddots & & & & & & & \vdots\\
    0 & 0 & \dots & 0 & 0 & 1 - p_x - \delta^i_X & p_x + \delta^i_X & 0 & 0\\
    0 & 0 & \dots & 0 & & 0 & 0 & 1 - p_x - \delta^i_X & p_x - \delta^i_X\\
\end{bmatrix}\\
b_{i, n} &= \begin{bmatrix}
             p^T_{10i}\\
             p^T_{11i}\\
             p^E_{10i}\\
             p^E_{11i}\\
             p^E_{12i}
             \vdots\\
             p^E_{1(n - 1)i}\\
             p^E_{1ni}\\
            \end{bmatrix}
\hspace{1in} \mathbf{x}_{i, n} = \begin{bmatrix}
                                P(Z = 1 \mid X = 0, Y = 0, C = i)\\    
                                P(Z = 1 \mid X = 1, Y = 0, C = i)\\
                                P(Z = 1 \mid X = 0, Y = 1, C = i)\\    
                                P(Z = 1 \mid X = 1, Y = 1, C = i)\\
                                \vdots\\
                                P(Z = 1 \mid X = 0, Y = N, C = i)\\    
                                P(Z = 1 \mid X = 1, Y = N, C = i)\\
                                \end{bmatrix}
\end{align*}
}
Since we assume $P(m, n)$ is true, the submatrix corresponding to ${A \mid b}_{m, n}$ in ${A \mid b}_{m + 1, n}$ is non-overlapping with $A_{m + 1, n}$ and $b_{m + 1, n}$, and can be put into RREF form, presented below.
\begin{align*}
    {A \mid b}_{m + 1, n} = \begin{bmatrix}
                        A^R_{0, n} & [\mathbf{0}] & \dots & [\mathbf{0}] & [\mathbf{0}]&  b^R_{0, n} \\
                        [\mathbf{0}] & A^R_{1, n} & \dots & [\mathbf{0}] & [\mathbf{0}] & b^R_{1, n}\\
                        [\mathbf{0}] & [\mathbf{0}] & \ddots & [\mathbf{0}] & [\mathbf{0}] & \vdots \\
                        [\mathbf{0}] & [\mathbf{0}] & \dots & A^R_{m, n} & [\mathbf{0}] & b^R_{m, n} \\
                        0 & \dots & 0 & 0 & [\mathbf{0}] & \frac{\sum_{j = 0}^{N} p^E_{1j0}p_{y_j} - p^T_{10}(1 - p_x - \delta^0_X) - p^T_{110}(p_x + \delta^0_X)}{p_{y_N}}\\
                        \vdots & \dots & 0 & 0 & [\mathbf{0}] & \vdots\\
                        0 & \dots & 0 & 0 & [\mathbf{0}] &\frac{\sum_{j = 0}^{N} p^E_{1j0}p_{y_j} - p^T_{10}(1 - p_x - \delta^0_X) - p^T_{110}(p_x + \delta^0_X)}{p_{y_N}}\\
                        0 & \dots & 0 & 0 & A_{m + 1, n} & b_{m + 1, n}
                \end{bmatrix}
\end{align*}
Now, putting $A_{m+ 1, n}$ and $b_{m + 1, n}$ in RREF form using a similar approach to Lemma \ref{lemma:RREF-y}, and shuffling all the rows that only have $0$ in every entry except the last gives us the RREF as stated and proved $P(m + 1, n)$. This concludes the proof of the double induction. 

Next, to complete this proof, we prove the following identities. 
\[\frac{\sum_{j = 0}^{N} p^E_{1ji}p_{y_j} - p^T_{10i}(1 - p_x - \delta^i_X) - p^T_{11i}(p_x + \delta^i_X)}{p_{y_N}} = 0\]

To prove this, note that
\[  p^T_{10i}(1 - p_x - \delta^0_X) = \sum_{j = 0}^N P(Z = 1\mid X = 0, Y = j, C = i)p_{y_j}(1 - p_x - \delta^0_X)\]
Along with
\[
p^T_{11i}(p_x + \delta^i_X) = \sum_{j = 0}^N P(Z = 1\mid X = 1, Y = j, C = i)p_{y_j}(p_x + \delta^i_X)
\]

Consequently, 
{\small
\begin{align*}
    &p^T_{10i}(1 - p_x - \delta^0_X) + p^T_{11i}(p_x + \delta^i_X) = \\
    & \sum_{j = 0}^N p_{y_j}\Bigg(P(Z = 1\mid X = 0, Y = j, C = i)(1 - p_x - \delta^0_X) + P(Z = 1\mid X = 1, Y = j, C = i)(p_x + \delta^i_X)\Big)\\
    &= \sum_{j = 0}^N p_{y_j}\Bigg(P(Z = 1\mid X = 0, Y = j, C = i)P^E(X = 0 \mid C = i)\\
    &\hspace{0.4in} + P(Z = 1\mid X = 1, Y = j, C = i)P^E(X = 1 \mid C = i)\Big)\\
    &= \sum_{j = 0}^N p_{y_j}p^E_{1ji}
\end{align*}
}
And this will be subtracted from $\sum_{j = 0}^N p_{y_j}p^E_{1ji}$, hence it equals $0$. 
\[
\frac{p^T_{10i}(1 - p_x- \delta^i_X) + p^T_{11i}(p_x + \delta^i_X) - \sum_{j = 0}^{N - 1} p^E_{1ji}p_{y_j}}{p_{y_N}(1 - p_x - \delta^i_X)} 
\]

Following a similar approach as before, the above will simplify to
\begin{align*}
    \frac{\sum_{j = 0}^N p_{y_j}p^E_{1ji} - \sum_{j = 0}^{N - 1} p_{y_j}p^E_{1ji}}{p_{y_N}(1 - p_x - \delta^i_X)} = \frac{p^E_{1Ni}}{1 - p_x - \delta^i_X}
\end{align*}
This concludes the proof of the Lemma.
\end{proof}

\bibliography{references}

\end{document}